\documentclass{article}
\usepackage[utf8]{inputenc}
\usepackage{xr,amsmath,amsfonts,amssymb,graphics,epsfig,verbatim,bm,latexsym,url,amsbsy}
\usepackage{siunitx}

\usepackage{amsthm}
\usepackage{xr}
\usepackage{mathabx}
\newcommand{\ubar}[1]{\underaccent{\bar}{#1}}
\usepackage{xr}
\usepackage{natbib}
\usepackage{graphicx}
\usepackage{lscape}
\usepackage{subcaption}
\usepackage{color}
\usepackage{wrapfig}
\usepackage{mathtools}
\usepackage{placeins}
\usepackage{tgpagella}
\usepackage[framemethod=TikZ]{mdframed}
\usepackage{xcolor}
\usepackage{tikz}
\usetikzlibrary {positioning}
\usepackage{graphicx}
\newcommand{\indep}{\rotatebox[origin=c]{90}{$\models$}}
\usepackage{bm}
\usepackage{paralist}
\usepackage{accents}
\usepackage{enumerate}
\usepackage{algorithm}

\usepackage{booktabs}

\usepackage[noend,]{algorithmic}

\usepackage{chngcntr}

\counterwithin*{equation}{section}

\renewcommand{\Pr}{{\mathrm{P}}}

\newcommand{\Ep}{\mathbb{E}}

\newcommand{\Enk}{{\mathbb{E}_{n,k}}}
\newcommand{\Gnk}{{\mathbb{G}_{n,k}}}

\newcommand{\EN}{{\mathbb{E}_{N}}}

\newcommand{\G}{\mathbb{G}}
\newcommand{\GN}{\mathbb{G}_N}

\newcommand{\E}{\mathbb{E}}
\usepackage{nccmath}
		{\end{pmatrix}\end{medsize}}%

\newtheorem{theorem}{Theorem}
\newtheorem{assumption}{Assumption}
\newtheorem*{assumption*}{Assumption}

\newtheorem{corollary}{Corollary}
\newtheorem{lemma}{Lemma}

\theoremstyle{remark}
\newtheorem{example}{Example}
\newtheorem*{example*}{Example}
\newtheorem{remark}{Remark}

\counterwithin*{corollary}{section}

\newtheorem{definition}{Definition}[section]
\renewcommand{\thesection}{\arabic{section}}
\renewcommand{\theequation}{\arabic{section}.\arabic{equation}}

\renewcommand{\theassumption}{\arabic{section}.\arabic{assumption}}

\renewcommand{\thelemma}{\arabic{section}.\arabic{lemma}}

\DeclareMathOperator{\eig}{eig}

\title{Debiased Machine Learning of Set-Identified Linear Models}

\author{ Vira Semenova\thanks{Email: semenovavira@gmail.com.  First version: September, 2018 (Job Market Paper).   I am grateful to  Victor Chernozhukov, Bryan Graham, Michael Jansson, Patrick Kline,   Anna Mikusheva,  Whitney Newey, Demian Pouzo and Jim Powell for their guidance and encouragement.  I am thankful to Alberto Abadie, Chris Ackerman, Sydnee Caldwell, Xiaohong Chen, Denis Chetverikov, Ben Deaner, Mert Demirer, Jerry Hausman, Peter Hull, Tetsuya Kaji, Kevin Li, Elena Manresa, Rachael Meager, Francesca Molinari, Denis Nekipelov, Oles Shtanko, Cory Smith, Jorg Stoye, Sophie Sun, Roman Zarate,  three anonymous referees, and the participants at the MIT Econometrics Lunch for helpful comments. }   \\
University of California, Berkeley 
 }
\date{}

\begin{document}
\maketitle 
 \begin{abstract} 
    This paper provides estimation and inference methods for an identified set's boundary  (i.e., support function)  where the selection among a very large number of covariates is based on modern regularized tools. I characterize the boundary using a semiparametric moment equation. Combining Neyman-orthogonality and sample splitting ideas, I construct a root-$N$ consistent, uniformly asymptotically Gaussian estimator of the boundary and propose a multiplier bootstrap procedure to conduct inference.   I apply this result to the Partially Linear Model, the Partially Linear IV Model and the Average Partial Derivative with an interval-valued outcome.  
    \end{abstract}

 \section{Introduction and Motivation.}

Interval-valued outcomes are ubiquitous in economic research. Examples of such outcomes include bidders' valuation in English auctions (\cite{HaileTamer}), income and wages (\cite{Trostel}, \cite{Gafarov}), house prices (\cite{GRT}, \cite{BerSasaki}), and county-level employment rates (\cite{Dorn}).  An outcome is interval-valued if the actual outcome $Y$ is missing, but there exist an observable lower bound $Y_L$ and an upper bound $Y_U$ so that
\begin{align}
\label{eq:ylyu}
Y_L \leq Y \leq Y_U \text{ a.s. }
\end{align}
When an outcome is interval-valued, the parameter of interest is a set, where each point corresponds to a possible random variable $Y$ in the band  \eqref{eq:ylyu}.

The main contribution of this paper is to provide an estimator of the identified set's boundary, where the selection among high-dimensional controls is based on modern machine learning/regularized methods. The paper focuses on identified sets whose boundary can be represented by a moment equation as in \cite{BM}, \cite{BMM}.  In this paper,  the equation depends on an identified functional nuisance parameter,  for example, a conditional mean function.  A naive approach would be to plug-in a machine learning estimate of the nuisance parameter into the moment equation and solve  for the boundary. However, modern regularized methods (machine learning techniques) have bias converging slower than the parametric rate, which cannot be made small by classic  techniques (e.g., undersmoothing). As a result, plugging such estimates into the moment equation produces a biased, suboptimal estimate of the boundary itself.

To overcome the transmission of the bias into the second stage, I  adjust the moment equation to make it insensitive or, formally, Neyman-orthogonal, to the biased estimation of the nuisance parameter.  While orthogonality has been extensively studied in the point-identified case, set identification presents several challenges. The first one is the non-smoothness of $x \rightarrow \min (x,0)$ function at $x=0$, often occurred in censored LAD (e.g., \cite{Powell}). The second one is to establish uniformly valid inference over the boundary in addition the pointwise one.

When the identified set is multi-dimensional, its  boundary consists of continuum points.  As a result, economists are interested in uniform inference in addition to the pointwise inference.  Establishing uniform inference is not trivial. To control the speed at which an empirical sample average concentrates around the population mean, I invoke maximal inequalities of \cite{CCKAS} instead of Markov inequality that is typically sufficient in the point-identified case. I propose multiplier bootstrap algorithm to conduct inference.  By virtue of orthogonality, only the moment function (the second stage),  not the nuisance parameter estimate (the first stage), needs to be resampled in simulation. As a result,  this multiplier bootstrap is faster to compute than the weighted bootstrap of  \cite{CCMS}, which is based on a non-orthogonal moment and involves resampling of both stages. I demonstrate the proposed approach in a simulation exercise and give a brief empirical illustration.

\paragraph{Literature review. Set identification} This paper bridges the gap between three literatures: set-identified models,  debiased/orthogonal machine learning, and non-smooth models. Set identification is a vast area of research,  encompassing a wide variety of approaches: linear and quadratic programming, random set theory, support function, and moment inequalities
 (\cite{Manski90}, \cite{ManskiPepper}, \cite{Manski:2002}, \cite{HaileTamer}, \cite{CHT},  \cite{BM}, \cite{Molinari2008}, \cite{CilibertoTamer}, \cite{LeeBound}, \cite{Stoye}, \cite{AndrewsShiECMA}, \cite{BMM2},  \cite{CCMS},  \cite{BMM3}, \cite{CherRigStoker}, \cite{BMM}, \cite{CLR}, \cite{FanPark}, \cite{KaidoWhite},  \cite{KaidoSantos},   \cite{KaidoWhite2}, \cite{Pakesetal}, \cite{ShiShum},  \cite{Kaido:2016}, \cite{Kasy2016}, \cite{KlineTamer}, \cite{AndrewsShi}, \cite{CanayBugniShi},   \cite{Kaido}, \cite{ChenTamerChristensen}, \cite{GafarovMeierOlea}, \cite{Shi}, \cite{Gafarov}, \cite{KaidoMolinariStoye}, \cite{SyrgkanisTamer}, \cite{Torgovitsky},\cite{MolinariStoye}, \cite{BerSasaki}, \cite{Honore}, \cite{AndrewsRothPakes},  \cite{kallus2020localized}, \cite{FanTao}, \cite{FanTao2}, \cite{MolinariMolchanovPeng}, \cite{HsiehShiShum}, \cite{DongHsiehShum}), see e.g. \cite{Tamer:2010} or  \cite{Molinari:2018} for a  review. This paper generalizes the \cite{BMM}'s model by allowing its components to depend on a functional nuisance parameter, covering e.g., \cite{CCMS} and \cite{Kaido} as special cases.

\paragraph {Orthogonality. }Next, this paper contributes to a large body of work on debiased inference for parameters following regularization or model selection  (\cite{Neyman:1959}, \cite{Neyman:1979}, \cite{HardleStoker1989}, \cite{NeweyStoker},  \cite{Newey1994}, \cite{Robins}, \cite{robinson:88}, \cite{ZhangZhang}, \cite{JM}, \cite{chernozhukov2016double}, \cite{LRSP}, \cite{Program}, \cite{sasaki2018estimation}, \cite{sasaki2020unconditional}, \cite{Sasaki}, \cite{chiang2019multiway}, \cite{ning2020doubly}, \cite{chernozhukov2021debiased}, \cite{chernozhukov2021automatic}, \cite{CherSem}, \cite{NSS}, \cite{singh2020debiased}, \cite{Colangelo}, \cite{Lieli}, \cite{ZimLech}).   A basic idea is to make the moment condition insensitive, or, formally, Neyman-orthogonal, to the biased estimation of the nuisance parameter. For a semiparametric GMM setting, the work by \cite{AckerbergChen} derives an orthogonal moment condition whose nuisance functions are identified by conditional moment restriction, such as  conditional mean and conditional quantiles. Combining Neyman-orthogonality and sample splitting,  \cite{LRSP} and \cite{chernozhukov2016double}  derive a root-$N$ consistent and asymptotically normal estimator for a single target parameter. This has idea has been extended for many functional parameters in $Z$-estimation framework, in the context of   distribution regression (\cite{BelCherWei}) and quantile regression  (\cite{sasaki2020unconditional}). Next, the paper is related to literature on the non-smooth estimating equations  (\cite{Powell}, \cite{Powell2}, \cite{PowellStockStoker}, \cite{kaplan_sun_2017}, \cite{franguridi2021conditional}).   Finally, the paper contributed to a small, but growing literature on machine learning for bounds and partially identified models (\cite{kallus2019assessing},  \cite{jeong2020robust}, \cite{SemSupp2}, \cite{Bonvini_2021}).

\paragraph{Structure of the paper.} The paper is organized as follows. Section \ref{sec:setup1} demonstrates main points for the partially linear model of \cite{robinson:88}.  Section \ref{sec:theory} states theoretical results. Section \ref{sec:application}  applies the results to models with an interval-valued outcome.   Section \ref{sec:montecarlo} presents finite-sample evidence. Section \ref{sec:empirical} contains an empirical illustration. Section \ref{sec:proofs} contains the proofs of main results.

\section{Set-Up.}
\label{sec:setup1}

\subsection{ General  Framework }
\label{sec:setup}
I focus on parameters that are linear in an unobserved scalar outcome $Y$. The identified set  takes the form
\begin{align}
\label{eq:idset}
    \mathcal{B} = \{ \beta = \Sigma^{-1} \E V (\eta_0) Y,  \quad Y_L \leq Y \leq Y_U\},
\end{align}
where the random vector $V(\eta_0)  \in \mathrm{R}^{d}$ depends on a nuisance function $\eta_0$.   Examples of the nuisance functions
$$
\eta_0 = \eta_0(X)
$$
include the propensity score, the conditional density, and the regression function, among others. The matrix  $\Sigma \in \mathrm{R}^{d \times d}$ is identified by a moment equation
\begin{align}
\label{eq:sigma}
\Sigma = \E A(W,\eta_0)
\end{align}
and is assumed invertible.  The key innovation of this framework is to allow  the vector $V(\eta)$, the matrix function $A(W, \eta)$, and the  bounds $Y_L, Y_U$ to depend on a functional nuisance parameter $\eta$, covering the models in \cite{BMM}, \cite{CCMS}, \cite{Kaido}, and many others as special cases.

\subsection{Examples}

\begin{example}[Partially Linear  Model]
\label{ex:plp}
Consider the partially linear model of \cite{robinson:88}
\begin{align}
\label{eq:plm}
Y &= D' \beta_0 + f_0(X) + U, \quad \E [U\mid D,X] = 0,
\end{align}
where $D \in \mathrm{R}^d$ is a treatment  (policy) variable, $\beta_0$ is a causal (structural) parameter,  $$X = (1,X_1, X_2, \dots, X_{p_X}) $$ is a vector of covariates whose dimension may be large relative to the sample size (e.g., $p_X \gg N$), and $f_0(\cdot)$ is an integrable function. The parameter $\beta_0$ can be represented as the minimizer of the least squares criterion function
\begin{align}
\label{eq:plppointlong}
\beta_0& =\arg \min_{b \in \mathrm{R}^d, f \in L_2(P)} \E (Y - D' b - f(X))^2.
\end{align}
As in Lemma \ref{lem:fromlongtoshort}, $\beta_0$ coincides with the minimizer of a shorter criterion function
\begin{align}
\label{eq:plppoint1}
\beta_0& =\arg \min_{b \in \mathrm{R}^d} \E (Y - (D - \eta_0(X))' b)^2.
\end{align}
Thus, the  identified set $\mathcal{B}$  is a special case of model \eqref{eq:idset}-\eqref{eq:sigma} with  $V(\eta)$ and $A(W,\eta)$ defined as follows. The treatment regression function is
\begin{align}
\label{eq:cexp}
\eta_0(X) = \E[ D \mid X],
\end{align}
 the treatment residual is
\begin{align}
\label{eq:resid}
V (\eta) = D-\eta(X),
\end{align}
the matrix  function is
\begin{align}
\label{eq:aweta}
A(W,\eta) =  (D - \eta(X))(D - \eta(X))'.
\end{align}

\end{example}

\begin{example}[Partially Linear IV Model]
\label{ex:plpiv}
Consider the following partially linear IV model
\begin{align}
\label{eq:plpiv}
Y &= D' \beta_0 + f_0(X) + U, \quad \E[ U \mid Z, X]=0 \\
D &= m_0(X) + E, \quad \E [ E \mid X ] =0, \label{eq:m0} \\
Z &= \eta_0(X) + V, \quad \E[ V \mid X ] =0, \label{eq:eta0}
\end{align}
where $Z \in \mathrm{R}^d$ is the instrument for $D$.  The parameter $\beta_0$ can be characterized by a moment equation
\begin{align*}
\E (Z - \eta_0(X)) ( Y - (D - m_0(X))' \beta_0)  = 0,
\end{align*}
which gives a closed-form expression for $\beta_0$
\begin{align*}
\beta_0 = (\E (Z - \eta_0(X)) (D - m_0(X))' )^{-1} \E (Z - \eta_0(X)) Y.
\end{align*} 
 The model is a special case of \eqref{eq:idset}-\eqref{eq:sigma} with 
\begin{align}
V (\eta) &= Z - \eta(X) \label{eq:plpivspec} \\
A (W, \eta,m) &=  (Z - \eta(X)) (D - m(X))', \label{eq:plpivspec2}
\end{align}
where $V = V(\eta_0)$ in \eqref{eq:eta0}.  If $Z = D$, the model \eqref{eq:plpiv}-\eqref{eq:eta0} coincides with \eqref{eq:plm}-\eqref{eq:cexp}.

\end{example}

\begin{example}[Average Partial Derivative]
\label{ex:apd} An important parameter in economics is the average partial derivative. This parameter shows the average effect of a small change in a  variable of interest $D$ on the outcome $Y$ conditional on the covariates $X$. To describe this change, define the conditional expectation function of an outcome $Y$ given the variable $D$ and exogenous variable $X$ as $$\mu(D,X):= \E [Y|D,X]$$ and its partial derivative with respect to $D$ as $\nabla_D\mu(D,X):=\nabla_D\mu(d,X)|_{d=D}$. Then, the average partial derivative is defined as
\begin{align}
\label{eq:apd}
    \beta &= \E \nabla_D\mu(D,X).
\end{align}
For example, when $Y$ is the logarithm of consumption, $D$ is the logarithm of price, and $X$ is the vector of other demand attributes,  the average partial derivative  stands for the average price elasticity. 

Assume that the  variable $D$ has bounded support $\mathcal{D} \subset \mathcal{R}^d$. Furthermore, the conditional density $f ( D \mid X)$  has positive density on this support a.s. in $X$. \cite{HardleStoker1989} have shown that the average partial derivative  can be represented as $\beta =   \E V Y,$
where $V = - \nabla_D \log f(D|X) = - \frac{\nabla_D f(D|X)}{f(D|X)}$ is the negative partial derivative of the logarithm of the density $f(D|X)$. When $Y$ is interval-valued,  the identified set $\mathcal{B}$ for $\beta$ is a special case of \eqref{eq:idset}-\eqref{eq:sigma} with $A(W, \eta) = \Sigma = I_d$, the nuisance function $\eta_0(D, X) = \frac{\nabla_D f(D|X)}{f(D|X)}$ and the vector $V ( \eta) = -\eta$. \cite{Kaido} studies a special case of this problem without covariates.

\end{example}

\subsection{Single treatment}
\label{sec:singletreat}

In this section, I derive an orthogonal moment equation for the upper bound $\beta_U$ on the causal parameter $\beta_0$ in Example \ref{ex:plp}. Because the bias due to sign mistake in $Y^{\text{best}}(\eta) \neq Y^{\text{best}}(\eta_0)$  proves to be second-order, I derive the orthogonal moment treating $Y^{\text{best}}(\eta_0)$ as observed. Next, I formally control the bias due to the sign mistake. 
The following two subsections elaborate on these points.

\paragraph{Moment equation for $\beta_U$. }Consider Example \ref{ex:plp} with a single treatment (i.e., $d=1$).    The identified set $\mathcal{B}$ becomes a closed interval $[\beta_L, \beta_U]$.  The upper bound  $\beta_U$ is
\begin{align}
\label{eq:betu1d}
\beta_U =   \max_{\{ Y: Y_L \leq Y \leq Y_U \}}  \bigg\{ \dfrac{\E  (D - \eta_0(X)) \cdot Y }{\E (D - \eta_0(X))^2 } \bigg\}.
\end{align}
To maximize the numerator of \eqref{eq:betu1d},  take $Y=Y_U$ for positive values of $D - \eta_0(X)$ and $Y=Y_L$ otherwise. Define the best-case outcome
\begin{align}
\label{eq:ubg}
	Y^{\text{best}} (\eta)  &= \begin{cases} Y_L, \quad  D - \eta(X) \leq 0, \\
	Y_U, \quad  D - \eta(X) > 0.
	\end{cases}
\end{align}
Plugging \eqref{eq:ubg} into \eqref{eq:betu1d} gives the moment function for $\beta_U$
 \begin{align}
 \label{eq:naive:1d}
  m (W, \beta_U, \eta) :=  (Y^{\text{best}}(\eta) -   (D- \eta(X)) \beta_U) (D -\eta(X)).
 \end{align}

  \paragraph{ Establishing orthogonality. }  In what follows,  $\eta$ corresponds to an instance of the nuisance parameter whose true value is $\eta_0$.    Consider an infeasible moment function
  \begin{align}
 \label{eq:naive:1d0}
   m_0(W, \beta_U, \pmb{\eta} ) :=  (Y^{\text{best}}(\eta_0) -   (D- \pmb{\eta}  (X)) \beta_U) (D - \pmb{\eta} (X)),
 \end{align}
 where  the best-case outcome $Y^{\text{best}}(\eta_0) $ is treated as if it was observed, and $ \pmb{\eta}$  indicates the instance of the nuisance parameter treated as unknown. The moment equation \eqref{eq:naive:1d0} is not orthogonal to the perturbations of $\widehat{\eta}(X) - \eta_0(X)$
 \begin{align*}
 \partial_{r} \E [ m_0 (W,  \beta_U, r(\eta - \eta_0) + \eta_0) ] |_{r=0} = -\E[ Y^{\text{best}}(\eta_0) ( \eta(X) - \eta_0(X)) ] \neq 0.
 \end{align*}
 Therefore, the bias of the estimation error $\widehat{\eta}(X)-\eta_0(X)$ translates into the moment \eqref{eq:naive:1d0}.  To overcome the transmission of this bias,  \cite{robinson:88} proposes an orthogonal moment equation
 \begin{align*}
 	 g_0 (W, \beta_U, \{ \pmb{\eta}, \pmb{\gamma}_U \})= (Y^{\text{best}}(\eta_0)- \pmb{\gamma}_U(X) -  (D- \pmb{\eta}  (X)) \beta_U) (D - \pmb{\eta} (X)),
 \end{align*}
 where the true value of $\gamma_U(x)$ is  $$\gamma_{U,0}(X) =  \E[ Y^{\text{best}}(\eta_0) \mid X]. $$  The orthogonality condition is
  \begin{align}
  \label{eq:derivative}
 &\partial_{r} \E [ g_0 (W, \beta_U, r(\eta - \eta_0) + \eta_0, \gamma_{U,0} ) ] |_{r=0}  \\
 &=- \E[ (Y^{\text{best}}(\eta_0) - \E[ Y^{\text{best}}(\eta_0)\mid X]) ( \eta(X) - \eta_0(X)) ] =0. \nonumber 
 \end{align} 
   Thus, the bias of the estimation error, $\widehat{\eta}(X) - \eta_0(X)$, does not translate into the moment  \eqref{eq:naive:1d0}.    The orthogonality condition with respect to $\gamma_{U,0}$ can be verified in a similar way.

  \paragraph{ Bias due to sign mistake. }  I now discuss whether the mistake in the best-case outcome  $Y^{\text{best}} (\eta) \neq  Y^{\text{best}} (\eta_0)$, which has been so far ignored, has any effect on the support function estimate. Consider the difference between the feasible moment $g (W, \beta_U, \{ \pmb{\eta},  \pmb{\gamma}_U \})$ and its infeasible analog 
  \begin{align*}
& g (W, \beta_U, \{ \pmb{\eta},  \pmb{\gamma}_U \})  -  g_0 (W, \beta_U,  \{ \pmb{\eta},  \pmb{\gamma}_U \}) =  (D - \eta(X)) ( Y^{\text{best}}(\eta) - Y^{\text{best}}(\eta_0) ) \\
 &= (D - \eta_0(X)) ( Y^{\text{best}}(\eta) - Y^{\text{best}}(\eta_0) ) \\
 &+ ( \eta_0(X) - \eta(X)) ( Y^{\text{best}}(\eta) - Y^{\text{best}}(\eta_0) ). 
  \end{align*}
Define the first-order bias $B_1( \eta, \eta_0)$ as
   \begin{align*}
 B_1( \eta, \eta_0):= \E [ (D - \eta_0(X)) ( Y^{\text{best}}(\eta) - Y^{\text{best}}(\eta_0) )]
 \end{align*}
 and the second-order one
  \begin{align*}
 B_2( \eta, \eta_0):= \E [ ( \eta_0(X) - \eta(X)) ( Y^{\text{best}}(\eta) - Y^{\text{best}}(\eta_0) )].
  \end{align*}
  Below, I  describe the conditions under which $B_1( \eta, \eta_0)$ and $ B_2( \eta, \eta_0)$ are negligible,  (i.e., $o(N^{-1/2})$). If they hold, the feasible moment equation $g (W, \beta_U, \{ \pmb{\eta},  \pmb{\gamma}_U \})$ is insensitive to the biased estimation of $\eta$. As a result, the support function estimator based on $g (W, \beta_U, \{ \pmb{\eta},  \pmb{\gamma}_U \})$ is asymptotically unbiased under plausible conditions.

 Consider the   best-case outcome $Y^{\text{best}} (\eta)$
$$
Y^{\text{best}} (\eta) = Y_L + (Y_U - Y_L) 1 \{ D- \eta(X) \geq 0 \}.
$$
 The sign mistake $Y^{\text{best}} (\eta) \neq  Y^{\text{best}} (\eta_0)$ occurs on the events
  \begin{align}
 \label{eq:e-}
 \mathcal{E}_{-} :&= \bigg\{  D - \eta_0(X) < 0 < D - \eta(X) \bigg\}, \\
   \label{eq:e+}  \mathcal{E}_{+} :&= \bigg\{  D - \eta(X) < 0 < D - \eta_0(X) \bigg\}.
 \end{align}
On these events, the residual cannot exceed estimation error in absolute value
  \begin{align}
  \label{eq:small}
  \bigg\{  \mathcal{E}_{-}  \cup \mathcal{E}_{+}  \bigg\} &\Rightarrow  \bigg\{ 0 < | D -\eta_0(X) | < | \eta(X) - \eta_0(X) | \bigg\}.
\end{align}
 If the width  $Y_U - Y_L$ is bounded by $M_{UL}$,  the  estimation error of $Y^{\text{best}}(\eta) $ is bounded as
  \begin{align}
  \label{eq:bestcaseerror}
   &|Y^{\text{best}}(\eta) - Y^{\text{best}}(\eta_0) |=(Y_U - Y_L) 1 \{  \mathcal{E}_{-}  \cup \mathcal{E}_{+} \} \\
   &\leq M_{UL} 1 \{ 0 < | D -\eta_0(X) | < | \eta(X) - \eta_0(X) | \}. \nonumber
  \end{align}
  Suppose  the conditional density $  h_{V (\eta_0) \mid X} (t,X) $ of $V(\eta_0)$ is bounded by $M_h$ a.s.. Invoking \eqref{eq:small} gives
\begin{align}
 \E | \eta (X) - \eta_0(X) | 1{\{\mathcal{E}_{+ } \cup  \mathcal{E}_{-}\}} &\leq \E_{X} \int_{- |  \eta (X) - \eta_0(X) | }^{|  \eta (X) - \eta_0(X) |} t h_{ V (\eta_0) | X} (t) dt  \nonumber \\
 &\leq  2  M_h \E_{X}  ( \eta (X) - \eta_0(X))^2. \label{eq:mainbound2}
\end{align}
As a result, the bias terms $B_1( \eta, \eta_0)$ and $B_2( \eta, \eta_0)$ shrink at  the quadratic rate.  Combining \eqref{eq:mainbound2} and \eqref{eq:derivative} gives a feasible moment function
    \begin{align}
    \label{eq:feasible2}
 	 g (W, \{ \pmb{\eta}, \pmb{\gamma}_U \})&=  (Y^{\text{best}}(\pmb{\eta}) - \pmb{\gamma}_U (X) -   (D- \pmb{\eta}(X)) \beta_U) (D -\pmb{\eta}(X)).
	  \end{align}

\subsection{Multi-dimensional case}
\label{sec:multid}

In this section, I derive an orthogonal moment for the support function, starting from a non-orthogonal one due to \cite{BMM}, \cite{BM}. 

 \paragraph{Moment Equation for Support Function. }   As shown in \cite{BM}, the identified set $\mathcal{B}$ in \eqref{eq:idset} is a compact and convex set. Thus, it can be described  by its projections onto a unit sphere 
\begin{align}
\label{eq:unitsphere}
\mathcal{S}^{d-1} := \{q \in \mathrm{R}^{d}, \quad \| q\| = 1\}.
\end{align}
For any direction $q \in \mathcal{S}^{d-1}$, define the support function as the upper bound on $q' \beta_0$ 
 \begin{align}
\label{eq:suppfun}
    \sigma(q):= \sup_{b \in \mathcal{B}} q' b.
\end{align}
As proposed in  \cite{BM} and \cite{BMM}, define the projected weighting vector 
\begin{align}
\label{eq:zp}
z(p,\eta) = p'  V (\eta), 
\end{align}  
 the best-case outcome $Y(p,\eta)$   
\begin{align}
\label{eq:yq}
   Y (p,\eta)     &= Y_L + (Y_U - Y_L) 1\{  z(p,\eta)>0\},
\end{align}
and the projection parameter $p(q)$ 
\begin{align}
\label{eq:pq}
p(q) = \Sigma^{-1} q.
\end{align}
  Then, the moment equation for $\sigma(q)$ is
\begin{align}
\label{eq:zqwq}
    \sigma(q) &= \E [z (p, \eta_0) Y (p,\eta_0)] \big|_{p = p(q)}.
\end{align}

\paragraph{Orthogonal Moment for Support Function. } The moment equation \eqref{eq:zqwq} is sensitive to the biased estimation of the nuisance parameter $\eta$. To avoid the transmission of this bias into the second stage, I construct another moment function $g(W, p, \xi),$ where $\xi(p)$ is the  nuisance parameter, in the following steps.
\begin{enumerate}

\item Starting from an infeasible, smooth moment
\begin{align}
\label{eq:smooth:d0}
m_0(W, p, \pmb{\eta}) := z (p,\pmb{\eta}) Y (p,\eta_0)
\end{align}
derive an infeasible orthogonal moment $g_0(W, p, \xi(p)) $ obeying \eqref{eq:derivative} for each $p$. 

\item Invoke Lemma \ref{lem:powell} to bound the bias 
\begin{align}
\label{eq:ubias}
\sup_{p \in \mathcal{P}} \E | z(p,\eta) \left(Y (p,\eta)- Y (p,\eta_0) \right) |  = O( \E \| \eta(X) - \eta_0(X) \|^2),
\end{align}
where $p$ belongs to a compact bounded set $\mathcal{P}$ defined below. When $\Sigma = I_d$, $\mathcal{P} = \mathcal{S}^{d-1}$.

\item Combine \eqref{eq:ubias} and  \eqref{eq:smooth:d0} to obtain the feasible orthogonal moment
\begin{align}
\label{eq:feasible}
g(W, p, \xi(p)) = g_0(W, p, \xi(p)) + z(p,\eta) (Y (p,\eta)- Y (p,\eta_0)).
\end{align}

\end{enumerate}

\begin{example*}[Example \ref{ex:plpiv}, cont.]
Consider Example \ref{ex:plpiv}. The projected weighting vector \eqref{eq:zp} is
\begin{align*}
z(p, \eta) = p' (Z - \eta(X)).
\end{align*}
The infeasible orthogonal moment $g_0(W, p, \xi(p)) $ is
\begin{align}
\label{eq:orthomom1}
g_0(W, p,  \xi(p)) =  z(p,\eta) (Y (p,\eta_0)  - \gamma (p, X)),
\end{align}
where  
\begin{align}
\label{eq:rieszgamma}
\gamma_0(p,x) &= \E [ Y (p,\eta_0) \mid X] \\
&=  \E [ Y_L \mid X=x] + \E [ (Y_U - Y_L) 1\{ p' V(\eta_0) > 0 \} \mid X=x] \nonumber
\end{align}
The nuisance function $\xi_0(p) = (\eta(\cdot), \gamma(p,\cdot))$. Invoking \eqref{eq:feasible} gives a feasible orthogonal moment
\begin{align}
\label{eq:orthomom}
g(W, p,  \xi(p)) =  z(p,\eta) (Y (p,\eta)  - \gamma (p, X)).
\end{align}

Corollary \ref{cor:plpiv} establishes the asymptotic theory for the support function estimator based on \eqref{eq:orthomom}. 

\end{example*}

\begin{example*}[Example \ref{ex:apd}, cont.]
Consider Example \ref{ex:apd}. The projected weighting vector is
\begin{align*}
z(q, \eta) = - q' \partial_D \log f(D \mid X).
\end{align*}
The infeasible orthogonal moment $g_0(W, p, \xi(p)) $ is
\begin{align}
     \label{eq:rho:apd}
        g_0(W,q,\xi(q)) = z(q, \eta)  Y(q, \eta_0) + q' \partial_D \log f(D \mid X)  \mu(q,D,X) + q' \nabla_D \mu(q,D,X),
\end{align}
where 
\begin{align*}
 \mu_0(q,D,X)  &= \E [ Y_L \mid D, X] + \E [ (Y_U - Y_L)  \mid D,X ] 1\{ -q' \partial_D \log f(D \mid X) > 0 \} \\
 &= \gamma_{L,0}(D, X) +  \gamma_{UL,0}(D, X) 1\{ -q' \partial_D \log f(D \mid X) > 0 \}.
 \end{align*}
  Invoking \eqref{eq:feasible} gives a feasible moment equation
\begin{align}
\label{eq:orthomom:apd}
g(W, p,  \xi(p)) &= g_0(W,q,\xi(q)) + z(q,\eta) (Y (q,\eta)  -Y (q,\eta_0)) \\
&= z(q,\eta) Y(q, \eta) +q' \partial_D \log f(D|X)  \mu(q,D,X) + q' \nabla_D \mu(D,X) \nonumber.
\end{align}
In absence of the conditioning covariates, \eqref{eq:orthomom:apd} coincides with the efficient score in  \cite{Kaido}.  Corollary \ref{cor:apd} establishes the asymptotic theory for the support function estimator based on \eqref{eq:rho:apd}. 

\end{example*}

\subsection{ Overview of Main Results}
\label{sec:overview}

The Support Function  Estimator $\widehat{\sigma}(q)$ has two stages.  In the first stage, I construct an estimate $\widehat{\xi}$ of the nuisance parameter $\xi_0$ using some regularized estimator. In the second stage, I compute the estimated  values $(\widehat{\xi}_i)_{i=1}^N$ and the support function estimate. I use different samples  in the first and the second stage in the form of cross-fitting. The number $K$ of cross-fit partitions is assumed to be fixed/finite relative to $N$.

 \begin{definition}[Cross-Fitting]
 \label{sampling}
\mbox{}
 	\begin{compactenum} 
	\item  For a random sample of size $N$, denote a $K$-fold random partition of the sample indices $[N]=\{1,2,...,N\}$ by $(J_k)_{k=1}^K$, where $K$ is  the number of partitions and the sample size of each fold is $n = N/K$. For each $k \in [K] = \{1,2,...,K\}$ define $J_k^c = \{1,2,...,N\} \setminus J_k$.
	\item For each $k \in [K]$, construct estimates $\widehat{\xi}_k = \widehat{\xi}( W_{i \in J_k^c})$ and $\widehat{\eta}_k = \widehat{\eta}( W_{i \in J_k^c})$    of the nuisance parameters $\xi_0$ and $\eta_0$ using only the data $\{ W_{j}: j \in J_k^c \}$. For any observation $i \in J_k$, define $\widehat{\xi}_i = \widehat{\xi}_k (W_i)$ and $\widehat{\eta}_i = \widehat{\eta}_k (W_{i})$.
		\end{compactenum}
\end{definition}

Definition \ref{sampling} introduces cross-fitting.  Cross-fitting plays an essential role in modern debiased inference in semi-parametric models; see, e.g., \cite{bch:2010,zheng:laan,chernozhukov2016double} for recent examples and \cite{hasminskii:debiased} and \cite{schick1986asymptotically} for early, classical uses of simpler sample-splitting methods for debiased inference. Cross-fitting is proposed for the cases where the nuisance parameter $\xi_0(\cdot)$ does not depend on $p \in \mathcal{P}$, such as Example \ref{ex:apd} and Examples \ref{ex:plp}--\ref{ex:plpiv}  under Assumption \ref{ass:suffcond}. 

\begin{definition}[Support Function Estimator]
\label{def:estimate:psi}
Let $\widehat{\xi}$ and $\widehat{\eta}$ be the estimates of $\xi_0$ and $\eta_0$. Define 
\begin{align}
\widehat{\Sigma} &= \dfrac{1}{N} \sum_{i=1}^N A(W_i, \widehat{\eta}_i), \quad \widehat{p}(q) = (\widehat{\Sigma}^{-1})' q    \label{sigma:est} \\
    \widehat{\sigma}(q)&= \dfrac{1}{N} \sum_{i=1}^N g(W_i, \widehat{p}(q) , \widehat{\xi}_i( \widehat{p}(q) )).    \label{eq:psi:est}
\end{align}

\end{definition}

\begin{definition}[Multiplier Bootstrap]
\label{def:bb}
Let  $(e_i)_{i=1}^N: e_i $ are  i.i.d. truncated exponential random variables $\text{Tr Exp}(1)$ on $[0, \bar{M}]$ independent of the data. Define the bootstrap analog of $ \widehat{\sigma}(q)$ as
\begin{align}
\widetilde{\Sigma} &= \dfrac{1}{N} \sum_{i=1}^N \dfrac{e_i}{\bar{e}} A(W_i, \widehat{\eta}_i), \quad \widetilde{p}(q) = (\widetilde{\Sigma}^{-1})' q    \label{sigma:est:boot} \\
    \widetilde{\sigma}(q)&= \dfrac{1}{N} \sum_{i=1}^N \dfrac{e_i}{\bar{e}} g(W_i, \widetilde{p}(q) , \widehat{\xi}_i( \widetilde{p}(q) )).    \label{eq:psi:boot}
\end{align}
\end{definition} 

Under mild conditions on $\xi$, the Support Function Estimator delivers a high-quality estimate $\widehat{\sigma}(q)$ of the support function $\sigma(q)$ with the following properties
\begin{enumerate}

\item With probability (w.p.) $\rightarrow 1$, the estimator converges uniformly over the unit sphere $\mathcal{S}^{d-1}$
\begin{align}
\label{eq:urate}
\sup_{ q \in \mathcal{S}^{d-1} } | \widehat{\sigma} (q) - \sigma(q) |  = O_P(1/\sqrt{N}) = o_P(1).  
\end{align}

\item The estimator $\widehat{\sigma}(q)$ is asymptotically Gaussian
\begin{align}
\label{eq:limit}
S_N(q) :=\sqrt{N} (\widehat{\sigma}(q) - \sigma(q)) = \G_N(q)+ o_P(1) \quad \text{ uniformly in } \mathcal{S}^{d-1},
\end{align}
where the empirical process $\G_N(q)$ is approximated by a Gaussian process $\G(q)$, which is a tight $P$-Brownian bridge in $ \ell^{\infty}(\mathcal{S}^{d-1})$.

\end{enumerate}

Define  the bootstrap statistic
\begin{align*}
\widetilde{S}_N(q):=\sqrt{N} (\widetilde{\sigma}(q) - \widehat{\sigma}(q)) 
\end{align*}

\paragraph{Pointwise asymptotics. } The sharp identified set for $q' \beta_0$ is $[ - \sigma(-q), \sigma(q)]$. Its $(1-\tau)$-pointwise confidence region (CR)  is 
\begin{align*}
[\underline{i}(q), \bar{i}(q)]:=[- \widehat{\sigma}(-q) + N^{-1/2} \widehat{C}_{\tau/2}(q), \quad  \widehat{\sigma}(q) +N^{-1/2} \widehat{C}_{1-\tau/2}(q) ],
\end{align*}
where the critical values  $\widehat{C}_{\tau/2} (q)$ and $\widehat{C}_{1-\tau/2} (q) $  are the $\tau/2$ and $1-\tau/2$ quantiles of the bootstrapped statistic $|\widetilde{S}_N(q)|$. Plugging $$q= \mathbf{e}_k = (0,0,\dots, 0, \underbrace{1}_{k}, 0, \dots, 0) \in \mathrm{R}^d, \quad k = 1,2, \dots, d$$ gives the CR for the projection of the identified set $\mathcal{B}$.

\paragraph{Uniform asymptotics. }  The $(1-\tau)$-uniform confidence region (CR) for  $[ - \sigma(-q), \sigma(q)]$ is 
\begin{align}
\label{eq:uniform}
[\underline{i}_u(q), \bar{i}_u(q)]:=[- \widehat{\sigma}(-q) + N^{-1/2} \widehat{C}^{*}_{\tau/2}, \quad  \widehat{\sigma}(q) +N^{-1/2} \widehat{C}^{*}_{1-\tau/2} ],
\end{align}
where $\widehat{C}^{*}_{\tau/2}$ and $\widehat{C}^{*}_{1-\tau/2}$ are the quantiles of the bootstrapped statistic $\sup_{q \in \mathcal{S}^{d-1}} |\widetilde{S}_N(q)|$.  Likewise, for any function $f(\cdot)$ and a critical value  $\widehat{c}_N = c_N + o_P(1)$ and $c_N = O_P(1)$,
\begin{align*}
\Pr ( f(S_N) \leq \widehat{c}_N) - \Pr^e ( f(\widetilde{S}_N) \leq \widehat{c}_N) \rightarrow_P 0,
\end{align*}
where $\Pr^{e} (\cdot)$ is the probability conditional on the data. 

\section{ Theoretical Results.}
\label{sec:theory}

\setcounter{example}{0}

\paragraph{Notation.} I use the empirical process notation. For a generic function $f$ and a generic sample $(W_i)_{i=1}^N$, denote the empirical sample average by $$ \EN f(W_i) := \dfrac{1}{N}  \sum_{i=1}^N f(W_i) $$ and the
scaled, demeaned sample average by  $$\GN f(W_i) := 1/\sqrt{N} \sum_{i=1}^N  [f(W_i) - \int f(w) d P(w)].$$ 
For two sequences of random variables  $\{ a_N, b_N, N \geq 1\}: a_N \lesssim_{P}  b_N$ means    $a_N = O_{P} (b_N)$. For two sequences of numbers $\{a_N, b_N, N \geq 1\}$, $a_N \lesssim  b_N$ means $a_N = O (b_N)$. Let $a \wedge b = \min \{ a, b\}, a \vee b = \max \{ a, b\} $. The $\ell_2$ norm of a vector is denoted by $\| \cdot \|$, the $\ell_1$ norm is denoted by $\| \cdot \|_1$, the $\ell_{\infty}$ norm is denoted by $\| \cdot \|_{\infty}$, and $\ell_0$ norm is denoted by $\| \cdot \|_{0}$. For a matrix $Q$, let $\|Q\|$ be the maximal eigenvalue of $Q$ and $\| Q \|_F$ be the Frobenius norm of $Q$. For a random vector $W$, let $\| W \|_{P,c}:= (\int |W|^c d P)^{1/c}$. The random sample $(W_i)_{i=1}^N$ is a sequence of independent copies of a random element $W$ taking values in a measurable space $(\mathcal{W}, \mathcal{A}_{\mathcal{W}})$  according to a probability law $P$. The $\| f \|_{P_N, 2}$ is the empirical $\ell_2$-norm, denoted as $\| f \|_{P_N, 2}:=(N^{-1} \sum_{i=1}^N f^2(W_i))^{1/2}$. Define the projection set 
\begin{align}
\label{eq:pee}
\mathcal{P} = \bigg\{ p \in \mathrm{R}^d: \quad 1/2 \min \eig (\Sigma^{-1})  \leq \| p \| \leq 2 \max \eig (\Sigma^{-1})  \bigg\}
\end{align}
and let $C_P:= 2 \max \eig (\Sigma^{-1}) $.  Let $\mathcal{F}_c$ be the space of continuous functions obeying two conditions: (1) $f(Z)$ has a continuous distribution when $Z$ is a tight Gaussian process  with non-degenerate covariance function and (b) $f(\xi_N + c) -f(\xi_N) = o(1)$ for any $c=o(1)$ and any $\| \xi_N \|= O_P(1)$. Note that $[p_X]:=\{1,2,\dots, p_X\}$.

\subsection{Assumptions}

Assumption \ref{ass:sigma} is a standard identification condition. It ensures that the eigenvalues of $\Sigma$ are bounded from above and below.
\begin{assumption}[Identification]
\label{ass:sigma}
There exist  constants $\lambda_{\min} >0$ and $\lambda_{\max} < \infty $ that bound the eigenvalues of $\Sigma$ in \eqref{eq:sigma} from above and below $0 < \lambda_{\min} \leq \min \eig (\Sigma) \leq \max \eig(\Sigma) \leq \lambda_{\max}$.
\end{assumption}

Assumption \ref{ass:jacobian} ensures that the support function is differentiable on the unit sphere.    It requires the  distribution of the weighting vector $V(\eta_0)$ to be sufficiently smooth. For example, if the vector $V(\eta_0)$ has a symmetric (i.e., spherical) distribution around zero,  the normalized vector $\| V(\eta_0) \|^{-1} V(\eta_0)$ is uniformly distributed  on the surface of the unit sphere $\mathcal{S}^{d-1}$, and Assumption \ref{ass:jacobian} holds. Assumption \ref{ass:jacobian} is a common regularity  condition  in set-identified models (e.g., Condition C.1 in \cite{CCMS}) and censored median regression (e.g., Assumption R.2 in \cite{Powell}).  

\begin{assumption}[Smooth boundary]
\label{ass:jacobian}
 Let $d \geq 2$. There exists a finite constant $\mathcal{C}_V$ such that
\begin{align}
\label{eq:smooth}
\sup_{q \in \mathrm{S}^{d-1}} \Pr \left( \dfrac{ |q' \Sigma^{-1/2} V (\eta_0) | }{ \| \Sigma^{-1/2} V (\eta_0)  \| } \leq \delta \right) \leq \mathcal{C}_V \delta.
\end{align}
\end{assumption}

\begin{example}[Gaussian Noise]
\label{ex:gaussian}
Consider Example \ref{ex:plpiv} with $$V(\eta_0) = Z - \eta_0(X) \sim N(0, \Sigma)$$ independent of $X$. Then,  $\Sigma^{-1/2} V(\eta_0) \sim N(0, I_d)$ is standard Gaussian vector and 
$\Sigma^{-1/2} V(\eta_0)/ \| \Sigma^{-1/2} V (\eta_0)  \|$ is uniformly distributed over the unit sphere $\mathcal{S}^{d-1}$. For any $q \in \mathcal{S}^{d-1}$, $|q'\Sigma^{-1/2} V(\eta_0)|/ \| \Sigma^{-1/2} V (\eta_0)  \|$ is uniformly distributed on $[0,1]$ (\cite{Pitman}).  As a result,  \eqref{eq:smooth} holds with $\mathcal{C}_V = 1$ conditional on $X$ uniformly in $X$. 
\end{example}

Assumption \ref{ass:jacobian} is a sufficient condition for the smoothness of the boundary. If it holds, the moment equation \eqref{eq:orthomom} is differentiable in $p$.  The gradient
\begin{align}
\label{eq:gp}
G(p):=  \E  V (\eta_0)  Y(p, \eta_0)
\end{align}
 is a uniformly continuous function of $p$ (see Lemma \ref{lem:uniderivative} in Online Appendix).  As a result, there exists a uniform Gaussian approximation for the support function estimator.  
 
 \begin{remark}
 \label{rm:discrete}
  Consider Example \ref{ex:plp}. If $D$ and $X$ consist of discrete variables only, the distribution of $V (\eta_0)$ cannot be continuous, and  Assumption \ref{ass:jacobian}  fails. Discrete distributions imply flat surfaces on the identified set, which may not be compatible with uniform Gaussian approximation. In this case, 
   \cite{CCMS} suggests adding a small amount of continuously distributed noise and work with a slightly expanded set with smooth boundary,  while \cite{Gafarov} provides an alternative approach. 
 \end{remark}

The moment functions $A(W, \eta)$ in  \eqref{eq:sigma} and $g(W, p, \xi(p))$ depend on the nuisance parameters $\eta_0$ and $\xi_0$, respectively.  Definition \ref{def:nearorthog}  introduces a sequence of nuisance realization sets $\Xi_N \subseteq \Xi$ and $\mathcal{T}_N \subseteq \mathcal{T}$ that contain $\xi_0$ and $\eta_0$, as well as their estimators $\widehat{\xi}$ and $\widehat{\eta}$, with probability $1-\phi_N$.  As the sample size increases, the sets $\Xi_N$ and $\mathcal{T}_N$ shrink. The shrinkage speed is measured by the rates below. 

\begin{definition}[Moment Rates]
\label{def:nearorthog}
Let $\{ \Xi_N, N \geq 1\}$ and $\{ \mathcal{T}_N, N \geq 1\}$ be sequences of subsets of  $\Xi$ and  $\mathcal{T}$, respectively, obeying the following conditions. (1)  The true values $\xi_0$ and $\eta_0$ belong to $\Xi_N$ and $\mathcal{T}_N$ for all $N \geq 1$. There exists a sequence of numbers $\phi_N = o(1)$ such that  the first-stage estimators $\widehat{\xi}$ of $\xi_0$ and $\widehat{\eta}$ of $\eta_0$ belong to $\Xi_N$ and $\mathcal{T}_N$ with probability at least $1-\phi_N$. Define the sequences $\mu_N, r_N'', r_N', A_N, \delta_N$ as
\begin{align}
\sup_{\xi \in \Xi_N} \sup_{p \in \mathcal{P}} | \E [ g(W, p, \xi(p)) - g (W,p, \xi_0(p)) ] | = \mu_N  \label{eq:mun} \\
\sup_{\eta \in \mathcal{T}_N} \| \E [ A (W, \eta) - A(W, \eta_0) ] \| = A_N \label{eq:an} \\
	\sup_{\xi \in \Xi_N} \sup_{p \in \mathcal{P}} (\E  (g(W, p, \xi(p)) - g (W,p, \xi_0(p)) )^2 )^{1/2} = r_N''  \label{eq:rnprimeprime} \\
	 \sup_{p, p_0 \in \mathcal{P} \| p - p_0 \| \lesssim \tau_N}  (\E ( g(W, p, \xi_0(p)) - g(W,p_0, \xi_0(p_0)) )^2 )^{1/2} = r_N'   \label{eq:rnprime}  \\
     \sup_{\eta \in \mathcal{T}_N}  (\E  \| A(W, \eta) - A(W, \eta_0) \|^2)^{1/2} = \delta_N,  \label{eq:delta} 
	\end{align}
	where $\tau_N:= N^{-1/2} \log N$ for the case when $A(W,\eta_0)$ is being estimated and $\tau_N:= r_N':=0$ when $A(W, \eta_0) = \Sigma$ is known.
\end{definition}

\begin{assumption}[Regularity Conditions]
\label{ass:ratebound2}
(1) There exist absolute constants $c'>2$ and $\bar B_A< \infty$ such that  the following matrix norms are bounded
$$\sup_{\eta \in \mathcal{T}_N} \E \| A(W, \eta)  \|^{c'} \leq \bar B_A, \quad \E \| A(W, \eta_0)  \|_F^{2} \leq \bar B_A$$ (2) There exists a sequence $v_N  = o(N^{-1/4})$ such that $$ \| \E_N A(W_i, \eta_0) - \Sigma \| = O_P (v_N) = o_P(1). $$ 
\end{assumption}

Assumption \ref{ass:ratebound} requires the rates of Definition \ref{def:nearorthog} to decay sufficiently fast. The matrix moment function $A(W, \eta)$ is assumed to be already orthogonal with respect to $\eta$. This is the case  for covariance matrices in Examples \ref{ex:plp} and \ref{ex:plpiv}.

\begin{assumption}[Convergence Rates]
\label{ass:ratebound}
(1) For $c_3>2$ in Assumption \ref{ass:ratebound2}, suppose the numerator moment rates $\mu_N,  r_N'', r_N'$ obey the following bounds: $\mu_N = o(N^{-1/2})$ and  $$(r_N'' + r_N' ) \log^{1/2} (1/(r_N'' + r_N')) + N^{-1/2+1/c_3} \log N = o(1).$$ 
(2)  The matrix rates $A_N$ and $\delta_N$ obey the following bounds:  $A_N = o(N^{-1/2})$ and $\delta_N = o(1)$. 
\end{assumption}

Assumption \ref{ass:concentration:chap1} bounds the complexity of the function class $$\mathcal{G}_{\xi} = \{ g (W,p, \xi(p)), p \in \mathcal{P}\}.$$

\begin{assumption}[Complexity Conditions]
\label{ass:concentration:chap1}
(1) There exists  a  measurable envelope function  $G_{\xi} = G_{\xi}(W)$ that almost surely bounds all elements in the class $$ \sup_{p \in \mathcal{P}} | g(W,p,\xi(p)) | \leq G_{\xi}(W)
\quad \text{a.s.}$$
 There exists $c>2$ such that $\|G_{\xi}\|_{P,c} := \left(\int_{w \in \mathcal{W}}  (G_{\xi}(w))^c \right)^{1/c} < \infty$. (2) There exist constants $a,v$ that do not depend on $N$ such that the uniform covering entropy of the function class $\mathcal{G}_{\xi}$ is  bounded  
 \begin{align}
 \label{eq:complexityq}
  \log \sup_{Q} N(\epsilon \| G_{\xi} \|_{Q,2}, \mathcal{G}_{\xi} , \| \cdot \|_{Q,2}) \leq  v \log (a/\epsilon), \quad \text{ for all } 0 < \epsilon \leq 1.
  \end{align}
\end{assumption}

\subsection{Results}

Define the influence function $h_g(W,q)$ as
\begin{align}
\label{eq:hwgq}
h_g(W, q) &= g(W, p(q), \xi_0(p(q))) - \sigma(q) 
\end{align}
and the influence function for the matrix estimation
\begin{align}
\label{eq:hwaq}
h_A(W, q) &=  -G(p(q))' \Sigma^{-1} ( A(W, \eta_0) - \Sigma) \Sigma^{-1} q  ,
\end{align}
where $G(p)$ is the gradient  defined in \eqref{eq:gp}. Finally, define
\begin{align}
\label{eq:hwq}
h(W,q) = h_g(W, q)+h_A(W, q).
\end{align}

\begin{theorem}[Limit Theory for the Support Function Process]
\label{thm:limit}
Suppose Assumptions  \ref{ass:sigma}-\ref{ass:concentration:chap1}  hold. Then, the support function process $S_N(q) = \sqrt{N} (\widehat{\sigma} (q) - \sigma(q))$ is asymptotically linear uniformly on $ \mathcal{S}^{d-1}$
$$S_N(q) = \G_N [h(W,q)] + o_{P} (1)\text{ uniformly on }  \mathcal{S}^{d-1},$$
where $h(W, q)$ is as in \eqref{eq:hwq}. Furthermore,   the process $S_N(q)$ admits the following approximation
  $$
S_N(q) =_d \G[h(q) ] + o_P(1) \quad \text{ in } \ell^{\infty} (\mathcal{S}^{d-1}),$$
where the process $ \G[h(q)] $ is a tight $P$-Brownian bridge in $ \ell^{\infty} (\mathcal{S}^{d-1})$ with a non-degenerate covariance function
\begin{align*}
    \Omega(q_1,q_2 ) = \E [h(W,q_1)h(W,q_2 )] - \E[h(W,q_1)] \E[h(W,q_2 )], \quad q_1, q_2  \in \mathcal{S}^{d-1}.
\end{align*}
\end{theorem}

Theorem \ref{thm:limit} is my first main result. It says that the Support Function Estimator is asymptotically equivalent to a tight Gaussian process with a non-degenerate covariance function. Previous work (e.g., \cite{CCMS}, \cite{Kaido}) has derived similar Gaussian approximations for the estimators based on classic nonparametric methods. Combing Neyman-orthogonality and sample splitting,  Theorem \ref{thm:limit} allows to accommodate both classic nonparametric and modern regularized/machine learning estimators. Corollary \ref{cor:limit} states that the inference properties of  support function estimator.

\begin{corollary}[Limit Inference on Support Function Process]
\label{cor:limit}
Suppose Assumptions  \ref{ass:sigma}--\ref{ass:concentration:chap1}  hold.  For any $f \in \mathcal{F}_c$, $\widehat{c}_N = c_N + o_P(1)$ and $c_N = O_P(1)$, 
\begin{align*}
\Pr ( f(S_N) \leq \widehat{c}_N) - \Pr ( f( \G[h(q)]) \leq \widehat{c}_N) \rightarrow 0. 
\end{align*}
If $c_N (1-\tau)$ is the $(1-\tau)$-quantile of $ f( \G[h(q)]) $ and $\widehat{c}_N(1-\tau) = c_N(1-\tau) + o_P(1)$  is any consistent estimate of this quantile, then 
 \begin{align*}
\Pr ( f(S_N) \leq \widehat{c}_N(1-\tau)) \rightarrow 1-\tau.
\end{align*}
\end{corollary}

\begin{theorem}[Limit Theory for the Bootstrap Support Function Process]
\label{thm:bb}
Under conditions of Theorem \ref{thm:limit},  the bootstrap support function process $\widetilde{S}_N(q) = \sqrt{N} (\widetilde{\sigma} (q) - \widehat{\sigma}(q))$ is asymptotically linear uniformly on $ \mathcal{S}^{d-1}$
\begin{align*}
\widetilde{S}_N(q) &=  \G_N [ (e-1) h (W,q)]+ o_{P}(1) \text{ uniformly on }  \mathcal{S}^{d-1}.
\end{align*}
Furthermore,  the  bootstrap support function process admits an approximation conditional on the data:
$$
\widetilde{S}_N(q) = \widetilde{\G[h(q)]} + o_{P^e}(1) \quad \text{ in } \ell^{\infty} (\mathcal{S}^{d-1}),  \text{ in probability } P,$$
where $\widetilde{\G[h(q)]}$ is a tight $P$-Brownian bridge in $\ell^{\infty} (\mathcal{S}^{d-1})$ with the same distribution as the process $\G[h(q)]$  defined in Theorem \ref{thm:limit}, and independent of $ \G[h(q)] $.  

\end{theorem}

Theorem \ref{thm:bb} is my second main result.  It establishes the validity of multiplier bootstrap for uniform inference on the support function.  In contrast to the weighted bootstrap of \cite{CCMS}, the multiplier bootstrap does not require re-estimating the first-stage nuisance parameter in each bootstrap repetition. Instead, the nuisance parameter is estimated  on an auxiliary sample once and plugged into the bootstrap sampling procedure.

\begin{corollary}[Limit Inference on Bootstrap Support Function Process]
\label{cor:bb}
For any $\widehat{c}_N = c_N + o_P(1)$ and $c_N = O_P(1)$, 
\begin{align*}
\Pr ( f(S_N) \leq \widehat{c}_N) - \Pr^e ( f(\widetilde{S}_N) \leq \widehat{c}_N) \rightarrow_P 0. 
\end{align*}
If $\widehat{c}_N(1-\tau)$ is the $(1-\tau)$-quantile of $ f(\widetilde{S}_N) $ under $P^e$, then 
 \begin{align*}
\Pr (f(\widetilde{S}_N) \leq \widehat{c}_N(1-\tau)) \rightarrow_P 1-\tau.
\end{align*}

\end{corollary}

\section{Partially Linear IV Model}
\label{sec:application}
\setcounter{assumption}{0}
\setcounter{lemma}{0}
\setcounter{corollary}{0}

\subsection{The First Stage }

In this section, I give  examples of the first-stage estimators as well as the nuisance realization sets. I focus on the partially linear IV model of Example \ref{ex:plpiv}. This discussion automatically covers  Example \ref{ex:plp} that is a special case of Example \ref{ex:plpiv} with $D=Z$.

Definition \ref{def:fsrate} introduces  sequences of nuisance realization sets $\{ \mathcal{T}_N, N \geq 1\}$  and  $\{ \mathcal{M}_N, N \geq 1\}$ that contain the true value of $\eta_0$ and $m_0$ and their estimates $\widehat \eta$ and $\widehat m$ with probability $1-o(1)$. As the sample size $N$ increases, the sets $\mathcal{T}_N$ and $\mathcal{M}_N$ 
shrink. The shrinkage speed is measured by mean square rates $\eta_N$ and $m_N$, respectively. 

\begin{definition}[Mean Square  Rates]
\label{def:fsrate}  Define the mean square rate for the expectation functions $\eta_0(x)$ and $m_0(x)$
\begin{align*}
  \sup_{\eta \in \mathcal{T}_N}  \left(\E \|\eta (X)- \eta_0(X)\|^2\right)^{1/2} = \eta_N \\
   \sup_{m \in \mathcal{M}_N}  \left(\E \| m (X)-m_0(X)\|^2\right)^{1/2} = m_N \end{align*}
\end{definition}

Definition \ref{def:fsrate} introduces mean square convergence rates for the expectation functions $\eta_0$ and $\gamma_0$. The bounds on $\eta_N$   are established for 
a wide variety of regularized estimators, including  partition estimators (\cite{CattaneoFarrell}, \cite{CattaneoFarrellFeng}),   $\ell_2$-boosting (\cite{Luo}), deep neural networks (\cite{Schmidt}, \cite{Farrell}), linear and nonlinear sieve estimators \cite{Chen2007}, penalized sieve estimators \cite{Chen2011},  random forest in small (\cite{WagerWalther}) and high (\cite{syrgkanis2020estimation}) dimensions with sparsity structure.   For the sake of brevity, the examples of nuisance realization sets  $ \mathcal{M}_N$ and $\mathcal{T}_N$ are omitted in this text, but they can be found in \cite{SemCher}, Appendix B or \cite{CGST}, Section 5.

\paragraph{First-Stage Fitted Values: Basic Case.}  In this paragraph, I consider the case when the nuisance parameter $\xi_0$ does not depend on $p$.    
\begin{assumption}[Independent and Symmetric Residual]
\label{ass:suffcond}
The following conditions hold. (1) The interval width $Y_U - Y_L$ is independent of $V(\eta_0)$ conditional on $X$:
\begin{align}
\label{eq:indep}
(Y_U - Y_L) \quad \indep \quad V(\eta_0) \mid X.
\end{align}
(2) The vector $V(\eta_0)$ is independent of $X$, for any $p \in \mathcal{P}$. (3) $V(\eta_0)$ has a spherical distribution, which implies $\Pr (p' V(\eta_0) >0) = 1/2$ for any $p \in \mathcal{P}$.
\end{assumption}

Suppose Assumption \ref{ass:suffcond} holds. Define
$$
\gamma_{L,0}(X):= \E[ Y_L \mid X], \quad  \gamma_{UL,0} (X)  = \E[ (Y_U - Y_L) \mid X]. 
$$
Then, the Riesz representer function takes the form
\begin{align}
 \gamma_0(p,X) &= \gamma_{L,0} (X)+1/2 \gamma_{UL,0} (X)  \label{eq:midpoint}
\end{align}
and the first-stage fitted values take the form
\begin{align}
\label{eq:crossfitgammal}
\widehat {\gamma} (X_i):= \widehat{\gamma}_L (X_i) +  1/2 \widehat{\gamma}_{UL} (X_i), \quad i=1,2,\dots, N,
\end{align}
where $(\widehat{\gamma}_L (X_i), \widehat{\gamma}_{UL}(X_i))_{i=1}^N$ are the cross-fit first-stage estimates of Definition \ref{sampling}.

\paragraph{First-Stage Fitted Values: High-Dimensional Sparse Case.} In this paragraph, I sketch a possible estimator of the Riesz representer function without relying on Assumption \ref{ass:suffcond}. Define
$$
\rho_0(p,X) = \E[ \mathcal{Y} (p,\eta_0) \mid X] = \E[ (Y_U - Y_L) 1\{ p' V(\eta_0) > 0 \} \mid X]. 
$$
Assume that the expectation function is approximated as 
\begin{align}
\label{eq:maindecomp}
\rho_0(p,X) = \Lambda(Z(X)' \nu_0(p)) + R_p(\eta_0,X), 
\end{align}
where $Z: \mathcal{X} \rightarrow \mathrm{R}^{p_X}$ is a set of $p_X$ measurable basis functions of the covariates $X$, $\nu_0(p)$ is a $p_X$-dimensional vector, and $R_p(\eta_0,X)$ is an approximation error, and $\Lambda: \mathrm{R} \rightarrow \mathrm{R}$ is the linear link function (Example \ref{ex:linearlasso}) and logistic link function (Example \ref{ex:lassologistic}). 

\begin{example}[Linear Lasso Estimator]
\label{ex:linearlasso}
Let $$\mathcal{Y}_i(p, \widehat \eta):= (Y_{U,i} - Y_{L,i}) 1 \{ p' V_i(\widehat \eta) >0 \}, \quad i=1,2,\dots, N$$ be an estimate of $\mathcal{Y}_i(p, \eta_0)$. Given the penalty level 
$$
\lambda_{Z} = \dfrac{1.1}{N^{1/2}} \Phi^{-1} \left(1 - \dfrac{\bar{\gamma}}{ 2 p_X N^d} \right), \quad \bar{\gamma} = .1/ \log N
 $$
and  the  diagonal matrix of penalty loadings $\widehat \Psi_{p}$,  define  
$$
\widehat \nu (p) \in \arg \min_{\nu \in \mathrm{R}^{p_X}} N^{-1} \sum_{i=1}^N ( \mathcal{Y}_i(p, \widehat \eta)- Z(X_i)' \nu )^2 + \lambda_{Z} \| \widehat \Psi_{p} \nu \|_1
$$
and the  fitted value as
$$
\widehat \rho(p,X_i):= Z(X_i)' \widehat \nu (p).
$$
Given the cross-fit estimate $ \widehat \gamma_L (\cdot)$ of the $\gamma_{L,0}(\cdot)$, define
\begin{align}
\label{eq:finalfitted}
\widehat \gamma(p, X_i) = \widehat \gamma_L (X_i) +Z(X_i)' \widehat \nu (p).
\end{align}
\end{example}
An important special case occurs when the brackets have constant width 
$$
Y_U - Y_L = \Delta \quad \text{a.s.},
$$
and the nuisance parameter reduces to the conditional probability
$$
 \E [ \mathcal{M} (p, \eta_0) \mid X]:=\Pr (p' V(\eta_0) > 0 \mid X).
$$
Suppose this probability can be approximated as 
\begin{align}
\label{eq:maindecomplogistic}
\Pr (p' V(\eta_0) > 0 \mid X) = \Lambda (Z(X)' \nu_0(p)) + R_p(\eta_0,X), 
\end{align}
where  $ \Lambda (t) = \exp t/ (\exp t+1)$  is the logistic link function.

\begin{example}[Logistic Lasso Estimator]
\label{ex:lassologistic}
Let $$L(y,t):= -1 (1\{ y=1 \} \log \Lambda(t) + 1\{ y=0 \} \log (1-\Lambda(t)) )$$ be the logistic loss function and let $\mathcal{M}_i(p, \widehat \eta) = 1\{ p' V_i (\widehat \eta)>0\}$ be the estimated outcome.
Given the penalty level $\lambda_Z$ and  the  diagonal matrix of penalty loadings $\widehat \Psi_{p}$, define
$$
\widehat \nu (p) \in \arg \min_{\nu \in \mathrm{R}^{p_X}} N^{-1} \sum_{i=1}^N L (\mathcal{M}_i(p, \widehat \eta) , Z(X_i)' \nu ) + \lambda_{Z} \| \widehat \Psi_{p} \nu \|_1
$$
The final estimator of  fitted values is
\begin{align}
\label{eq:finalfittedlogis}  \widehat \gamma(p, X_i) &= \widehat \gamma_L (X_i) + \widehat \rho(p,X_i)=\widehat \gamma_L (X_i) + \Delta  \Lambda( Z(X_i)' \widehat \nu(p)).
\end{align}

\end{example}

Lemma \ref{lem:riesz} provides the first-stage convergence rates for linear Lasso estimator. In contrast to regular linear Lasso model, the outcome $\mathcal{Y}(p, \eta_0)$ depends on the nuisance parameter $\eta_0$ and therefore has to be estimated. I show that the linear Lasso estimator remains valid as long as the first-stage mean square rate decays sufficiently fast.

\begin{lemma}[Validity of Linear Lasso with Estimated Outcome]
\label{lem:riesz}
The following conditions hold for $N$ large enough and a sequence $\zeta_N=o(1)$ and $\Lambda(t)=t$. (i) The model \eqref{eq:maindecomp} is approximately sparse with $s_{\nu} =s_{\nu}(N)$
\begin{align*}
\sup_{ p \in \mathcal{P}} \| \nu_0(p) \|_0 \leq s_N
\end{align*}
and $\log (p_X \vee N) \leq \zeta_N N^{1/3}$. (ii) Heteroscedasticity. There exists a constant $c_{\zeta} >0$ so that $0<c_{\zeta} \leq \E[ \zeta^2(p, \eta_0)  \mid X ]  \text{ a.s. }$. (iii) Lipschitz property of $\nu_0(p)$. For some finite constant $C_L, \sup_{p_1, p_2 \in \mathcal{P}} \| \nu_0(p_1) - \nu_0(p_2) \|_1 \leq  C_L \| p_1 - p_2 \|$. (iv) The approximation error decays fast:
$$ \sup_{p \in \mathcal{P}} (\E R^2_0(p,X))^{1/2} = o (\bar{\sigma}_N), \quad  \bar{\sigma}_N:=\sqrt{s_N \log (p_X \vee N)/N}.$$
(vi)  First Stage.  (a) The first-stage mean square rate is fast enough  $\eta_N = o (\bar{\sigma}_N)$ and (b) There exists $\eta^{\infty}_N = o(1)$ so that
$\sup_{\eta \in \mathcal{T}_N} \sup_{x \in \mathcal{X}} \| \eta(x) - \eta_0(x) \| \leq \eta^{\infty}_N = o(1)$. Then, under additional  conditions on $Z(X)$ in Assumption \ref{lem:rieszappendix1},
the estimate $\widehat \nu (p)$ of Example \ref{ex:linearlasso} is uniformly sparse, that is $\sup_{p \in \mathcal{P}} \| \widehat \eta(p) \|_0 \leq C_X s_N$, 
 and the following performance bounds hold:
\begin{align}
\sup_{ p \in \mathcal{P}}  \| Z(X)' (\widehat \nu (p) - \eta_0(p)) \|_{P_N, 2} &\leq C_X \sqrt{\dfrac{s_N \log p_X }{N}} \label{eq:fsstage} \\
 \sup_{ p \in \mathcal{P}}  \| \widehat \nu (p) - \eta_0(p) \|_1 &\leq C_X \sqrt{\dfrac{s^2_N \log p_X }{N}}.\label{eq:fsstage2}
\end{align}

\end{lemma}

\begin{lemma}[Validity of Logistic Lasso with Estimated Outcome]
\label{lem:rieszlogistic}
Suppose the conditions of Lemma \ref{lem:riesz} hold for  \eqref{eq:maindecomp} with logistic link function $\Lambda(t)$. Then, under  Assumption \ref{lem:rieszappendix2},
the estimate $\widehat \nu (p)$ of Example \ref{ex:lassologistic} is uniformly sparse, that is $\sup_{p \in \mathcal{P}} \| \widehat \eta(p) \|_0 \leq \widetilde{C} s_N$, and the bounds \eqref{eq:fsstage}--\eqref{eq:fsstage2} hold.

 \end{lemma}

\subsection{The Second Stage }

In this section, I describe the  Support Function Estimator for the Partially Linear IV Model. The estimator for Example \ref{ex:plp} is obtained by replacing Steps 1 and 2 and 4 of Algorithm \ref{alg:plpiv} by their analogs in Algorithm \ref{alg:plp}. As an input, Algorithm \ref{alg:plp} takes a direction $q \in \mathcal{S}^{d-1}$ and the first-stage fitted  values $(\widehat{\eta}(X_i), \widehat{m}(X_i), \widehat{\gamma}(p,X_i))_{i=1}^N$.

 \begin{algorithm}[H]
\begin{algorithmic}[1]
\STATE The  treatment residual $$\widehat{V}_i:= Z_i - \widehat{\eta}(X_i),  \quad \widehat{E}_i:= D_i - \widehat{m}(X_i) \qquad i=1,2,\dots, N.$$
\STATE The sample covariance matrix: $$\widehat{\Sigma}:= \dfrac{1}{N} \sum_{i=1}^N \widehat{V}_i \widehat{E}_i' .$$
\STATE The best-case outcome as a function of $p$   $$Y_{i}(p, \widehat{\eta}) :=Y_{L,i} + (Y_{U,i}-Y_{L,i})1\{ p' \widehat{V}_i>0  \},   \quad i=1,2,\dots, N. $$
\STATE The IV coefficient of the second-stage residual $Y_{i}(\widehat{p}(q), \widehat{\eta}) - \widehat{\gamma}(\widehat{p}(q), X_i)$  on the instrument residual $\widehat{V}_i$
\begin{align}
\label{eq:betaq}
\widehat{\beta}_q =  \widehat{\Sigma}^{-1} \dfrac{1}{N} \sum_{i=1}^N\widehat{V}_i [ Y_{i}(\widehat{p}(q) , \widehat{\eta}) - \widehat{\gamma} (\widehat{p}(q) ,X_i)], \quad \widehat{p}(q) =\widehat{\Sigma}^{-1}q
\end{align}
\STATE Report: the projection of  $\widehat{\beta}_q$ onto the direction $q$: $$\widehat{\sigma}(q) = q' \widehat{\beta}_q.$$
\end{algorithmic}
\caption{Support Function Estimator for Partially Linear IV Model.}
\label{alg:plpiv}
\end{algorithm}

 \begin{algorithm}[H]
Input: a direction $q \in \mathcal{S}^{d-1}$, estimated values $(\widehat{\eta}(X_i), \widehat{\gamma}(p,X_i))_{i=1}^N$. Estimate the following quantities:
\begin{itemize}[1]
\item[$1'$, $2'$] The  treatment residual and the sample covariance matrix $$\widehat{V}_i:= D_i - \widehat{\eta}(X_i),   \qquad i=1,2,\dots, N, \quad \widehat{\Sigma}:= \dfrac{1}{N} \sum_{i=1}^N \widehat{V}_i V_i' .$$
\item[$4'$] The OLS coefficient of the second-stage residual $Y_{i}(\widehat{p}(q), \widehat{\eta}) - \widehat{\gamma}(\widehat{p}(q), X_i)$  on the treatment residual $\widehat{V}_i$ as in \eqref{eq:betaq}.
\end{itemize}
\caption{Support Function Estimator for Partially Linear Model.}
\label{alg:plp}
\end{algorithm}

\subsection{Results }
\label{sec:ver}
 In this section, I verify  Assumptions \ref{ass:ratebound}-- \ref{ass:concentration:chap1} for Example \ref{ex:plpiv}. Then, I establish asymptotic Gaussian approximation for the Support Function Estimator.

\begin{assumption}[Bounded Width $Y_U - Y_L$]
\label{ass:boundedwidth}
The width $Y_U - Y_L$ is bounded by a finite constant $M_{UL}$ a.s., that is 
$$
Y_U - Y_L \leq M_{UL} \text{ a.s. }
$$
\end{assumption}

\begin{assumption}[Regularity Conditions]
\label{ass:regularity}
(1)  For every $p \in \mathcal{P}$,   the residual vector $V(\eta_0) $ has  a conditional density $h_{\{p' V(\eta_0) \mid X=x\}}(\cdot,x)$ that is bounded uniformly over $X$ by $M_h$.  (2)  The residual vector norms $\| V(\eta_0) \| =\| Z-\eta_0(X)\|$ and $ \| E(m_0) \|= \|D-m_0(X)\|$ are subGaussian random variables conditional on $X$. (3) The random variable $Y_L$ has finite conditional second moment $\sup_{x \in \mathcal{X}} \E[ Y^2_L \mid X=x] \leq  \bar C_L$ and $\| Y_L \|_{P,4} \leq \bar C_L$ for some finite $ \bar C_L< \infty$. Finally, for some $c'>2$, $\| \| V(\eta_0) \| (Y_L - \gamma_{L,0} (X))\|_{P,c'}$ and $\| V (\eta_0) \|_{P,c'}$ are finite.  

\end{assumption}

Lemma \ref{lem:powell} shows that the moment equation \eqref{eq:zqwq} incurs only a second-order bias due to the sign mistake of $V(\eta_0)$ as long as $V(\eta_0)$ is continuously distributed.
\begin{lemma}[First-Order  Bias]
\label{lem:powell}
Under Assumptions \ref{ass:boundedwidth} and \ref{ass:regularity} (1),   the  first-order bias   shrinks at the quadratic speed
\begin{align*}
\sup_{p \in \mathcal{P}} | \E [ z(p, \eta_0) ( Y(p, \eta) - Y(p, \eta_0) )]  | \leq 2 C_P^2 M_{UL} M_h \E \| \eta(X) - \eta_0(X) \|^2.
\end{align*}
Likewise, the second-order bias shrinks at the quadratic speed
\begin{align*}
\sup_{p \in \mathcal{P}} | \E [ (z(p, \eta) - z(p, \eta_0)) ( Y(p, \eta) - Y(p, \eta_0) )]  |  \leq 2 C_P^2 M_{UL} M_h \E \| \eta(X) - \eta_0(X) \|^2.
\end{align*}
\end{lemma}

\begin{lemma}[Verification of Assumption \ref{ass:ratebound2}]
\label{cor:powell2}
Suppose Assumption \ref{ass:regularity}(2) holds. For  any sequence $\ell_N \rightarrow \infty$, Assumption \ref{ass:ratebound2}  is satisfied with $v_N = \sqrt{ \ell_N /N}$, in particular, one can take  $\ell_N = \log N$ to ensure that $v_N = o(N^{-1/2} \log N)$. 
\end{lemma}

Lemma \ref{cor:powell} verifies the Assumption \ref{ass:ratebound}  for Example \ref{ex:plpiv}.  Define the mean square rates for the expectation functions
\begin{align*}
  \sup_{\gamma_L \in \Gamma_{L,N}}  \left(\E (\gamma_{L} (D,X) - \gamma_{L,0}(D,X))^2 \right)^{1/2} =: \gamma_{L,N} \\
 \sup_{\gamma_{UL} \in \Gamma_{UL,N}}  \left(\E (\gamma_{UL} (D,X) - \gamma_{UL,0}(D,X))^2 \right)^{1/2} =: \gamma_{UL,N} 
\end{align*}

\begin{lemma}[Verification of Assumptions \ref{ass:ratebound}--\ref{ass:ratebound2}]
\label{cor:powell}
Let $\eta_0(X)$ be as in \eqref{eq:eta0}, $V(\eta)$ be as in \eqref{eq:plpivspec}, the matrix function  $A(W, \eta, m)$ be as in \eqref{eq:plpivspec2} and the  orthogonal moment function be as in \eqref{eq:orthomom}.   Suppose Assumptions \ref{ass:boundedwidth} and  \ref{ass:regularity} and \ref{ass:suffcond} hold.  Furthermore, suppose (1) the elements of $\Gamma_{L,N}$ are bounded by finite constant $M_{\gamma}$ in the sup-norm: $\sup_{ \gamma_L \in \Gamma_{L,N} } \sup_{x \in \mathcal{X}} | \gamma_L(x) | \leq M_{\gamma}$ and $\sup_{ \gamma_{UL} \in \Gamma_{UL,N} } \sup_{x \in \mathcal{X}} | \gamma_{UL}(x) | \leq M_{UL\gamma}$   (2) the elements of $\mathcal{M}_N$ and $\mathcal{T}_N$ are bounded by finite constant $M_{\eta}< \infty$ in the sup-norm: $$\sup_{ m \in \mathcal{M}_N }  \sup_{x \in \mathcal{X}} \| m(x) \| \leq M_{\eta} , \quad \sup_{ \eta \in \mathcal{T}_N }  \sup_{x \in \mathcal{X}} \| \eta(x) \| \leq M_{\eta}. $$ Then, Assumption \ref{ass:ratebound2}(1) holds. Furthermore,  the bias rates in Definition \ref{def:nearorthog} can be bounded as follows with $\gamma^2_N := 2(\gamma^2_{L,N} + \gamma^2_{UL, N})$
\begin{align*}
\mu_N &\leq 4 C^2_P M_{UL} M_h \eta^2_N + C_P \eta_N \cdot \gamma_N \\
A_N &\leq \eta_N \cdot m_N.
\end{align*}
In addition,  for $\tau_N := N^{-1/2} \log N$, the sequences $r_N'', r_N', \delta_N$ obey
$$
r_N'' = O (\eta_N + \gamma_N), \quad r_N' = O ( N^{-1/4} \log^{1/2} N), \quad \delta_N = O (\eta_N + m_N). 
$$
Thus, if  $\eta_N = o(N^{-1/4})$ and $ (m_N + \gamma_N) \cdot \eta_N = o(N^{-1/2})$ and $m_N = o(1)$ and $\gamma_N = o(1)$,  Assumption \ref{ass:ratebound} holds. 

\end{lemma}

Combining the statements in Lemmas \ref{cor:powell2}--\ref{cor:powell}, I obtain the following corollary. 

\begin{corollary}[Asymptotic Theory for Partially Linear IV Model with Interval-Valued Outcome]
\label{cor:plpiv}
Suppose Assumptions \ref{ass:sigma}, \ref{ass:jacobian} and  \ref{ass:suffcond} and the conditions of Lemma \ref{cor:powell} hold with the fast enough first-stage rates $\eta_N, m_N$ and $\gamma_N:= 2(\gamma_{L,N} + \gamma_{UL, N})$ such that Assumption \ref{ass:ratebound} holds.  Then,  Theorems \ref{thm:limit} and \ref{thm:bb} and Corollaries \ref{cor:limit} and \ref{cor:bb} hold for the Support Function Estimator of the Algorithm \ref{alg:plpiv} with the first-stage fitted values \eqref{eq:crossfitgammal} and   with the  influence function $h(W,q)$  equal to \eqref{eq:hwq}. 
\end{corollary}

\begin{corollary}[Asymptotic Theory for Partially Linear IV Model with Interval-Valued Outcome]
\label{cor:plpiv1}
Suppose Assumptions \ref{ass:sigma}, \ref{ass:jacobian}  and the conditions of Lemma \ref{lem:riesz} and  Lemma \ref{cor:powell} hold with the fast enough first-stage rates $\eta_N, m_N$ and $\gamma_N:= \gamma_{L,N}$  such that Assumption \ref{ass:ratebound} holds.   Then, Theorems \ref{thm:limit} and \ref{thm:bb} and Corollaries \ref{cor:limit} and \ref{cor:bb} hold for the Support Function Estimator of the Algorithm \ref{alg:plpiv} with the  influence function $h(W,q)$  equal to \eqref{eq:hwq}, where the first-stage fitted values are as in  \eqref{eq:finalfitted} (assuming Lemma \ref{lem:riesz} holds with $\bar{\sigma}_N = o(N^{-1/4})$ )  or as in \eqref{eq:finalfittedlogis} (assuming Lemma \ref{lem:rieszlogistic} holds with $\bar{\sigma}_N = o(N^{-1/4})$).
\end{corollary}

\begin{remark}
\label{rm:sharp}
Consider Example \ref{ex:plp}. Note that the short least squares regression \eqref{eq:plppoint1} uses only $d$ out of (infinitely) many restrictions implied by the exogeneity restriction \eqref{ex:plp}.  Therefore, the identified set
$$
\mathcal{B}_1:=\bigg\{ b \in \mathrm{R}^d: \exists f_b \in \mathcal{L} \text{ and } Y \in [Y_L, Y_U]:  \quad \E[ Y - D b - f_b(X) \mid D, X] =0 \bigg\} 
$$
is a convex subset of $\mathcal{B}$, but may not coincide with $\mathcal{B}$. 
\end{remark}

\section{Average Partial Derivative}
\label{sec:apd}

In this section, I present the identification, estimation and inference results for Example \ref{ex:apd}.

\setcounter{assumption}{0}
\setcounter{lemma}{0}
\setcounter{corollary}{0}
\setcounter{example}{0}
\setcounter{definition}{0}

Assumption \ref{ass:regcond:apd} states the sufficient conditions for compactness and convexity of the identified set $\mathcal{B}$.

\begin{assumption}[Regularity Conditions for Average Partial Derivative]
\label{ass:regcond:apd}
The following conditions hold. (1) There exists a compact and convex set $\mathcal{D}$ with nonempty interior containing the support of $D$, such that $f_0 (d \mid X) = 0$ on the boundary of $\mathcal{D}$ a.s. in $X$. (2) The random variables $D$ and $f_0(D \mid X)$ and $\nabla f_0 (D \mid X)$ have the density conditional on $X$. (3) The random vector $$\eta_0(D,X) = \nabla f_0 (D \mid X)/ f_0 (D \mid X)$$ is $L_{P,2}$-integrable, that is, there exists a finite constant $C_{\text{APD}}< \infty$ such that 
$$
\| \| V(\eta_0) \| \|_{P,2} := \| \| \nabla f_0 (D \mid X)/ f_0 (D \mid X) \| \|_{P,2} \leq C_{\text{APD}}.
$$
(4)  The functions $\gamma_{L,0}(D,X)$ and $\gamma_{UL,0}(D, X)$ are bounded by some finite constant $C_{\gamma}< \infty$ on the support of $D$ and $X$. (5) Furthermore, $\gamma_{L,0}(D,X)$ and $\gamma_{UL,0}(D, X)$ are continuously differentiable w.r.t to $D$  with bounded derivatives a.s. in $D$ and $X$. (5) The random variables $\| V(\eta_0) \|$ and  $
\| V(\eta_0) \| (Y_L - \gamma_{L,0}(D,X))
$ are $L_{P,2}$-integrable and $L_{P,c'}$-integrable for $c'>2$.

\end{assumption}

\begin{lemma}
Suppose Assumption  \ref{ass:regcond:apd} (1)-(4) hold. Then, (a) the identified set $\mathcal{B}$ for $\beta_0$ is compact and convex and (b) the support function of  $\mathcal{B}$ is given by \eqref{eq:zqwq}. 
In addition, if Assumption  \ref{ass:jacobian} holds, which implies that $\mathcal{B}$ is strictly convex.
\label{lem:apd}
\end{lemma}
Lemma \ref{lem:apd} extends the Theorem 2.1 of \cite{Kaido} to allow the density $D$ to be conditioned on  $X$. 

\begin{assumption}[Margin Condition]
\label{ass:margin}
(1) There exists an absolute constant  $\bar C_f <\infty$ so that, in some neighborhood $(0, \bar t)$ of zero, 
\begin{align*}
\sup_{q \in \mathcal{S}^{d-1}} \Pr \left( |q'  \nabla_D f_0(D \mid X)/ f_0 (D \mid X) | \leq \delta \right) \leq \bar C_f \delta ,  \quad \delta \in (0, \bar{t}).
\end{align*}
\end{assumption}

The margin condition is commonly used in classification analysis (\cite{MammenTsybakov}, \cite{Tsybakov}) and empirical welfare maximization (\cite{KitagawaTetenov}, \cite{MbakopTabord}) and bounds \cite{SemSupp2}. This paper is the first one to introduce it for the study of average partial derivative with an interval-valued variable. 

\begin{definition}[Worst-Case Rates]
Let $f_0(D \mid X)$ and $\nabla_D f_0(D \mid X)$ be the conditional density and its derivative. Let $\{ F_N, \quad N \geq 1 \}$ and $\{ F^1_N, \quad N \geq 1 \}$ be sequences of realization sets of the estimates of $f_0 (D \mid X)$ and  $\nabla_D f_0 (D \mid X)$, respectively. Assume that these sequences shrink at the following worst-case rates $f^{\infty}_N$ and $f^{\infty}_{1,N}$
\begin{align*}
\sup_{ f \in F_N } \sup_{d, x}   | f (d \mid x) - f_0( d \mid x) |  \leq f^{\infty}_N \\
\sup_{ f \in F^1_N } \sup_{d, x}   \| \nabla_{d} f (d \mid x) - \nabla_d f_0( d \mid x) \|  \leq f^{\infty}_{1,N}
\end{align*}
Define
$$
\eta^{\infty}_N:= f^{\infty}_{1,N} + f^{\infty}_N.
$$
Furthermore, assume that the elements of $F_N$ are bounded by some $\bar B_f< \infty$:
$$
\sup_{ f \in F_N } \sup_{d, x}  \{ | f^{-1} (d \mid x ) |,  | f (d \mid x ) |\} \leq \bar B_f.
$$
and
$$
\sup_{ f \in F^1_N } \sup_{d, x}   \| \nabla_{d} f (d \mid x) \| \leq \bar B_f.
$$

\end{definition}

\subsection{The Algorithm}

 \begin{algorithm}[H]
\begin{algorithmic}[1]
\STATE The best-case outcome as a function of $q$   $$Y_{i}(q, \widehat{\eta}) :=Y_{L,i} + (Y_{U,i}-Y_{L,i})1\{ -q' \widehat{\eta}(D_i, X_i)>0  \},   \quad i=1,2,\dots, N. $$
\STATE The expectation function of $Y(q, \eta)$ and its derivative
\begin{align*}
\widehat \mu (q, D_i, X_i) = \widehat{ \gamma}_L (D_i, X_i) + \widehat{ \gamma}_{UL} (D_i, X_i)  1\{ -q' \widehat{\eta}(D_i, X_i)>0  \} \\
\nabla_D \widehat \mu (q, D_i, X_i) = \nabla_D \widehat{ \gamma}_L (D_i, X_i) + \nabla_D \widehat{ \gamma}_{UL} (D_i, X_i) 1\{ -q' \widehat{\eta}(D_i, X_i)>0  \} 
\end{align*}
\STATE The sample moment
\begin{align}
\label{eq:betaqapd}
\widehat{\beta}_q =   \dfrac{1}{N} \sum_{i=1}^N -\widehat{\eta}(D_i, X_i) [ Y_{i}(q, \widehat{\eta}) - \widehat{\mu} (q,D_i, X_i)] + \nabla_D \widehat{\mu} (q,D_i, X_i) \end{align}
\STATE Report: the projection of  $\widehat{\beta}_q$ onto the direction $q$: $$\widehat{\sigma}(q) = q' \widehat{\beta}_q.$$
\end{algorithmic}
\caption{Support Function Estimator for Average Partial Derivative.}
\label{alg:apd}
\end{algorithm}

\subsection{Results}

Lemma \ref{lem:bias2} shows that the moment equation \eqref{eq:zqwq} incurs only a second-order bias due to the sign mistake of $V(\eta_0)$ as long as $V(\eta_0)$ is continuously distributed. In contrast to the setup of Lemma \ref{lem:powell}, the estimation error is not orthogonal to the space of $V(\eta_0)$. As a result, the first-order bias is bounded by invoking the margin assumption and the $\ell_{\infty}$-rate.

\begin{lemma}[First-Order Bias]
\label{lem:bias2}
Suppose Assumptions \ref{ass:boundedwidth} and \ref{ass:margin} holds. Then, the first-order bias shrinks at quadratic speed, that is, for some constant $\bar C$ large enough, 
\begin{align*}
\sup_{ \eta \in \mathcal{T}_N } | \E [z(q, \eta_0) (Y(q, \eta) - Y(q, \eta_0)) ] |  = O ((\eta^{\infty}_N)^2). 
\end{align*}
Likewise, the second-order bias shrinks at quadratic speed
\begin{align*}
\sup_{ \eta \in \mathcal{T}_N } |\E [ (z(q, \eta) - z(q, \eta_0)) (Y(q, \eta) - Y(q, \eta_0)) ] | = O ((\eta^{\infty}_N)^2). 
\end{align*}

\end{lemma}

\begin{definition}[Mean Square Rates]
\label{def:sqrate}
Define the mean square rates for the expectation functions
\begin{align*}
  \sup_{\gamma_L \in \Gamma_{L,N}}  \left(\E (\gamma_{L} (D,X) - \gamma_{L,0}(D,X))^2 \right)^{1/2} =: \gamma_{L,N} \\
 \sup_{\gamma_{UL} \in \Gamma_{UL,N}}  \left(\E (\gamma_{UL} (D,X) - \gamma_{UL,0}(D,X))^2 \right)^{1/2} =: \gamma_{UL,N} 
\end{align*}
and  for their derivatives 
\begin{align*}
  \sup_{\gamma_L \in \Gamma_{L,N}}  \left(\E \| \nabla_D \gamma_{L} (X) -\nabla_D \gamma_{L,0}(X)  \|^2 \right)^{1/2} =: \gamma^1_{L,N} \\
 \sup_{\gamma_{UL} \in \Gamma_{UL,N}} \left(\E \| \nabla_D \gamma_{UL} (X) -\nabla_D \gamma_{UL,0}(X)  \|^2 \right)^{1/2} =: \gamma^1_{UL,N} 
\end{align*}
and let $$
\gamma_N := \gamma_{L,N} + \gamma_{UL,N}  + \gamma^1_{L,N} +\gamma^1_{UL,N}.
$$

\end{definition}

\begin{corollary}[Asymptotic Theory for Average Partial Derivative with an Interval-Valued Outcome]
\label{cor:apd}
Suppose Assumptions   \ref{ass:regcond:apd} and \ref{ass:boundedwidth}   and \ref{ass:margin} hold. In addition, suppose $| \gamma_{L} (D,X) | + \| \nabla_D \gamma_{L} (D,X)  \| + | \gamma_{UL} (D,X) | + \| \nabla_D \gamma_{UL} (D,X)  \|< M_{\gamma} \text{ a.s.}$. Then, the sequences $\mu_N$ and $r_N'$ can be bounded as follows
$$
\mu_N = O  ( (\eta^{\infty}_N)^{3/2}+  \eta^{\infty}_N \cdot  \gamma_N).
$$
and $r_N' = O (\gamma_N + (\eta^{\infty}_N)^{1/2} ) $.   In particular, if $\eta^{\infty}_N = o (N^{-1/3})$ and $\eta^{\infty}_N \cdot  \gamma_N = o (N^{-1/2})$ and $\gamma_N = o(1)$, Assumption \ref{ass:ratebound} holds. Assumption \ref{ass:ratebound2} holds automatically since $A(W, \eta_0) = \Sigma = I_d$ is a known matrix. Finally,  Assumption \ref{ass:concentration:chap1} holds.  Then, Theorems \ref{thm:limit} and \ref{thm:bb} hold for the Support Function Estimator of Algorithm \ref{alg:apd} and the influence function equal to
\begin{align*}
h(W,q) =  g(W,q,\xi(q))  - \E [ g(W,q,\xi(q)) ],
\end{align*}
where $g(W,q,\xi(q))$ is given in (\ref{eq:rho:apd}).
\end{corollary}

\section{Simulation Study}
\label{sec:montecarlo}

In this section, I compare the performance of the classic (series-based non-orthogonal) and the proposed (lasso-based orthogonal) approaches in a moderate-dimensional sparse design. The first approach is to plug a least squares series first-stage estimator into the non-orthogonal moment equation \eqref{eq:zqwq}. The second one is to plug an $\ell_1$-regularized least squares series into the orthogonal moment equation \eqref{eq:orthomom}. Regularization helps to leverage the sparsity assumption and to reduce risk.

Consider the partially linear model of  Example \ref{ex:plp} with $d=2$ treatments.  The outcome equation is generated as
\begin{align}
Y &=D_1 \beta_1 + D_2 \beta_2 + X \cdot (c_{\theta} \theta_0) + U, \label{eq:theta0}
\end{align}
where $D=(D_1, D_2)$ is the treatment vector, $X \in \mathrm{R}^{p_X}$ is the $p_X$-covariate vector with $p_X=50$, and $U \sim N(0, \sigma_U^2)$ is a normal shock independent of $D$ and $X$ with $\sigma_U =1$. For each treatment $m=1,2$, its reduced form is 
\begin{align}
D_m &= X\cdot (c_{D_m} \alpha_m) + V_m, \label{eq:alpha1} \quad m=1,2,
\end{align}
where the coefficients $$\theta_0 = \alpha_1 = \alpha_2 = (1, 1/2^2, \dots, 1/j^2, \dots, 1/p_X^2). $$ The parameters in \eqref{eq:theta0} and \eqref{eq:alpha1} are chosen as
\begin{align*}
c_D& = (c_{D_1}, c_{D_2}) = (2,1), c_{\theta}=1, \beta_0 = (1,1).
\end{align*}
The covariates $X$ are generated from $N(0, \Omega)$, where $\Omega$ is a Toeplitz matrix with correlation coefficient $\rho = 0.5$. That is, for every $(i,j) \in \{1,2,\dots, p_X\}^2$, $\Omega_{ij} = \rho^{|i-j|}$.  The  vector $V$ is independent of $(X,U)$ and is drawn from the bivariate normal distribution 
\begin{align}
\label{eq:vee}
 V = (V_1, V_2) \sim N(0, \Sigma), \quad \Sigma = \begin{pmatrix} 1 & 0.5  \\ 0.5 & 1 \end{pmatrix}.
 \end{align}
The outcome $Y$ is not included into the data. Instead, the support of $Y$ is partitioned into  the  bins $ \cup_{s=1}^{S} [b_s, b_{s+1})$ of width $\Delta$:
$$
b_{s+1} = b_s + \Delta,  \quad s=1, 2, \dots, S-1, 
$$
where $b_0 = -\infty$ and $b_{S+1}  = \infty$.  The observed bounds $Y_L$ and $Y_U$ are taken to be
\begin{align*}
[Y_L, Y_U]:&= \sum_{s=1}^{S} [b_s, b_{s+1}) \cdot 1 \{ Y \in [b_s, b_{s+1})  \}.
\end{align*}
Thus, the observed data vector $W=(X, D, Y_L, Y_U)$ but does not contain $Y$.

I now derive the true (population) support function. Plugging $Y_U - Y_L = \Delta$ into \eqref{eq:zqwq} gives
$$
\sigma(q) =  q' \Sigma^{-1} \E V(\eta_0) Y(q, \eta_0) = q' \Sigma^{-1} \E V Y_L +  \Delta \E \max (q' \Sigma^{-1} V,0).
$$
Invoking \eqref{eq:vee} gives $q'  \Sigma^{-1} V \sim N(0, q' \Sigma^{-1} \Sigma \Sigma^{-1}  q) = \sqrt{q' \Sigma^{-1} q} N(0, 1)$, which implies
$$
\E \max (q' V, 0)  = \sqrt{q' \Sigma^{-1} q/2\pi}.
$$
Thus,  the support function is
\begin{align}
\label{eq:sigmatrue}
\sigma(q) = q'\kappa_0 + \Delta \sqrt{q' \Sigma^{-1} q/2 \pi}  , \quad \kappa_0 := \Sigma^{-1} \E V Y_L.
\end{align}

The classic  and the proposed approaches are implemented via Algorithm \ref{alg:plpiv} with different sets of the first-stage fitted values. For each treatment $m \in \{1,2\}$,  the first-stage regression parameter $\alpha$ is estimated as
\begin{align}
\label{eq:lasso}
\widehat{\alpha}_{m} :&= \arg \min_{b \in \mathrm{R}^{p_X}} \dfrac{1}{N} \sum_{i =1}^N (D_{im} - X_i' b)^2 + \lambda_{D_m} \| b \|_1,
\end{align}
where $\lambda_{D_m} = 0$ in the series-based case and $\lambda_{D_m} > 0$ in the lasso-based case. For the lasso estimator, the  penalty parameter $\lambda_{D_m}$ is chosen according to Algorithm 1 in \cite{Program} (i.e., the default value of \url{rlasso} package).  In both cases, the treatment fitted values are
\begin{align*}
\widehat{\eta}(X)&= X' (\widehat{\alpha}_1, \widehat{\alpha}_2). 
\end{align*}
In the orthogonal case (the proposed approach),  the best-case outcome fitted values are
\begin{align*}
\widehat{\gamma}_U(X)&= X' \widehat{\gamma}_L + \dfrac{1}{2} \Delta,
\end{align*}
where  $\gamma_L$ is estimated by Lasso regression of $Y_L$ on $X$ similarly to \eqref{eq:lasso}. Since the classic case does not require partialling out,
the fitted values $\widehat{\gamma}_U(X)$ are set to zero.

The  estimator's  performance  is summarized in terms of  its risk and coverage. The total risk of the estimator is defined as
\begin{align}
\label{eq:total}
R_H := \sup_{q \in \mathcal{S}^{d-1}} | \widehat{\sigma}(q) - \sigma(q) |.
\end{align}
It coincides with the Hausdorff distance $d(\widehat{\mathcal{B}},\mathcal{B})$ between the estimated ($\widehat{\mathcal{B}}$) and the true ($\mathcal{B}$) sets. Furthermore, the outer and the inner risks are defined as
\begin{align}
\label{eq:outerinner}
R_O:= \sup_{q \in \mathcal{S}^{d-1}} \max (\widehat{\sigma}(q) - \sigma(q), 0), \quad R_I:= \sup_{q \in \mathcal{S}^{d-1}} \max ( \sigma(q) - \widehat{\sigma}(q), 0).
\end{align}
 The rejection frequency is the share of rejected simulation draws
\begin{align}
\label{eq:boot}
 \dfrac{1}{N_S} \sum_{s=1}^{N_S} 1\{ R_H^s  > c^{*}_{1- \alpha} \},
\end{align}
where $c^{*}_{1- \alpha} $ is the $(1-\alpha)$-quantile of $R_H^b$ of the bootstrap process
\begin{align}
\label{eq:totalboot}
R^b_H := \sup_{q \in \mathcal{S}^{d-1}} | \widehat{\sigma}^b(q) - \widehat{\sigma}(q) |.
\end{align}

Table \ref{tab:sims} compares the classic  and the proposed estimators in terms of risk and coverage. Across the board, the proposed  estimator has a smaller total risk than the classic one, by a factor of ranging from $2.0$ for $\Delta=1$ to $3.0$ for $\Delta=3$. The classic estimator has higher risk due to the excessive noise of series estimators $\widehat{\alpha}_1$  and $\widehat{\alpha}_2$ in a regime with $p_X =50$ covariates. 

I investigate the importance of  the first-stage regularization  and second-stage orthogonalization, applying one at a time. Table \ref{tab:simsapp} compares the non-orthogonal (Columns (1)-(4)) and the orthogonal (Columns (5)-(8)) estimators based on the true treatment first stage. Across the board, the orthogonal  estimator has  smaller total risk than the non-orthogonal one, by a factor of $2.5$ on average. The variance reduction occurs due to partialling out the relevant controls from the outcome $Y^{\text{best}}(\eta_0)$. In contrast, the risks of series-based non-orthogonal (Table \ref{tab:sims}, Column (3)) and orthogonal (Table \ref{tab:simsapp}, Column (7)) estimators are close to each other. In particular, the high risk of the series-based estimator cannot be improved by orthogonalization.

Next, I compare the ortho (Table \ref{tab:sims}, Columns 5--8)  and non-ortho (Table \ref{tab:simsapp2}, Columns 1--4)  estimators based on the lasso-based first stage. 
Across the board, the risk of the ortho version is substantially smaller than the non-ortho one, by a factor ranging from $2$ to $5.5$. The non-ortho version has a higher risk, because the estimates of non-zero coefficients in  $\alpha_m$ are shrunk to zero. While the shrinkage bias could be reduced by invoking the post-lasso instead of the lasso estimator, post-single-selection inference may not be robust to moderate deviations from zero.

\section{Empirical illustration}
\label{sec:empirical}

This section demonstrates the proposed approach by estimating the gender wage gap with a bracketed wage variable. First, I show that a frequent empirical practice -- midpoint regression -- gives biased results. Instead of this approach, I propose reporting identified set (i.e., the lower and the upper bound) for the parameter of interest and demonstrate how to estimate the set.

The sample for the analysis comes from the U.S. March Supplement of the Current Population Survey (CPS) in 2015, as studied in \cite{MulliganRubinstein} and \cite{CCMS}. The selected sample consists of white non-hispanic individuals, aged 25 to 64 years, working more than 35 hours per week during at least 50 weeks of the year. The resulting sample comprises $32,523$ workers, including $18,137$ men and $14,386$ women. The object of interest is the gender wage gap -- the average difference in log wages between men and women after controlling for the observed characteristics. The data set is augmented by the lower bound $Y_L$ and the upper bound $Y_U$ defined as
\begin{align*}
[Y_L, Y_U]:&= \sum_{s=1}^{S} [b_s, b_{s+1}) \cdot 1 \{ Y \in [b_s, b_{s+1})  \},
\end{align*}
where $b_0=1$ and $$  b_{s+1} - b_s = \Delta  \quad s=1,2,\dots, S.$$

I assume that the log wage variable follows the partially linear regression of Example \ref{ex:plp}
$$
Y = D' \beta_0 + X' \gamma_0+ U, \quad \E[ U \mid X, D] =0.
$$
Here, the outcome variable $Y$ is the logarithm of the hourly wage rate, the treatment/policy variable $D$ is an indicator for female gender, and the vector of $p_X=260$ controls includes demographic indicators, region, and experience indicators, as well as their interactions. The coefficients $\beta_0$ and $\gamma_0$ are the target and the nuisance parameters, respectively. The parameter $\beta_0$ is identified as a minimizer of the least squares loss function
 $$
 \beta_0:=\arg \min_{b \in \mathrm{R}} \E (Y - (D- \eta_0(X))b)^2.
 $$
When $Y$ is unobserved,  a frequent approach is to replace $Y$ by the bracket midpoint
 $$Y_M:=(Y_L+Y_U)/2.$$ 
 The ``mid-point'' regression parameter is taken to be
 $$
 \beta_{M}:=\arg \min_{b \in \mathrm{R}} \E (Y_M - (D- \eta_0(X))b)^2,
 $$
 which can differ from $\beta_0$. I report the estimate and the $95 \%$ CI of $\beta_0$ and $\beta_M$.    The first-stage  treatment and the outcome expectation functions are  estimated via logistic and linear Lasso regression of \cite{Program} implemented in the hdm $R$ package, respectively.  In addition, I also consider the random forest estimator as implemented in the ranger $R$ package. The final estimator  is taken to be the Double Machine Learning estimator of \cite{chernozhukov2016double} with $K=2$-fold cross-fitting. The Lasso-based first-stage fitted values (Columns (1)-(2)) and the random-forest-based (Columns (3)-(4)).

Instead of reporting $\beta_M$ -- which is a biased measure of $\beta_0$ -- I propose reporting  the lower and the upper bound on $\beta_0$. The moment equation for $\beta_U$ is given in \eqref{eq:betu1d} and its estimate is defined in  the Algorithm \ref{alg:plp} with the fitted values described below. The symmetry-based specification (SYM) imposes Assumption \ref{ass:suffcond}, and the first-stage fitted values are taken to be \eqref{eq:crossfitgammal}. The sparsity-based specification (SPRS) imposes the model \eqref{eq:maindecomp}, and the first-stage are taken to be as in Example \ref{ex:lassologistic}, \eqref{eq:finalfittedlogis}. The treatment expectation function $\eta_0(\cdot)$ is estimated the same as in the point-identified specifications.

Table \ref{tab:empapp} summarizes the findings for the bracket width $\Delta \in \{1,2,3\}$. The ``ground-truth'' gender wage gap ranges between $18 \%$ (Columns (3)-(4)) and $20 \%$ (Columns (1)-(2)). I call this estimate ``ground-truth'' because this is the estimate to be reported if the wage $Y$ was observed. When the bracket width $\Delta=1$ is small, the midpoint estimate $\beta_M$ is close to the estimate of $\beta_0$. When $\Delta=2$, the midpoint estimate $\beta_M$ ranges between $25\%$ (RF) and $28 \%$ (Lasso). Furthermore, the 95 $\%$ CI for $\beta_M$ does not contain the ``ground-truth`` estimate of $\beta_0$ for either RF or Lasso first-stage method. When $\Delta=3$, the magnitude of the bias remains substantial, which speaks against midpoint regression for large values of bracket width.  

To estimate $[\beta_L, \beta_U]$, I consider four specifications: SYM-Lasso,  SYM-RF, SPRS-Lasso and SPRS-RF. Across the board, the bounds contain $\beta_0$. Furthermore, the specifications yield close results  despite being based on very different starting assumptions. That said, the bounds $[\beta_L, \beta_U]$ come out wide. As discussed in Remark \ref{rm:sharp}, the bounds $[\beta_L, \beta_U]$ may not be sharp for $\beta_0$, since they are utilize only $d$ (out of infinitely many) moment restriction implied by \eqref{ex:plp}. The derivation of sharp bounds for $\beta_0$ is left for the future work.

\begin{table}
\centering
\caption{Finite-sample performance of the classic (series-based) and the proposed (Lasso-based) methods}
\begin{tabular}{c|cccccccc}
\toprule
  & \multicolumn{4}{c}{Series-based}& \multicolumn{4}{c}{Lasso-based} \\ 
   \hline \\
 $N$  & Total & Outer & Inner & Rej.freq & Total & Outer & Inner & Rej.freq   \\
  \\  
   & \multicolumn{8}{c}{Panel A: Bracket width $\Delta = 1$}\\
\\
$250$ &  0.24 & 0.18 & 0.23 & 0.03 & 0.13 & 0.11 & 0.11 & 0.09 \\ 
 $500$ &    0.19 & 0.09 & 0.19 & 0.25 & 0.09 & 0.08 & 0.08 & 0.08 \\ 
  $700$ &  0.19 & 0.07 & 0.19 & 0.43 & 0.08 & 0.07 & 0.07 & 0.04 \\ 
  $1,000$ &  0.18 & 0.06 & 0.18 & 0.99 & 0.07 & 0.06 & 0.06 & 0.10 \\ 
  \\
   & \multicolumn{8}{c}{Panel B: Bracket width $\Delta = 2$}\\
   \\
$250$ &  0.35 & 0.24 & 0.34 & 0.05 & 0.17 & 0.14 & 0.13 & 0.09 \\ 
 $500$ &   0.33 & 0.13 & 0.33 & 0.85 & 0.12 & 0.10 & 0.09 & 0.09 \\
    $700$ & 0.32 & 0.09 & 0.32 & 0.99 & 0.10 & 0.08 & 0.08 & 0.03 \\ 
   $1,000$ &  0.32 & 0.07 & 0.32 & 1.00 & 0.09 & 0.07 & 0.07 & 0.10 \\ 
     \\
       & \multicolumn{8}{c}{Panel C: Bracket width $\Delta = 3$}\\
    \\
$250$ & 0.48 & 0.32 & 0.47 & 0.10 & 0.21 & 0.17 & 0.16 & 0.07 \\ 
 $500$ & 0.46 & 0.16 & 0.46 & 0.98 & 0.15 & 0.12 & 0.12 & 0.07 \\ 
    $700$ &   0.46 & 0.12 & 0.46 & 1.00 & 0.13 & 0.11 & 0.10 & 0.04 \\
   $1,000$ & 0.46 & 0.09 & 0.46 & 1.00 & 0.11 & 0.09 & 0.09 & 0.06 \\ 
\bottomrule
\end{tabular}
\label{tab:sims}
\caption*{Notes. Results are based on 10, 000 simulation runs.  Panels A, B and C correspond to the bin width $\Delta = 1, 2, 3$.    Table shows the total risk \eqref{eq:total}, the outer and inner risks \eqref{eq:outerinner}, and the rejection frequency \eqref{eq:boot} for the nominal size $\alpha = 0.05$. The supremum over $\mathcal{S}^1$ is approximated by the maximum over the grid consisting of $ 50$ evenly spaced points on unit circumference $\mathcal{S}^1$. Columns (1--4) and (5--8) correspond to the classic and the proposed approach. The number of bootstrap repetitions $B=2, 000$. The true support function $\sigma(q)$ is in \eqref{eq:sigmatrue}. For the description of estimators, see text.    } \end{table}

\newpage

\begin{table}[H]
\centering
\small
\caption{Bounds on gender wage gap with bracketed log wage}
\begin{tabular}{c|cc|cc|}
\toprule
  & \multicolumn{2}{c}{Lasso}& \multicolumn{2}{c}{RF} \\ 
  & Estimated Set & 95 $\%$ CI   & Estimated Set & 95 $\%$ CI \\ 
     \\  
   & \multicolumn{4}{c}{Panel A: Bracket width $\Delta = 1$}\\
\\
True $\beta_0$ & -0.200  & (-0.213, -0.186) & -0.180 & (-0.194, -0.167) \\ 
 $\beta_M$ & -0.199  & (-0.215, -0.184) & -0.181  & (-0.196, -0.165) \\ 
 $[\beta_L, \beta_U]$ (SYM) & [ -1.195, 0.796] & (-1.211, 0.813)  & [-1.133, 0.771] & (-1.150, 0.788)  \\ 
  $[\beta_L, \beta_U]$ (SPRS)  &  [-1.193, 0.794] & (-1.209, 0.810) & [-1.115, 0.754] & (-1.131, 0.770) \\ 
  \\  
   & \multicolumn{4}{c}{Panel B: Bracket width $\Delta = 2$}\\
\\
   
True $\beta_0$ & -0.200  & (-0.213, -0.186) & -0.180 & (-0.194, -0.167) \\ 
$\beta_M$  &-0.279 & (-0.302, -0.256) & -0.251 & (-0.273, -0.228) \\ 
 $[\beta_L, \beta_U]$ (SYM) &[-2.270,  { }1.712] & (-2.295, 1.737) & [-2.155, 1.654] & (-2.180, 1.679) \\ 
 $[\beta_L, \beta_U]$ (SPRS)   &[-2.266  { }1.708] & (-2.290, 1.731) & [-2.120, 1.618] & (-2.144, 1.643) \\

     \\  
   & \multicolumn{4}{c}{Panel C: Bracket width $\Delta = 3$}\\
\\
True $\beta_0$ & -0.200  & (-0.213, -0.186) & -0.180 & (-0.194, -0.167) \\ 
$\beta_M$  &  -0.158  & (-0.179, -0.137) & -0.131  & (-0.152, -0.110) \\ 
  $[\beta_L, \beta_U]$ (SYM)  &  [-3.145, {} 2.828] & (-3.172, 2.854) & [-2.988, 2.725] & (-3.015, 2.752) \\ 
   $[\beta_L, \beta_U]$ (SPRS)  & [-3.139, {}2.822] & (-3.162, 2.844) & [-2.935, 2.672] & (-2.959, 2.697) \\ 
\bottomrule
\end{tabular}
\label{tab:empapp}
\caption*{Notes. Estimated parameter (square brackets) and the $95\%$ confidence bands (parentheses) for the parameter.  The true parameter $\beta_0$ is based on the observed outcome $Y$. The midpoint parameter $\beta_M$ is based on the midpoint outcome $Y_M$. The upper bound $\beta_U$ is as defined in \eqref{eq:betu1d}, and $\beta_L$ is its analog.  The first-stage treatment expectation function is estimated by linear Lasso (Columns (1)-(2)) and random forest (Columns (3)-(4)).  The symmetry-based specification (SYM, Row 3) is based on Assumption \ref{ass:suffcond}, and the first-stage fitted values are given in 
\eqref{eq:crossfitgammal}. The sparsity-based specification (SPRS, Row 4) is based on Example \ref{ex:lassologistic}, and the first-stage fitted values are given in   \eqref{eq:finalfittedlogis}.  For more details, see text.  }
\end{table}

\newpage 

\section{Proofs}
\label{sec:proofs}

\paragraph{Empirical process notation. }  Let $\widehat{\eta}_k$ and $\widehat{\xi}_k, \quad k=1,2,\dots, K$ be as in Definition \ref{sampling}.  Define an event 
\begin{align*}
\mathcal{E}_N &:= \{ \widehat{\eta}_k, (\widehat{\xi}_k(p))_{p \in \mathcal{P}} \in \Xi_N \quad \forall k=1,2,\dots,K \}.
\end{align*}  By union bound, this event holds with probability approaching one $$\Pr (\mathcal{E}_N ) \geq 1- K \epsilon_N = 1-o(1).$$ 
For a given partition $k$ in $\{1,2, \dots, K\}$, define the partition-specific averages
\begin{align*}
\Enk f(W_i) :&= \dfrac{1}{n} \sum_{i \in J_k} f(W_i), \\
 \Gnk f(W_i) :&= \dfrac{1}{\sqrt{n}} \sum_{i \in J_k} [f(W_i) - \int f(w) dP (w)].
 \end{align*}
Define the function $\psi_0(p)$ 
 \begin{align}
\psi_0(p) &= \psi(p, \xi_0) = \E [ g(W, p, \xi_0(p))] \label{eq:psi0}
\end{align}
and observe that plugging $p_0(q)$ into $\psi_0(p)$ gives the support function $ \sigma(q) $ at $q$:
$$ \psi_0(p_0(q))= \psi_0(\Sigma^{-1}q)=\sigma(q).$$ 
For $i \in J_k$ and $k=1,2,\dots, K$, define the partition-specific conditional expectation
\begin{align}
\label{eq:psi}
\psi(p, \widehat{\xi}_k)&:= \E [ g (W_i, p, \widehat{\xi}_k(p))  \mid  (W_i)_{i \in J_k^c} ], \quad i \in J_k
\end{align}
and its weighted sample analog 
\begin{align}
\label{eq:psihat}
\widehat{\psi}_k^v (p, \widehat{\xi}_k) := \Enk v_i g (W_i, p, \widehat{\xi}_k(p)).
\end{align}
Finally, define the matrix error terms 
\begin{align*}
\widehat{\Sigma}_k(\widehat{\eta}_k)&:= \Enk A(W_i, \widehat{\eta}_k), \quad \widehat{\Sigma}(\widehat{\eta}) := \dfrac{1}{K} \sum_{k=1}^K \widehat{\Sigma}_k(\widehat{\eta}_k)
\end{align*}
and the weighted matrix error 
\begin{align*}
\widehat{\Sigma}^v_k(\widehat{\eta}_k)&:= \Enk v_i A(W_i, \widehat{\eta}_k), \quad \widehat{\Sigma}^v(\widehat{\eta}) := \dfrac{1}{K} \sum_{k=1}^K \widehat{\Sigma}^v_k(\widehat{\eta}_k).
\end{align*}

 \paragraph{Empirical process remainder terms.} Define the remainder term
\begin{align*}
R_{1,k} (p)&= \sqrt{N}( (\widehat{\psi}_k(p, \widehat{\xi}_k) - \widehat{\psi}_k(p_0, \xi_0))  - (\psi(p, \widehat{\xi}_k)- \psi_0(p_0))  ), \quad k=1,2,\dots, K,
\end{align*}
the bias term
\begin{align*}
R_{2,k} (p)&= \sqrt{N}(  \psi(p, \widehat{\xi}_k)- \psi_0(p) ), \quad k=1,2,\dots, K,
\end{align*}
 the second-order remainder term
\begin{align*}
R(p, p_0)&= \sqrt{N}( \psi_0(p) - \psi_0(p_0) -G(p_0)' (p- p_0))
\end{align*}
and the bootstrap term
\begin{align*}
R_{1,k}^v (p)&= \sqrt{N}( (\widehat{\psi}^v_k(p, \widehat{\xi}_k) - \widehat{\psi}^v_k(p_0, \xi_0))  - (\psi(p, \widehat{\xi}_k)- \psi_0(p_0))  ), \quad k=1,2,\dots, K.
\end{align*}
Thus, the support function process $S_N(q)$ of Theorem \ref{thm:limit} can be decomposed as 
\begin{align*}
S_N(q):&= \sqrt{N}  (\widehat{\sigma}(q) - \sigma(q))  \\
&= \sqrt{N} \left(\dfrac{1}{K} \sum_{k=1}^K  \widehat{\psi}_k (\widehat{p}(q), \widehat{\xi}_k) - \sigma(q) \right) \\
 &=\sqrt{N} \left(\dfrac{1}{K} \sum_{k=1}^K \widehat{\psi}_k(p_0(q), \xi_0) - \sigma(q) + G(p_0)' (\widehat{p}(q) - p_0(q)) \right) \\
  &+\dfrac{1}{K} \sum_{k=1}^K [R_{1,k} (\widehat{p}(q))+ R_{2,k} (\widehat{p}(q)) ] + R(\widehat{p}(q), p_0(q)).
\end{align*}

\paragraph{Misclassification events.}  For $p, p_0 \in \mathcal{P}$, define the events  $\mathcal{E}_{+}(p), \mathcal{E}_{-}(p),  \mathcal{E}_{-}(p,p_0), \mathcal{E}_{+}(p,p_0) $ 
 \begin{align}
 \label{eq:e+p}
 \mathcal{E}_{+}(p) &= \bigg\{ p' V ( \eta)  < 0 < p' V ( \eta_0) \bigg\}, \\
   \mathcal{E}_{-}(p) &= \bigg\{ p' V ( \eta_0)  < 0 < p' V ( \eta) \bigg\}  \label{eq:e-p} 
 \end{align} 
and
\begin{align}
       \mathcal{E}_{-}(p,p_0) &= \{ p_0' V(\eta_0)  < 0 < p' V(\eta_0)   \} \label{eq:e-pp0}  \\
      \mathcal{E}_{+}(p,p_0) &= \{ p' V(\eta_0) < 0 < p_0' V(\eta_0)   \}. \label{eq:e+pp0} 
      \end{align}

\subsection{Proofs of Main Results.}

\begin{proof} [Proof of Lemma \ref{lem:powell}]
 Define 
 \begin{align*}
 z(p, \eta):&= p' V(\eta) \\
 B_1 (W, \eta, p)  :&=  p' V ( \eta_0) ( Y(p, \eta)  - Y(p, \eta_0) ) \\
 B_2 (W, \eta, p)  :&=  p' ( V (\eta) - V ( \eta_0)) ( Y(p, \eta)  - Y(p, \eta_0) ).
 \end{align*}
 Observe that 
 \begin{align*}
 z(p, \eta) ( Y(p, \eta) - Y(p, \eta_0)) =  B_1 (W, \eta, p)  +  B_2 (W, \eta, p).
 \end{align*}
The mistake in $Y(p, \eta)$ can only occur if $p' V ( \eta_0)$ is small enough
 \begin{align}
 \bigg\{ Y(p, \eta) \neq  Y(p, \eta_0) \bigg\} &\Leftrightarrow   \bigg\{ \mathcal{E}_{+}(p)  \text{ or } \mathcal{E}_{-}(p)\bigg\} \nonumber \\
 &\Rightarrow \bigg\{ 0 < | p' V ( \eta_0)  | < | p' (V ( \eta) - V (\eta_0))| \bigg\} \nonumber \\
 &\Rightarrow  \bigg\{ 0 < | p' V ( \eta_0)  | <  C_P \| V ( \eta) - V (\eta_0) \| \bigg\}=: \mathcal{E}_{+-}(p) .  \label{eq:peta0bound} 
 \end{align}
Recall that  
\begin{align}
\label{eq:pp0}
Y(p, \eta) -Y(p, \eta_0)=(Y_U - Y_L) 1\{  \mathcal{E}_{+}(p) \cup \mathcal{E}_{-}(p)\} 
\end{align}
Invoking $| \E X | \leq \E | X| $ and \eqref{eq:pp0} gives
 \begin{align}
| \E B_1 (W, \eta, p) |&=  | \E p' V ( \eta_0)  (Y(p, \eta) -Y(p, \eta_0) )  | \nonumber \\
&\leq \E | p' V ( \eta_0)  | (Y_U - Y_L) 1\{  \mathcal{E}_{+}(p) \cup \mathcal{E}_{-}(p)\} \nonumber  \\
&\leq  \E | p' V ( \eta_0)  | (Y_U - Y_L) 1\{  \mathcal{E}_{+-}(p) \}  \label{eq:peta0bound2} 
\end{align} 
Invoking definition of $\mathcal{E}_{+-}(p)$ in \eqref{eq:peta0bound} and $Y_U - Y_L \leq M_{UL} \text{ a.s. }$ gives
 \begin{align}
&\E | p' V ( \eta_0)  | (Y_U - Y_L) 1\{   \mathcal{E}_{+-}(p) \} \nonumber \\
&\leq C_P M_{UL} \E \| V ( \eta) - V (\eta_0) \| 1\{  \mathcal{E}_{+-}(p) \}. \label{eq:peta0bound3} 
\end{align} 
The second-order bias term is bounded as
\begin{align}
| \E B_2 (W, \eta, p) | \leq C_P M_{UL} \E \| V ( \eta) - V (\eta_0)\| 1\{  \mathcal{E}_{+}(p) \cup \mathcal{E}_{-}(p)\}. \label{eq:peta0bound4} 
\end{align}
Invoking Assumption \ref{ass:regularity} gives
 \begin{align}
  &\E \|  \eta_0(X) - \eta(X) \| 1\{  \mathcal{E}_{+}(p) \cup \mathcal{E}_{-}(p)\}  \nonumber  \\
&=\E_{X} \| \eta(X) - \eta_0(X) \| \int_{- C_P \| \eta(X) - \eta_0(X) \|    }^{C_P \| \eta(X) - \eta_0(X) \|  }h_{p' V(\eta_0) \mid X} (t, X) dt \nonumber  \\
&\leq 2 C_P M_h   \E_{X} \| \eta(X) - \eta_0(X) \|^2 \label{eq:lipbound}.
\end{align}
 Combining the bounds  gives
\begin{align}
| \E [B_1 (W, \eta, p) + B_2 (W, \eta, p)] | &\leq |\E B_1 (W, \eta, p)  | + |  \E B_2 (W, \eta, p)  | \nonumber \\
&\leq 4 C_P^2 M_{UL} M_{h} \E  \| \eta(X) - \eta_0(X) \|^2  \label{eq:powell}.
\end{align}
 \end{proof}

 \begin{proof}[Proof of Theorem \ref{thm:limit}]
 \textbf{Step 1.} This step is required only if $\Sigma$ is unknown, such as in Example \ref{ex:plpiv}. As shown in Lemma \ref{lem:matrixlinarization}, for $v=1$ (regular case) and $v=e$ (bootstrap case), 
\begin{align}
\label{eq:sigma4main}
 (\widehat{\Sigma}^v (\widehat{\eta}))^{-1}  - \Sigma^{-1}  =  -\Sigma^{-1} ( \widehat{\Sigma}^v (\eta_0) - \Sigma ) \Sigma^{-1} + M^v,
 \end{align}
 where the remainder matrix $M^v$ obeys $\| M^v \| = o_P (N^{-1/2})$ for both cases. Take $v=1$. Post-multiplying the LHS above  by $q$ gives
  \begin{align}
 \sqrt{N} (\widehat{p}(q) - p_0(q)) &= -  \Sigma^{-1} ( \widehat{\Sigma} (\eta_0) - \Sigma ) \Sigma^{-1}q + o_P(1) \nonumber
 \end{align}
 Likewise, taking $v=e$ gives
  \begin{align*}
 \sqrt{N} (\widetilde{p}(q) - p_0(q)) &=  ((\widehat{\Sigma}^e (\widehat{\eta}))^{-1}  - \Sigma^{-1})'q \\
 &=  -  \Sigma^{-1} ( \widetilde{\Sigma}^e (\eta_0) - \Sigma ) \Sigma^{-1}q + o_P(1).
 \end{align*}
For some $N$ large enough, $ \widehat{p}(q)  \in \mathcal{P} \quad \forall q \in \mathcal{S}^{d-1}$  and $ \widetilde{p}(q)  \in \mathcal{P} \quad \forall q \in \mathcal{S}^{d-1}$ with probability $1-o(1)$.

\textbf{ Step 2. } Let $k=1,2,\dots, K$ denote the partition index. We  bound $R_{1,k} (\widehat{p}(q))$ and  $R^e_{1,k} (\widehat{p}(q))$. Define the function class
\begin{align*}
     \mathcal{F}_{2k}^v := \{ v\cdot(g(\cdot,p,\widehat{\xi}_k(p)) - g(\cdot,p_0,\xi_0(p_0))), \quad p , p_0 \in \mathcal{P},  \quad \| p - p_0 \| \leq \tau_N \},
\end{align*}   
where $v=1$ (regular case) and $v=e$ (bootstrap case). The class $\mathcal{F}_{2k}$ is obtained as $$\mathcal{F}_{2k} \subset \mathcal{G}_{\widehat{\xi}_k} - \mathcal{G}_{\xi_0},$$ where $\mathcal{G}_{\xi_0}$ and $\mathcal{G}_{\widehat{\xi}_k} $ are defined in Assumption \ref{ass:concentration:chap1}. On the event $\mathcal{E}_N$, 
\begin{align*}
     &\sup_{p \in \mathcal{P} } |g(W_i,p,\widehat{\xi}_k(p)) - g(W_i,p_0,\xi_0(p_0)) | \\
     &\leq   \sup_{p \in \mathcal{P}}  | g(W_i,p,\widehat{\xi}_k(p)) |   +    \sup_{p \in \mathcal{P}}   | g(W_i,p,\xi_0(p)) | \\
    &\leq G_{\widehat{\xi}_k} + G_{\xi_0},
\end{align*}
and $G_{\widehat{\xi}_k \xi_0} := G_{\widehat{\xi}_k} + G_{\xi_0}$ is a measurable envelope for the class $\mathcal{F}_{2k}$. Note that $\| G_{\widehat{\xi}_k \xi_0} \|_{P,c} \leq  \| G_{\xi_0} \|_{P,c} +  \| G_{\widehat{\xi}_k } \|_{P,c} \leq 2 C_1$.  The uniform covering entropy of the function class $ \mathcal{G}_{\widehat{\xi}_k} - \mathcal{G}_{\xi_0}$ is bounded as
\begin{align*}
& \log \sup_{Q} N(\epsilon \| G_{\xi}  +  G_{\xi_0}  \|_{Q,2}, \mathcal{G}_{\widehat \xi} - \mathcal{G}_{\xi_0} , \| \cdot \|_{Q,2}) \\
 &\leq  \log \sup_{Q} N(\epsilon/2 \| G_{\xi}    \|_{Q,2},  \mathcal{G}_{\xi} , \| \cdot \|_{Q,2}) + \log \sup_{Q} N(\epsilon/2 \| G_{\xi_0}    \|_{Q,2},  \mathcal{G}_{\xi_0} , \| \cdot \|_{Q,2}) \\ 
 &\leq  2v \log (2a/\epsilon)
\end{align*}
by the proof of Theorem 3 in \cite{andrews:1994b} and Assumption \ref{ass:concentration:chap1}. The class  $\mathcal{F}_{2k}^e $ is obtained by multiplication of $\mathcal{F}_{2k}$ by an integrable random variable independent of the data, and therefore retains $P$-Donsker and uniform covering properties of the class  $\mathcal{F}_{2}$. In particular,
$G^v_{\widehat{\xi}_k \xi_0} := |v| (G_{\widehat{\xi}_k} + G_{\xi_0})$ is a valid envelope for $     \mathcal{F}_{2k}^v $. Next, for $v=1$ and $v=e$,
\begin{align*}
&\sup_{\xi \in \Xi_N} \E  v^2 (g(W,p,\xi(p)) - g(W,p_0,\xi_0(p_0)) )^2 \\
&\leq  2  \bigg( \sup_{\xi \in \Xi_N} \sup_{p \in \mathcal{P}} \E v^2 ( g(W,p,\xi(p)) - g(W,p,\xi_0(p)))^2 \\
  &+   \sup_{p_0, p \in \mathcal{P}, \quad \| p - p_0 \| \leq \tau_N}  \E v^2 (g(W,p,\xi_0(p))  - g(W,p_0,\xi_0(p_0)) )^2 \bigg) \\
  &\leq 2((r_N'')^2 + (r_N')^2).
\end{align*}
Invoking Lemma \ref{lem:maxineq} conditional on $(W_i)_{i \in J_k}$ and taking $\sigma = 2 (r_N' + r_N'')$ gives 
\begin{align*}
\sup_{f \in \mathcal{F}_2^v}  | \Gnk [f] | &= \sup_{ p_0, p \in \mathcal{P}} | \widehat{\psi}_k (p, \widehat{\xi}_k) - \widehat{\psi}_k(p_0, \xi_0) - (\psi(p, \widehat{\xi}_k) - \psi(p_0, \xi_0)) | \\
 &\lesssim_P (r_N''+r_N' )\log^{1/2} (1/(r_N''+r_N' )) + N^{-1/2+1/c} \log N \\
 &= o_P(1),
\end{align*}
where the last equality follows from Assumption \ref{ass:concentration:chap1}. By Lemma \ref{lem:cond}, $\sup_{f \in \mathcal{F}_2^v}  | \Gnk [f] |  = o_P(1)$ holds unconditionally.  as shown in Step 1, wp $1-o(1)$, $\widehat{p}(q) \in \mathcal{P} \quad \forall q$, and 
\begin{align*}
 \sup_{q \in \mathcal{S}^{d-1}}| R_{1,k} (\widehat{p}(q)) |  \lesssim_P \sup_{p \in \mathcal{P}}| R_{1,k} (p) |  \lesssim_P \sup_{f \in \mathcal{F}_2}  | \Gnk [f] | = o_P(1).
\end{align*}

\textbf{ Step 3. Bound on $R_{2,k} (\widehat{p}(q))$. }   On the event $\mathcal{E}_N$,   for any $k=1,2,\dots, K$
\begin{align*}
 \sup_{q \in \mathcal{S}^{d-1}}| R_{2,k} (\widehat{p}(q)) | &\leq \sup_{p \in \mathcal{P}} \sqrt{N}  | \psi(p, \widehat{\xi}_k) - \psi_0(p) | \\
&\leq \sup_{ \xi \in \Xi_N} \sup_{p \in \mathcal{P}}\sqrt{N} | \psi(p, \xi) - \psi_0(p) |  \lesssim \sqrt{N} \mu_N.
\end{align*}
Combining the bounds over a finite set $k=1,2,\dots, K$ gives
\begin{align*}
\dfrac{1}{K} \sum_{k=1}^K  \bigg[R_{1,k} (\widehat{p}(q)) + R_{2,k} (\widehat{p}(q)) \bigg]  = o_P(1).
\end{align*}

\textbf{ Step 4. Bound on $R(\widehat{p}(q), p_0)$. }  
By Assumption \ref{ass:jacobian}, Lemma \ref{lem:uniderivative} and Step 1, on the event $\mathcal{E}_N$, 
\begin{align*}
\sup_{q \in \mathcal{S}^{d-1}} | R(\widehat{p}(q), p_0(q)) |   = o (\sqrt{N} \| (\widehat{\Sigma} (\widehat{\eta}))^{-1} - \Sigma^{-1} \|)  = o_P(1).
\end{align*}

\textbf{ Step 5.  Conclusion. } For the influence function $h(W,q)$ in \eqref{eq:hwq}, the function class 
\begin{align*}
\mathcal{H} &= \bigg\{ h(\cdot, q), \quad q \in \mathcal{S}^{d-1}  \bigg\} \subseteq \mathcal{G}_{\xi_0} +\mathcal{H}_A
\end{align*}
is included into the sum $\mathcal{G}_{\xi_0} + \mathcal{H}_A $.  The function classes $\mathcal{G}_{\xi_0}$ and $\mathcal{H}_A $ are Donsker classes with square integrable envelopes (by Assumption \ref{ass:concentration:chap1} and Lemma \ref{lem:maxineq:suppfun},  respectively). The statement of the lemma follows from  the Skorohod-Dudley-Wichura construction, as in \cite{Skorohod}, \cite{Dudley} and \cite{Wichura}. 

\end{proof}
The bootstrap support function process can be decomposed as
\begin{align*}
\widetilde{S}_N(q) :&=\sqrt{N} \left( (\widetilde{\sigma}(q) - \sigma(q)) - (\widehat{\sigma}(q) - \sigma(q)) \right).
\end{align*}
The first summand can be decomposed as
\begin{align*}
\widetilde{\sigma}(q) - \sigma(q) &=\dfrac{1}{K} \sum_{k=1}^K \bar{e}^{-1} \widehat{\psi}_k^v(\widetilde{p}(q), \widehat{\xi}_k)       - \sigma(q)  \\
&=\dfrac{1}{K} \sum_{k=1}^K \widehat{\psi}_k^v(\widetilde{p}(q), \widehat{\xi}_k) / (1+ o_P(1)) - \sigma(q) \\
&=\dfrac{1}{K} \sum_{k=1}^K \widehat{\psi}_k^v(\widetilde{p}(q), \widehat{\xi}_k)  - \sigma(q)  + o_P(1).
\end{align*}
The weighted moment can be decomposed as 
\begin{align*}
\sqrt{N}  \dfrac{1}{K} \sum_{k=1}^K \widehat{\psi}_k^v(\widetilde{p}(q), \widehat{\xi}_k) &= \sqrt{N} \dfrac{1}{K} \sum_{k=1}^K \widehat{\psi}_k^v(p_0(q), \xi_0) + \sqrt{N}  G(p_0(q))' (\widetilde{p}(q) - p_0(q)) \\
&+  \dfrac{1}{K} \sum_{k=1}^K  \bigg[R_{1,k}^e (\widetilde{p}(q)) + R_{2,k} (\widetilde{p}(q)) \bigg]  + R(\widetilde{p}(q), p_0(q)).
\end{align*}

\begin{proof}[Proof of Theorem \ref{thm:bb}]
By Comment B.1 in \cite{CCMS}, if the bootstrap random element  converges in probability $P$ unconditionally (i.e,  $Z_N=o_P(1)$), then $Z_N = o_{P^e} (1)$
 in $L^1(P)$ sense and hence in probability $P$, where $P^e$ denotes the probability measure conditional on the data.   
 
\textbf{ Step 1. } As shown in Step 1 of the proof of Theorem \ref{thm:limit}, $\widetilde{p}(q) \in \mathcal{P} \quad \forall q$. The bound on $R_{1,k}^e (\widetilde{p}(q))$ is established in the proof of Theorem \ref{thm:limit}, Step 2. By construction,
\begin{align*}
 \sqrt{N}(\psi(p,  \widehat{\xi}_k) - \psi(p, \xi_0)) = R_{2,k} (p) = \E [ v \cdot (g(W, p, \widehat{\xi}_k) - g(W, p, \xi_0)) \mid  (W_i)_{i \in J_k} ].
\end{align*}
Thus, the bounds on $R_{2,k}(p)$ and  $R(p, p_0)$  are  established in Steps 3 and 4 of Theorem \ref{thm:limit}. 

\textbf{ Step 2. } As shown in Step 5 of Theorem \ref{thm:limit}, the function class $\mathcal{H} \subseteq \mathcal{G}_{\xi_0} + \mathcal{H}_A$ is a Donsker class with
square-integrable envelopes.  Then by the Donsker theorem for exchangeable bootstraps,  weak convergence holds conditional on the
data, 
\begin{align*}
\G_N [ (e-1) h(W,q) ] /\bar{e} \Rightarrow  \widetilde{\G[h(q)]}   \text{ under} P^e \text{ in probability } P,
\end{align*}
where $ \widetilde{\G[h(q)]} $ is a P-Brownian bridge independent of $G[h(q)]$ with the same distribution as $G[h(q)]$.

\end{proof}

\newpage
 
 \begin{abstract} 
    This appendix contains supplementary statements for the paper ``Debiased Machine Learning of Set-Identified Linear Models'' by Vira Semenova. Appendix A contains useful technical statements. Appendix B contains additional proofs. Appendix C contains supplementary tables.
    \end{abstract}

  \section*{Appendix A: Technical Lemmas }

\renewcommand{\theequation}{A.\arabic{equation}}
\renewcommand{\thelemma}{A.\arabic{lemma}}
\renewcommand{\thesection}{A}
\setcounter{equation}{0}
\setcounter{section}{0}
\setcounter{lemma}{0}
\setcounter{table}{0}
\medskip

\begin{lemma}[Conditional Convergence Implies Unconditional]
\label{lem:cond}
Let $\{X_m\}_{m \geq 1}$ and $\{Y_m\}_{m \geq 1}$ be sequences of random vectors. (i) If for $\epsilon_m \rightarrow 0, \Pr( \| X_m \| > \epsilon_m \mid Y_m ) \rightarrow_{P} 0,$ then  $\Pr( \| X_m \| > \epsilon_m) \rightarrow 0$. In particular, this occurs if  $\Ep [\| X_m \|^q /\epsilon_m^q \mid Y_m]  \rightarrow_P 0$ for some $q \geq 1$, by Markov inequality. (ii) Let $\{A_m\}_{m \geq 1}$ be a sequence of positive constants. If $\|X_m \| = O_{P} (A_m) $   conditional on $Y_m$, namely, that for any $\ell_m \rightarrow \infty, \Pr( \| X_m \| >  \ell_m A_m \mid Y_m) \rightarrow_P 0$ , then $ X_m = O_{P} (A_m)$ unconditionally, namely, that for any $\ell_m \rightarrow \infty, \Pr( \| X_m \| >  \ell_m A_m) \rightarrow 0$.
\end{lemma}
Lemma \ref{lem:cond} is a restatement of Lemma 6.1  in \cite{chernozhukov2016double}.

\begin{lemma}[Maximal Inequality]
\label{lem:maxineq}
Let  $\mathcal{F}$ be a set of suitably measurable functions $f: \mathcal{W} \rightarrow \mathrm{R}$.  Suppose that $F: \mathcal{W} \rightarrow  \mathrm{R}$, $F  \geq \sup_{f \in \mathcal{F}} | f| $ is a measurable envelope for $\mathcal{F}$ with $\| F \|_{P,c} < \infty$ for some $c \geq 2$. Let $M = \max_{i \leq N} F(W_i)$ and $\sigma'^2$ be any positive constant such that $\sup_{f \in \mathcal{F}} \| f \|_{P,2}^2 \leq \sigma'^2 \leq \| F \|_{P,2}^2$. Suppose that there exist constants $a \geq e$ and $v' \geq 1$ such that
$$
\log \sup_{Q} N( \epsilon \| F \|_{Q,2}, \mathcal{F}, \| \cdot \|_{Q, 2}) \leq v' \log (a / \epsilon), \quad 0 < \epsilon \geq 1.
$$
Then,
\begin{align}
\E_P [ \| \G_N \|_{\mathcal{F}} ]  \leq K \left( \sqrt{v' \sigma'^2 \log \left( \dfrac{a \| F \|_{P,2}}{\sigma'} \right) } + \dfrac{v' \| M \|_{P,2}}{\sqrt{N} }  \log \left( \dfrac{a \| F \|_{P,2}}{\sigma'} \right) \right),
\end{align}
where $K$ is an absolute constant. Furthermore, with probability $> 1- c_2 (\log N)^{-1}$, 
\begin{align}
\label{eq:maxineqkcc2}
\| \G_N \|_{\mathcal{F}} \leq K(c,c_2) \left( \sqrt{v' \sigma'^2 \log \left( \dfrac{a \| F \|_{P,2}}{\sigma'} \right) } + \dfrac{v' \| M \|_{P,c}}{\sqrt{N} }  \log \left( \dfrac{a \| F \|_{P,2}}{\sigma'} \right) \right),
\end{align}
where $\| M \|_{P,c} \leq N^{1/c} \| F \|_{P,c}$ and $K(c,c_2)>0$ is a constant depending only on $c$ and $c_2$.

\end{lemma}
Lemma \ref{lem:maxineq} is a restatement of Lemma 6.2  in \cite{chernozhukov2016double} (cf.  \cite{CCKAS}).

Define the subGaussian Orlicz norm
\begin{align*}
\| X \|_{\psi_2}  = \inf \{ t >0: \Ep \exp (X^2/t^2) \leq 2 \}
\end{align*}
and the subExponential Orlicz norm
\begin{align*}
 \|X \|_{\psi_1} =  \inf \{ t >0: \Ep \exp (|X|/t) \leq 2 \}.
 \end{align*}
 A random variable $X$ is $K$-subGaussian ($X$ is $K$-subExponential) if there exist a finite $K<\infty$ such that
 $$
 \| X \|_{\psi_t} \leq K,
 $$
 where $t=2$ corresponds to subGaussian case and $t=1$ to subExponential case. For a $d$-vector $X$, if $\| X \|$ is $K$-subGaussian, since
 $$
 | \alpha' X | \leq \| X \|, \quad \alpha \in \mathcal{S}^{d-1}
 $$
 $\alpha' X$ is $K$-subGaussian.

 We rely on the following properties of Orlicz norms. The product $XY$ of two $K$-subGaussian random variables $X$ and $Y$ obeys
 \begin{align}
 \label{eq:cauchy}
\| X Y \|_{\psi_1} \leq \| X \|_{\psi_2} \| Y \|_{\psi_2} \leq K^2.
\end{align}
Furthermore, if $X$ is $K^2$-subExponential random variable, its centered analog $X - \E X$ obeys
\begin{align}
\label{eq:centering}
\| X - \E X \|_{\psi_1} \leq C\| X \|_{\psi_1}.
\end{align}
(see the page 92 in \cite{Vershynin}, the proof of Theorem 4.6.1).

\begin{lemma}[Tail Bound on SubGaussian Covariance Matrix]
\label{lem:tailbound}
Let $(E_i, V_i)_{i=1}^N$ be an i.i.d sequence of $K$-subGaussian vectors.  Define
\begin{align}
\label{eq:covmat}
\Sigma = \E VE' \qquad \widehat \Sigma = N^{-1} \sum_{i=1}^N V_i E'_i
\end{align}
Then, with probability at least $1-2 e^{-u^2}$,
\begin{align*}
\| \widehat \Sigma - \Sigma \|  \leq CK^2 \left(\sqrt{\dfrac{d+u}{N}}+ \dfrac{d+u}{N}\right) =: CK^2 \delta.
\end{align*}
As a result,  for any $\ell_N \rightarrow \infty$,  
\begin{align}
\label{eq:sigmatailbound}
\| \widehat \Sigma - \Sigma \|  \lesssim_P \sqrt{(d + \ell_N)/N}.
\end{align}

\end{lemma}

\begin{proof}[Proof of Lemma \ref{lem:tailbound}]

\textbf{Step 1.}  As shown in Corollary 4.2.13 of \cite{Vershynin}, there exist an $1/4$-net $\mathcal{N}$ on the sphere $\mathcal{S}^{d-1}$ with cardinality $| \mathcal{N} | \leq 9^d$. Thus,
\begin{align}
\label{eq:volumetric}
\| A \| = \max_{ \alpha \in \mathcal{S}^{d-1}} | \alpha' A \alpha | \leq 2 \max_{ \alpha \in \mathcal{N}} | \alpha' A \alpha |.
\end{align}
\textbf{Step 2.} For any $\alpha \in \mathcal{N}$,  the random variables $E (\alpha) = \alpha' E$ and $V(\alpha)=\alpha'V$ are $K$-subGaussian random variables. Invoking  \eqref{eq:cauchy} and \eqref{eq:centering}  gives
$$
\| E(\alpha) V(\alpha) - \E E (\alpha) V (\alpha) \|_{\psi_1} \leq C \| E(\alpha) V(\alpha) \|_{\psi_1} \leq CK^2.
$$
Therefore, $\{ E_i(\alpha) V_i(\alpha) - \E E (\alpha) V (\alpha)\}_{i=1}^N$ is an i.i.d, mean zero, and subExponential random variables with its $\psi_1$-norm bounded by $CK^2$.  Take 
$$
\epsilon:= CK^2 \max(\delta, \delta^2)
$$
and invoke Bernstein's inequality with $\delta <1$, noting that 
$$\min (\max(\delta, \delta^2)^2, \max(\delta, \delta^2)) = \delta^2.$$ 
Bernstein's inequality gives
\begin{align*}
\Pr \left( \bigg| N^{-1} \sum_{i=1}^N E_i(\alpha) V_i(\alpha) - \E E (\alpha) V (\alpha) \bigg|  \geq \epsilon/2 \right) &\leq 2 \exp^{-c_1 \min ( \epsilon^2/C^2 K^4, \epsilon/CK^2) N }  \\
&\leq 2 \exp^{-c_1 \delta^2 N} \\
&\leq 2 \exp^{-c_1 (\bar C^2 (N + u^2))}
\end{align*}
for $\bar C$ large enough.
\textbf{Step 3.} For any fixed $\alpha \in \mathcal{S}^{d-1}$,
\begin{align*}
\alpha' (\widehat \Sigma - \Sigma) \alpha &= N^{-1} \sum_{i=1}^N [V_i (\alpha) E_i (\alpha) - \E  V (\alpha) E (\alpha)].
\end{align*}
Invoking \eqref{eq:volumetric} and the union bound over the net $\mathcal{N}$ gives
\begin{align*}
\Pr \left( \| \widehat \Sigma - \Sigma \|  \geq  \epsilon \right) &\leq \Pr ( \max_{\alpha \in \mathcal{N}} | \alpha' ( \widehat \Sigma - \Sigma) \alpha | \geq \epsilon/2) \\
&\leq 9^d 2  \exp^{-c_1 \bar C^2 (N + u^2)}\leq 2 \exp^{-u^2}.
\end{align*}
\end{proof}

\begin{lemma}[Weighted Covariance Matrix]
\label{lem:tailboundboot}
Suppose $ \E \| A(W, \eta_0) \|_F^2$ is finite. Let $\{ e_i \}_{i=1}^N$ be i.i.d Exp (1) draws independent from the data $\{ W_i \}_{i=1}^N$ .  Then,  the following bound holds
\begin{align}
\label{eq:sigma4}
 \| \dfrac{1}{N} \sum_{i=1}^N A(W_i, \eta_0) (e_i - 1) \| = O_P (N^{-1/2}).
\end{align}
\end{lemma}

\begin{proof}[Proof of Lemma \ref{lem:tailboundboot}]
\textbf{ Step 1. } Note that
 \begin{align*}
& \E \| \dfrac{1}{N} \sum_{i=1}^N A(W_i, \eta_0) (e_i - 1) \|_F^2 \\
&= \sum_{k=1}^d \sum_{j=1}^d \E  (\dfrac{1}{N} \sum_{i=1}^N A_{kj} (W_i, \eta_0)  (e_i - 1))^2 \\
&= N^{-1}  \sum_{k=1}^d \sum_{j=1}^d \E A^2_{jk} (W, \eta_0) = N^{-1} \E \| A(W, \eta_0) \|_F^2. 
 \end{align*}
 Invoking Markov inequality gives 
 $$
\| \dfrac{1}{N} \sum_{i=1}^N A(W_i, \eta_0) (e_i - 1) \| \leq \| \dfrac{1}{N} \sum_{i=1}^N A(W_i, \eta_0) (e_i - 1) \|_F = O_P (N^{-1/2}),
$$
which implies \eqref{eq:sigma4}.
 \end{proof}
 
 Lemma \ref{lem:uniderivative} is similar to  Lemma 3  in \cite{CCMS}. For completeness, we provide it below with proof.     Recall that
$$
\psi_0(p) = \E z(p, \eta_0) Y(p, \eta_0) = \E p' V(\eta_0) Y(p, \eta_0)
$$
and
$$
G(p_0) = \E  V(\eta_0) Y(p, \eta_0).
$$
  Let  $ \mathcal{E}_{-}(p,p_0)$ and $ \mathcal{E}_{+}(p,p_0)$ be  
 \begin{align}
       \mathcal{E}_{-}(p,p_0) &= \{ p_0' V(\eta_0)  < 0 < p' V(\eta_0)   \} \label{eq:e-pp01}  \\
      \mathcal{E}_{+}(p,p_0) &= \{ p' V(\eta_0) < 0 < p_0' V(\eta_0)   \}. \label{eq:e+pp01} 
      \end{align}

\begin{lemma}[Uniform Gradient]
\label{lem:uniderivative}
Let $\psi_0: \mathcal{P} \rightarrow \mathrm{R}$ be as defined in \eqref{eq:psi0}. (1) If Assumption \ref{ass:jacobian} holds, the function $\psi_0(p)$ 
has a gradient that is uniformly continuous in $p$. For any $p, p_0 \in \mathcal{P}$, 
\begin{align*}
\psi_0(p) - \psi_0(p_0) = G(p_0)' (p - p_0) +R (p, p_0),
\end{align*}
where  $R(p, p_0) = o(\| p - p_0 \|)$ uniformly over $p, p_0 \in \mathcal{P}$. 

\end{lemma}

\begin{proof}[Proof of Lemma \ref{lem:uniderivative}]
Note that
    \begin{align}
   \label{eq:p0bound}
     \mathcal{E}_{+}(p,p_0) \cup   \mathcal{E}_{-}(p,p_0) &\Rightarrow \bigg\{ 0 < |p_0' V(\eta_0) | < | (p-p_0)' V(\eta_0) | \bigg\} \\
     &\leq \bigg\{ 0 < \dfrac{ |p_0' V(\eta_0) | }{\| \Sigma^{-1/2} V(\eta_0) \| }< \| \Sigma^{1/2} \| \| p-p_0 \| \bigg\}   \nonumber \\
      &\leq \bigg\{ 0 < \dfrac{ |p_0' \Sigma^{1/2} \Sigma^{-1/2} V(\eta_0) | }{\| \Sigma^{1/2} p_0 \| \| \Sigma^{-1/2} V(\eta_0) \| }< (\| \Sigma^{1/2} \|/\| \Sigma^{1/2} p_0 \| ) \| p-p_0 \| \bigg\}   \nonumber \\
     &=:  \mathcal{E}_{+-}(p,p_0).  \nonumber 
    \end{align}
Note that
\begin{align*}
\psi_0(p) - \psi_0(p_0)  &=  \E [p' V(\eta_0) Y(p, \eta_0) - p_0' V(\eta_0) Y(p_0, \eta_0)] \\
&=  \E [p' V(\eta_0) (Y(p, \eta_0) - Y(p_0, \eta_0) )] + G(p_0)'(p-p_0) \\
&=: R(p,p_0) +  G(p_0)' (p - p_0).
\end{align*}
We bound the remainder term $R(p,p_0)$ uniformly over $p, p_0 \in \mathcal{P}$. On the event $ \mathcal{E}_{+-}(p,p_0)$,  
\begin{align*}
\sup_{p, p_0 \in \mathcal{P}} | R(p,p_0) | &\leq \E |p_0' V(\eta_0)| (Y_U - Y_L) 1\{  \mathcal{E}_{+-}(p,p_0) \} \\
&\leq  \E |(p-p_0)' V(\eta_0)| (Y_U - Y_L) 1\{  \mathcal{E}_{+-}(p,p_0) \}.
\end{align*}
Invoking Cauchy inequality gives 
\begin{align*}
\sup_{p, p_0 \in \mathcal{P}} | R(p,p_0) |  &\leq \| p-p_0 \| \E \| V(\eta_0) \| (Y_U - Y_L) 1\{  \mathcal{E}_{+-}(p,p_0) \}  \\
&\leq  \| p-p_0 \| M_{UL} \| V(\eta_0) \|_{P,2} \Pr^{1/2} ( \mathcal{E}_{+-}(p,p_0) ) 
\end{align*}
and Assumption \ref{ass:jacobian} implies
$$
\sup_{p, p_0 \in \mathcal{P}} | R(p,p_0) |  \leq  \| p-p_0 \| M_{UL} \| V(\eta_0) \|_{P,2} C_V  \| p-p_0 \|^{1/2}.
$$
for $C_V$ depending on $\Sigma$ and $\mathcal{C}_V$ large enough. 
\end{proof}

Let $\mathcal{P}$ be as in \eqref{eq:pee}. Define the function classes
 \begin{align*}
 \mathcal{L}_{\eta} &:= \bigg\{ p' V(\eta) , \quad p \in \mathcal{P} \bigg\}\\
\mathcal{I}_{\eta}  &:= \bigg\{  1\{  p' V(\eta) >0  \},\quad p \in \mathcal{P}  \bigg\} \\
 \mathcal{L}_{\eta} \cdot \mathcal{I}_{\eta}  &:= \bigg\{  p' V(\eta) \cdot 1\{  p' V(\eta) >0  \},\quad p \in \mathcal{P}  \bigg\} \\
 \mathcal{R}_{\eta} &:=\{ \quad (p, p' V(\eta)), \quad p \in \mathcal{P} \}
 \end{align*}
 and
 \begin{align*} 
\mathcal{Y}_{\eta}  &:= (Y_U - Y_L)  \cdot \mathcal{L}_{\eta} \cdot \mathcal{I}_{\eta}   = \{ \mathcal{Y} (p, \eta),\quad p \in \mathcal{P}\} \\
\mathcal{G}_{1, \eta} &:= \bigg \{ \E[ \mathcal{Y} (p, \eta) \mid X],\quad p \in \mathcal{P}  \bigg\} \\
\mathcal{F}_{1,2} &= \bigg\{ W \rightarrow p' V(\eta_0) \gamma_0(p, X) , \quad p \in \mathcal{P} \bigg\} \subseteq \mathcal{L}_{\eta_0} \cdot (\gamma_{L,0}(X) +\mathcal{G}_{1,\eta_0}) \\
\mathcal{H}_{\eta}  &= \bigg\{ q' A(W, \eta)  q, \qquad q \in \mathcal{S}^{d-1}  \bigg\}  \\
\mathcal{H}_A &= \bigg\{  G(\Sigma^{-1}q) \Sigma^{-1} ( A(W, \eta_0) - \Sigma) \Sigma^{-1} q, \quad q \in \mathcal{S}^{d-1}  \bigg\} 
\end{align*}
 \begin{lemma}[Entropy Bounds]
\label{lem:maxineq:suppfun}
Suppose Assumptions \ref{ass:sigma}, \ref{ass:jacobian}, \ref{ass:ratebound2}(2), \ref{ass:boundedwidth}--\ref{ass:regularity} hold.  Let $Q$ be any probability measure whose support concentrates on a finite set.   For every function class $ \mathcal{F} \in \{ \mathcal{L}_{\eta} , \mathcal{I}_{\eta} ,  \mathcal{R}_{\eta}, \mathcal{Y}_{\eta} , \mathcal{Y}_{\eta}+Y_L, \mathcal{G}_{1,\eta} , \mathcal{G}_{1,\eta} +\E[ Y_L \mid X], \mathcal{F}_{1,2}, \mathcal{H}_{\eta}, \mathcal{H}_A \}$, there exists an integrable envelope $F$ so that the following bound holds
\begin{align}
\label{eq:entropyboundapp}
 \log \sup_{Q} N(\epsilon \| F \|_{Q,2}, \mathcal{F} , \| \cdot \|_{Q,2}) \lesssim 1+  d \log (1/\epsilon), \quad \text{ for all } 0 < \epsilon \leq 1.
 \end{align}
 \end{lemma}
 \begin{proof}[Proof of Lemma \ref{lem:maxineq:suppfun}]
\textbf{Step 1. }As shown in \cite{andrews:1994b}, the statement holds for the linear class $\mathcal{L}_{\eta}$, the class of indicators $\mathcal{I}_{\eta}$, and  $\mathcal{L}_{\eta} \cdot \mathcal{I}_{\eta}$. Multiplying each element of $\mathcal{L}_{\eta} \cdot \mathcal{I}_{\eta}$ by $Y_U-Y_L$ gives  $\mathcal{Y}_{\eta}$. Therefore, \eqref{eq:entropyboundapp} holds for  the entropy of $\mathcal{Y}_{\eta}$ and  $\mathcal{Y}_{\eta}+Y_L$ , by Lemma L.1 in \cite{BelCherWei} (cf. \cite{andrews:1994b}).  $R_{\eta}$ is a cartesian product of the linear class and the index set $\mathcal{P}$. 

\textbf{Step 2. } Each element of  $\mathcal{G}_{1, \eta}$ is a  conditional on $X$ expectation of an element in $\mathcal{Y}_{\eta}$. By Lemma L.2 from \cite{BelCherWei} (cf. \cite{vdv}), the covering entropy of class $\mathcal{G}_{1, \eta}$ is bounded.   The  class $\mathcal{F}_{1,2}$ is included into the product of $ \mathcal{L}_{\eta_0}$ and
$ \mathcal{G}_{1,\eta_0} + \gamma_{L,0}(X)$ obeying  \eqref{eq:entropyboundapp}. Define the envelope 
\begin{align*}
F_{1,2} :&= C_P \| V(\eta_0) \| (\gamma_{L,0}(X) + M_{UL})
\end{align*}
and note that $\| F_{1,2} \|_{P,c}$ is finite for any $c \geq 2$ since $\| V(\eta_0) \|$ is subGaussian conditional on $X$.   Therefore, \eqref{eq:entropyboundapp} holds for $\mathcal{F}_{1,2}$ with $F_{1,2}$. 

\textbf{Step 3. }  For any two elements of $\mathcal{H}_{\eta}$,
\begin{align*}
| q_1' A(W, \eta) q_1 -  q_2' A(W, \eta) q_2| \leq 2  \| A(W, \eta) \|  \| q_1 - q_2 \|,
\end{align*}
which implies $F^A(W):= 2  \| A(W, \eta) \| $ is a valid envelope function for \eqref{eq:entropyboundapp} to hold. 

\textbf{Step 4. }  Recall  that $G(p) = \E V(\eta_0) (Y_L + \mathcal{Y}(p, \eta_0))$. Therefore, $\| G(p) \| \leq   \E  \| V(\eta_0) \| (Y_L   + M_{UL} )=: C_G$, which is finite by Assumption \ref{ass:regularity}. Then, the function class $$\mathcal{H}_A \subseteq \{ \mu (A(W, \eta_0) - \Sigma) \Sigma^{-1} q, \quad q \in \mathcal{S}^{d-1}, \mu \in \mathrm{R}^d, \quad \| \mu \| \leq C_G \} $$
has a bounded VC index. Each element of $\mathcal{H}_A$ is bounded by $ 2 C_G \|  A(W, \eta_0) - \Sigma \| \| \Sigma^{-1} \|$.

\end{proof}

\begin{lemma}[Entropy Bounds, cont.]
\label{lem:maxineq:suppfun2}
Let $Z(X)$ be a vector of basis functions such that $\| Z(X) \|_{\infty} \leq K_N$ \text{a.s.} for some deterministic sequence $\{ K_N  \geq 1\}$.  Suppose Assumptions \ref{ass:boundedwidth} and \ref{ass:regularity} hold.  Define
\begin{align}
\label{eq:zetapeta}
\zeta(p, \eta) = \mathcal{Y}(p, \eta) - \E[ \mathcal{Y}(p, \eta) \mid X].
\end{align}
For $j \in [p_X]=: \{1,2,\dots, p_X\}$, define
\begin{align*}
\mathcal{F}_{1j}:&= \{ W \rightarrow Z_j(X) (\zeta(p, \eta) - \zeta(p, \eta_0)), \quad p \in \mathcal{P} \} \\
\mathcal{F}_{2j}:&= \{ W \rightarrow Z^2_j(X) (\zeta(p, \eta) - \zeta(p, \eta_0))^2, \quad p \in \mathcal{P} \} \\
\mathcal{F}_{3j}:&= \{ W \rightarrow Z^2_j(X) (\mathcal{Y}(p, \eta) - \mathcal{Y}(p, \eta_0))^2, \quad p \in \mathcal{P} \}  \\
\mathcal{F}_{4j}:&= \{ W \rightarrow Z^2_j(X) (\zeta(p_1, \eta_0) - \zeta(p_2, \eta_0))^2, p_1, p_2 \in \mathcal{P}, \| p_1 - p_2 \| \leq N^{-1} \} 
\end{align*}
and
\begin{align*}
\mathcal{M} &= \bigg\{ W \rightarrow Z(X)' \zeta, \quad \zeta \in \mathrm{R}^{p_X},  \| \zeta \|_0 \leq C s_N,  \| \zeta \| \leq C \bigg\} \\
\mathcal{F}_{1,1} &= \bigg\{ W \rightarrow p' V(\eta_0) \Lambda(Z(X)' \zeta), \quad p \in \mathcal{P}, \zeta \in \mathrm{R}^{p_X}   \bigg\}.
\end{align*}
For  $j \in [p_X]$, for every function class $ \mathcal{F} \in  \{\mathcal{F}_{1j}, \mathcal{F}_{2j}, \mathcal{F}_{3j}, \mathcal{F}_{4j} \}$,  there exists an integrable envelope $F$ so that with $a_N = p_X +N$,
 \begin{align}
\label{eq:entropyboundapp3}
 \log \sup_{Q} N(\epsilon \| F \|_{Q,2}, \mathcal{F} , \| \cdot \|_{Q,2}) \lesssim 1+   \log (a_N/\epsilon), \quad \text{ for all } 0 < \epsilon \leq 1.
 \end{align}
 Furthermore,  for every function class $ \mathcal{F} \in  \{ \mathcal{M}, \mathcal{F}_{1,1}\}$,  there exists an integrable envelope $F$ so that with $a_N = p_X +N$,
 \begin{align}
\label{eq:entropyboundapp2}
 \log \sup_{Q} N(\epsilon \| F \|_{Q,2}, \mathcal{F} , \| \cdot \|_{Q,2}) \lesssim 1+  s_N \log (a_N/\epsilon), \quad \text{ for all } 0 < \epsilon \leq 1.
 \end{align}
\end{lemma}

\begin{proof} [ Proof of Lemma \ref{lem:maxineq:suppfun2}]

\textbf{Step 1. } Note that $0 \leq  \mathcal{Y}(p, \eta) \leq M_{UL} \text{ a.s.}$, which implies $\zeta^2(p, \eta)  \leq M^2_{UL} \text{ a.s. }$ and $(\zeta(p, \eta) - \zeta(p, \eta_0))^2 \leq 2 M^2_{UL} \text{ a.s. }$.
Each class in $ \mathcal{F}_j \in \{ \mathcal{F}_{1j} \mathcal{F}_{2j}, \mathcal{F}_{3j}, \mathcal{F}_{4j} \}$  is derived by multiplying an element in  $\mathcal{Y}_{\eta}-\mathcal{G}_{\eta} $ or   $(\mathcal{Y}_{\eta}-\mathcal{G}_{\eta})^2$
by an  integrable random variable $Z_j(X)$ for $j \in [p_X]$.  Therefore, for each $j \in [p_X]$ and with $F(W) := 4 | Z_j(X) | M_{UL}$ for $\mathcal{F}_{1j}$ and  $F_j(W) := 2Z^2_j(X) M^2_{UL}$ for $\mathcal{F}_j \in \{ \mathcal{F}_{2j}, \mathcal{F}_{3j}, \mathcal{F}_{4j} \}$.
$$
 \log \sup_{Q} N(\epsilon \| F_j \|_{Q,2}, \mathcal{F}_j , \| \cdot \|_{Q,2}) \lesssim 1+   \log (e/\epsilon), \quad \text{ for all } 0 < \epsilon \leq 1.
$$
Hence,
by Lemma L.1(2) in \cite{BelCherWei}, the uniform entropy numbers of $$\mathcal{F} \in \{ \cup_{j \in [p_X]} \mathcal{F}_{1j}, \cup_{j \in [p_X]} \mathcal{F}_{2j}, \cup_{j \in [p_X]} \mathcal{F}_{3j}, \cup_{j \in [p_X]} \mathcal{F}_{4j} \}$$ obeys \eqref{eq:entropyboundapp3}

\textbf{Step 2. }  The class $\mathcal{M} $ is a union over ${p_X\choose C s_N }$ VC-subgraph classes with indices $O( s_N)$. Therefore, \eqref{eq:entropyboundapp2} holds for  $\mathcal{M} $. By Lemma L.3 in \cite{BelCherWei}, the rule holds for $\Lambda(\mathcal{M} )$.

\textbf{Step 3.} The class $\mathcal{F}_{1,1}\subseteq \mathcal{L}_{\eta_0} \cdot \Lambda(\mathcal{M} ) $,  where the class $\mathcal{L}_{\eta_0} $ obeys \eqref{eq:entropyboundapp} and $\mathcal{M}$ obeys \eqref{eq:entropyboundapp2}. Define the envelope function
\begin{align*}
F_{1,1} :&= C_P  \| V(\eta_0) \| K_N \quad \Lambda(t) = t \\
F_{1,1} :&= C_P  \| V(\eta_0) \| \quad \Lambda(t) = \exp t/ (\exp t +1).
\end{align*}
and note that for any $c \geq 2$, $\| F_{1,1} \|_{P,c} \leq C_P K_N \| V(\eta_0) \|_{P,c} < \infty$ since $\| V(\eta_0) \|$ is subGaussian by Assumption \ref{ass:regularity}.  Therefore, $\mathcal{F}_{1,1}$ obeys  \eqref{eq:entropyboundapp2} with $F_{1,1}$.

\end{proof}

\begin{lemma}[Equivalence of Long and Short Definitions of the Partially Linear Parameter]
\label{lem:fromlongtoshort}
The minimizers of \eqref{eq:plppointlong} and \eqref{eq:plppoint1} coincide.
\end{lemma}

\begin{proof}
Fix a random variable $Y$ in a random interval $[Y_L,Y_U]$. For each $b$ in \eqref{eq:plppointlong} we solve for $f(X) = f_b(X)$ as a function of $b$. The solution $f_b(X)$ is a conditional expectation function:
\begin{align*}
    f_b(X) &= \E [Y - D' b\mid X] = \E[Y\mid X] - \eta_0(X)' b.
\end{align*}
Substituting $f_b(X)$ into \eqref{eq:plppointlong}  gives:
\begin{align}
\label{eq:concentrated}
\beta &= \arg \min_{b \in \mathrm{R}^d} \E (Y - \E[Y\mid X] - (D-\eta_0(X))' b)^2.
\end{align}
Expanding $(m+n)^2=m^2+2mn+n^2$ with $m=Y -(D-\eta_0(X))' b$ and $n =\E[Y\mid X]  $ gives:
\begin{align*}
\beta &=^{i} \arg \min_{b \in \mathrm{R}^d} \E (Y - (D-\eta_0(X))' b)^2 \\
&- 2  \E (Y - (D-\eta_0(X))' b) \E [Y\mid X] \\
&+ \E (\E [Y\mid X])^2 \\
&=^{ii} \arg \min_{b \in \mathrm{R}^d} \E (Y - (D-\eta_0(X))' b)^2 +\E (\E [Y\mid X])^2\\
&=^{iii}\arg \min_{b \in \mathrm{R}^d} \E (Y - (D-\eta_0(X))' b)^2.
\end{align*}
Since $\E[(D-\eta_0(X))' b] \E[Y\mid X] = 0$, $ii$ follows. Since the third term does not depend on $b$, $iii$ follows. 
\end{proof}

Recall that 
\begin{align*}
\widehat{\Sigma}_k(\widehat{\eta}_k)&:= \Enk A(W_i, \widehat{\eta}_k), \quad \widehat{\Sigma}(\widehat{\eta}) := \dfrac{1}{K} \sum_{k=1}^K \widehat{\Sigma}_k(\widehat{\eta}_k)
\end{align*}
and the weighted matrix error 
\begin{align*}
\widehat{\Sigma}^v_k(\widehat{\eta}_k)&:= \Enk v_i A(W_i, \widehat{\eta}_k), \quad \widehat{\Sigma}^v(\widehat{\eta}) := \dfrac{1}{K} \sum_{k=1}^K \widehat{\Sigma}^v_k(\widehat{\eta}_k).
\end{align*}

\begin{lemma}[Matrix Concentration]
\label{lem:matrixbern}
Let $\delta_N$ and $A_N$ be as in Definition \ref{def:nearorthog}.  For $v=1$ (regular case) and $v=e$ (bootstrap case), for every partition $k=1,2,\dots, K$, 
\begin{align}
\| \widehat{\Sigma}_k^v (\widehat{\eta}) - \widehat{\Sigma}_k^v (\eta_0)  \|  =  O_P \left(N^{-1/2}\delta_N +  N^{-1 + 1/c'} +A_N\right) = o_P(1). 
\label{eq:sigmabound} 
  \end{align}
Combining the bounds gives
\begin{align*}
\| \widehat{\Sigma}^v (\widehat{\eta}) - \widehat{\Sigma}^v (\eta_0) \|  =  O_P \left(N^{-1/2}\delta_N +  N^{-1 + 1/c'} + A_N\right).
\end{align*}
\end{lemma}

\begin{proof}[Proof of Lemma \ref{lem:matrixbern}]
\textbf{Step 1. } Note that
$$
\widehat{\Sigma}^v (\widehat{\eta}) - \widehat{\Sigma}^v (\eta_0) = \dfrac{1}{K} \sum_{k=1}^K \Enk v_i [A(W_i, \widehat{\eta}_k) - A(W_i, \eta_0)] .
$$
For each $k=1,2,\dots, K$,
\begin{align*}
&\Enk [A(W_i, \widehat{\eta}_k) - A(W_i, \eta_0)] v_i \\
&= n_k^{-1/2} \Gnk  [A(W_i, \widehat{\eta}_k) - A(W_i, \eta_0)] v_i \\
&+ \Sigma (\widehat{\eta}_k) - \Sigma (\eta_0) = I^v_1 (\widehat{\eta}_k)+ I^v_2 (\widehat{\eta}_k).
\end{align*}
Under Assumption \ref{ass:ratebound}, on the event $\mathcal{E}_N$, 
\begin{align*}
\| I^v_2 (\widehat{\eta}_k) \| = \| \Sigma (\widehat{\eta}_k) - \Sigma (\eta_0) \| \leq \sup_{ \eta \in \mathcal{T}_N} \| \Sigma (\eta) - \Sigma (\eta_0) \| \leq A_N.
\end{align*}
We establish the bound on $\|  I^v_1 (\widehat{\eta}_k) \|$ with $v=1$ in Step 2 and $v=e$ in Step 3, respectively.  

\textbf{Step 2. } Consider the function class 
$$
\mathcal{F}_{A}:= \{   \alpha' (A(W, \eta) - A(W, \eta_0)) \alpha, \quad \alpha \in \mathcal{S}^{d-1} \}
$$
and note that
$$
\sup_{\alpha \in \mathcal{S}^{d-1}} | \alpha' (A(W, \eta) - A(W, \eta_0)) \alpha| \leq \| A(W, \eta) - A(W, \eta_0) \|=: F^A (W).
$$
The matrix operator norm $ \| A(W, \eta)  \|$ is integrable in $L_{P,c}$  for $c=2$ and some $c'>2$. Thus, for any $\eta \in \mathcal{T}_N$,
$$
 \| F^A (W) \|_{P,c} \leq 2 \| A(W, \eta)  \|_{P,c} \leq 2 \bar B_A. 
$$
Note that  $\mathcal{F}_{A}$ is a parametric class obeying \eqref{eq:entropyboundapp}.  Assumption \ref{ass:ratebound} implies that, for any $\eta \in \mathcal{T}_N$, 
$$
\sup_{\eta \in \mathcal{T}_N} (\E \| A(W, \eta) - A(W, \eta_0) \|^2)^{1/2}  \leq \delta_N. 
$$

\textbf{Step 3. } Consider the function class 
$$
\mathcal{F}^e_{A}:= \{   e (\alpha' (A(W, \eta) - A(W, \eta_0))) \alpha, \quad \alpha \in \mathcal{S}^{d-1} \}
$$
and note that
$$
\sup_{\alpha \in \mathcal{S}^{d-1}} | e\cdot \alpha'  (A(W, \eta) - A(W, \eta_0)) \alpha| \leq  |e | F^A (W),
$$
where $|e | F^A (W)$ is integrable in $L_{P,c}$  for $c=2$ and some $c'>2$. The function class $\mathcal{F}^e_{A}$ is obtained by multiplying $\mathcal{F}_A$ by an exponential random variable independent of the data, and, therefore, retains the $P$-Donsker and covering properties  $\mathcal{F}_A$.  Finally,  for any $\eta \in \mathcal{T}_N$, 
$$
\sup_{\eta \in \mathcal{T}_N}  \E e^2 \| A(W, \eta) - A(W, \eta_0) \|^2 =  \sup_{\eta \in \mathcal{T}_N} 2 \E \| A(W, \eta) - A(W, \eta_0) \|^2   \leq 2 \delta^2_N. 
$$
Invoking \eqref{eq:maxineqkcc2} in  Lemma \ref{lem:maxineq} with $$\sigma'^2 = \delta^2_N, \quad v=d, \quad a=e$$   conditional on the data $(W_i)_{i \in \mathcal{I}^c_k}$ gives 
\begin{align*}
 \sup_{f \in \mathcal{F}^v_A}  | \Gnk f(W_i) v_i  | \lesssim_P \delta_N + N^{-1/2+1/c'} = o(1). 
\end{align*}
Since $K$ is finite, 
$$
\| \widehat{\Sigma}^v (\widehat{\eta}) - \widehat{\Sigma}^v (\eta_0) \|  =  O_P \left(K (N^{-1/2}\delta_N + N^{-1 + 1/c'} + A_N)\right) = o_P(1).
$$

\end{proof}

\begin{lemma}[Matrix Inverse Linearization]
\label{lem:matrixlinarization}
Under Assumption \ref{ass:ratebound}-\ref{ass:ratebound2}, for $v=1$ (regular case) and $v=e$ (bootstrap case)
\begin{align}
\label{eq:sigmav}
(\widehat{\Sigma}^v (\widehat{\eta}))^{-1}  - \Sigma^{-1}   = -\Sigma^{-1} ( \widehat{\Sigma}^v (\eta_0) - \Sigma ) \Sigma^{-1}  + M^v,
\end{align}
where $\| M^v \| = o_P (N^{-1/2})$.
\end{lemma}

\begin{proof}[Proof of Lemma \ref{lem:matrixlinarization}]
We invoke the identity
 $$(A+B)^{-1}  - A^{-1} = -A^{-1} B (A+B)^{-1}$$ 
with $$A= \Sigma, \quad B=  \widehat{\Sigma}^v (\widehat{\eta}) - \Sigma. $$ Furthermore, we decompose 
$$
B = \widehat{\Sigma}^v (\widehat{\eta}) - \Sigma = (\widehat{\Sigma}^v (\eta_0) - \Sigma) + (\widehat{\Sigma}^v (\widehat{\eta}) - \widehat{\Sigma}^v (\eta_0))  = B^v_1 + B^v_2.
$$
Plugging $A$ and $B = B^v_1+ B^v_2$ gives
\begin{align}
&((\widehat{\Sigma}^v (\widehat{\eta}))^{-1} -  \Sigma^{-1} = -A^{-1} B_1 (A+B)^{-1}  -A^{-1} B_2 (A+B)^{-1}  \\
&= -\Sigma^{-1} ( \widehat{\Sigma}^v (\eta_0) - \Sigma ) ((\widehat{\Sigma}^v (\widehat{\eta}))^{-1}  -  \Sigma^{-1} ( \widehat{\Sigma}^v (\widehat{\eta}) - \widehat{\Sigma}^v (\eta_0) ) ((\widehat{\Sigma}^v (\widehat{\eta}))^{-1}  \label{eq:sigmad} \\
&= -\Sigma^{-1} ( \widehat{\Sigma}^v (\eta_0) - \Sigma ) \Sigma^{-1} \nonumber  \\
 &-  \Sigma^{-1} ( \widehat{\Sigma}^v (\widehat{\eta})  - \widehat{\Sigma}^v (\eta_0) )((\widehat{\Sigma}^v (\widehat{\eta}))^{-1}  \nonumber  \\
&-\Sigma^{-1} (  \widehat{\Sigma}^v  (\eta_0) - \Sigma ) ( (\widehat{\Sigma}^v (\widehat{\eta}))^{-1}  - \Sigma^{-1} ) \nonumber  \\
&= M^v_1  + M^v_2 + M^v_3. \label{eq:sigma3}
\end{align}
By Assumption \ref{ass:sigma}, $ \| \Sigma^{-1} \| \leq \lambda^{-1}_{\text{min}} < \infty$ is bounded. For $v=1$ (regular case), Assumption \ref{ass:ratebound2}  gives
$$
\|  \widehat{\Sigma} (\eta_0) - \Sigma \| = O_P (v_N) = o_P(1). 
$$
For $v=e$ (bootstrap case),  Lemma \ref{lem:tailboundboot} and Assumption \ref{ass:ratebound2} give
\begin{align}
\label{eq:sigma2}
\| B^v_1 \| = \|  \widehat{\Sigma}^v (\eta_0) - \Sigma \| \leq \|  \widehat{\Sigma}^v (\eta_0) -  \widehat{\Sigma} (\eta_0)  \|  + \| \widehat{\Sigma} (\eta_0)  - \Sigma \| = O_P ( v_N + N^{-1/2} ).
\end{align}
In both cases, for $v=1$ and $v=e$,
\begin{align}
\label{eq:mv1}
\| M^v_1   \| \leq \| \Sigma^{-1} \| \|  \widehat{\Sigma}^v (\eta_0) - \Sigma \| \| \Sigma^{-1} \|  = O_P ( v_N + N^{-1/2} ).
\end{align}
As shown in Lemma \ref{lem:matrixbern}, for $v=1$ and $v=e$,
\begin{align}
\label{eq:sigma4}
\| \widehat{\Sigma}^v (\widehat{\eta}) - \widehat{\Sigma}^v (\eta_0) \| = O_P (N^{-1/2} \delta_N + N^{-1+1/c} + N^{-1/2}A_N ) = o_P(1).
\end{align}
Combining \eqref{eq:sigma2} and \eqref{eq:sigma4} gives
\begin{align*}
\| \widehat{\Sigma}^v (\widehat{\eta}) - \Sigma \| \leq \| \widehat{\Sigma}^v (\widehat{\eta}) - \widehat{\Sigma}^v (\eta_0) \|  + \|  \widehat{\Sigma}^v (\eta_0) - \Sigma \| = O_P ( v_N + N^{-1/2} ),
\end{align*}
and $\| \widehat{\Sigma}^v (\widehat{\eta})  \| \geq \lambda_{\text{min}}/2$ wp $1-o(1)$. On the same event,  $\| (\widehat{\Sigma}^v (\widehat{\eta}))^{-1} \|$  is bounded away from $2/\lambda_{\text{min}}$.  Invoking  Assumption  \ref{ass:ratebound} gives
\begin{align}
\label{eq:m2}
&\| M^v_2 \| \leq \| \Sigma^{-1} \| \| \widehat{\Sigma}^v (\widehat{\eta}) - \widehat{\Sigma}^v (\eta_0) \|  \| ( \widehat{\Sigma}^v (\widehat{\eta}))^{-1}  \| \nonumber \\
&= O_P \left( N^{-1/2} \delta_N  + N^{-1 + 1/c'} + N^{-1/2}A_N \right) = o_P(N^{-1/2}).
\end{align}
Definition of $M^v_3$ implies
\begin{align*}
\| M^v_3 \| &\leq  \| \Sigma^{-1} \| \| \widehat{\Sigma}^v (\eta_0)  - \Sigma  \|  \| ((\widehat{\Sigma}^v (\widehat{\eta}))^{-1} - \Sigma^{-1} \|.
\end{align*}
As shown above, $\| ((\widehat{\Sigma}^v (\widehat{\eta}))^{-1}  - \Sigma^{-1} \| \leq \| ((\widehat{\Sigma}^v (\widehat{\eta}))^{-1}  \| + \| \Sigma^{-1} \| =O_P(1)$. Therefore, 
\begin{align}
\label{eq:m3}
\| M^v_3   \| \leq \| \Sigma^{-1} \| \|  \widehat{\Sigma}^v (\eta_0) - \Sigma \| O_P(1) = O_P(v_N + N^{-1/2}).
\end{align}
Collecting the bounds gives
\begin{align*}
\| ((\widehat{\Sigma}^v (\widehat{\eta}))^{-1} -  \Sigma^{-1} \| \leq \sum_{k=1}^3 \| M^v_k \| = O_P (v_N + N^{-1/2}).
\end{align*}
The following bound holds
\begin{align}
\label{eq:m2red}
\| M^v_3 \| &\leq  \| \Sigma^{-1} \| \| \widehat{\Sigma}^v (\eta_0)  - \Sigma  \|  \| ((\widehat{\Sigma}^v (\widehat{\eta}))^{-1} - \Sigma^{-1} \| = O_P ((v_N+ N^{-1/2})^2) = o_P (N^{-1/2}).
\end{align}
As a result,  $\| M^v \|  \leq \| M^v_2 \| + \| M^v_3 \| =  o_P (N^{-1/2})$,  and \eqref{eq:sigmav} holds.
\end{proof}

\begin{proof}[Proof of Corollary \ref{cor:limit}]
Corollary \ref{cor:limit} follows from Theorem \ref{thm:limit} and Steps 4 and 5 of the proof of Theorems 1 and 2 of \cite{CCMS}. Assumptions \ref{ass:ratebound}--\ref{ass:ratebound2} are 
\end{proof}

\begin{proof}[Proof of Corollary \ref{cor:bb}]
Corollary \ref{cor:bb} follows from Theorem \ref{thm:bb} and Steps 4 and 5 of the proof of Theorems 3 and 4 of \cite{CCMS}.
\end{proof}

\renewcommand{\thesection}{B}
\section{Proofs of Section \ref{sec:application} } 

\renewcommand{\theequation}{B.\arabic{equation}}
\renewcommand{\thelemma}{B.\arabic{lemma}}
\renewcommand{\theassumption}{B.\arabic{assumption}}

\setcounter{equation}{0}
\setcounter{section}{0}
\setcounter{lemma}{0}
\setcounter{table}{0}

 \subsection{Proof of Lemmas \ref{lem:riesz} and \ref{lem:rieszlogistic}}

Assumption \ref{lem:rieszappendix} is a restatement of Assumption 6.1 in \cite{Program}. In this subsection, Assumption \ref{lem:rieszappendix} is verified from the primitive conditions on the data generating process. Recall that \eqref{eq:maindecomp}  implies
$$
\mathcal{Y} (p, \eta_0) =: \Lambda (Z(X)' \eta_0(p)) + R_0(p, X) + \zeta (p, \eta_0).
$$
Decompose
\begin{align*}
\mathcal{Y} (p, \eta) &= \E [ \mathcal{Y} (p, \eta) - \mathcal{Y} (p, \eta_0)\mid X] + \E[ \mathcal{Y} (p, \eta_0)\mid X] + \zeta (p, \eta) \\
&=: \Lambda (Z(X)' \eta_0(p)) + R_0(p, X) + R(p, \eta, X) + \zeta (p, \eta).
\end{align*}
Define the following terms
\begin{align*}
A_{11N}(\eta):&= \sup_{p \in \mathcal{P}, j \in [p_X]} |  [\E_N - \E] Z_j^2(X) \mathcal{Y}^2(p, \eta) | \\
A_{12N}(\eta):&= \sup_{p \in \mathcal{P}, j \in [p_X]}   | [\E_N - \E] Z_j^2(X) \zeta^2(p, \eta)|\\
B_N(\eta):&= {\sup}_{\substack{p_1, p_2 \in \mathcal{P}  \\  d_{\mathcal{P}} (p_1, p_2) \leq 1/N \\ j \in [p_X]}}  | [\E_N - \E] Z_j^2(X) (\zeta(p_1, \eta) - \zeta(p_2, \eta))^2 |   \\
G_N(\eta):&= {\sup}_{\substack{p_1, p_2 \in \mathcal{P}  \\  d_{\mathcal{P}} (p_1, p_2) \leq 1/N \\ j \in [p_X]}} | \E Z^2_j(X) (\zeta(p_1, \eta) - \zeta(p_2, \eta))^2 | \\
C_N(\eta):&= {\sup}_{\substack{p_1, p_2 \in \mathcal{P}  \\  d_{\mathcal{P}} (p_1, p_2) \leq 1/N \\ j \in [p_X]}}   | \E_N Z_j(X) (\zeta(p_1, \eta) - \zeta(p_2, \eta)) |\\
D_N (\eta) &=   \sup_{p \in \mathcal{P}, j \in [p_X]}|    [\E_N - \E ]Z_j(X) (\zeta(p, \eta) - \zeta(p, \eta_0)) |\\
E_N(\eta) :&= \sup_{p \in \mathcal{P}, j \in [p_X]} | [\E_N - \E] Z_j^2(X) (\zeta(p, \eta) - \zeta(p, \eta_0))^2 | \\
F_N (\eta) &= \sup_{p \in \mathcal{P}, j \in [p_X]}| [\E_N - \E] Z_j^2(X) (\mathcal{Y} (p, \eta) - \mathcal{Y} (p, \eta_0))^2 | \\
L_N (\eta):&=\sup_{p \in \mathcal{P}, j \in [p_X]} | \E Z_j^2(X) (\zeta(p, \eta) - \zeta(p, \eta_0))^2 |
\end{align*}

In the statement of the following assumption, $\Delta_N = o(1)$ and $\zeta_N = o(1)$  are fixed sequences approaching zero from above at a speed at least polynomial in $N$ (for example, $\zeta_N \geq N^{-c}$ for some $c>0$, and $c_{\zeta}, C_{\zeta}, c_z, C_Z$ are positive finite constants.

    \begin{assumption}[Assumption 6.1, \cite{Program}]
\label{lem:rieszappendix}
The following conditions hold for $N \geq N_0$ and $a_N:=p_X$. (i) The model \eqref{eq:maindecomp} is approximately sparse with $s_{\nu} =s_{\nu}(N)$
\begin{align}
\sup_{ p \in \mathcal{P}} \| \nu_0(p) \|_0 \leq s_N \nonumber \\
\label{eq:mainboundapp} 
\sup_{ p \in \mathcal{P}} (\E (R(p, \eta, X) + R_0(p,X))^2)^{1/2}  \lesssim \bar{\sigma}_N
\end{align}
 (ii) The set $\mathcal{P}$ has uniform covering entropy obeying $ \log N(\epsilon, \mathcal{P}, d_{\mathcal{P}}) \leq d \log (1/ \epsilon) \vee 0$ with $d_{\mathcal{P}} (p_1, p_2) = \| p_1 - p_2 \|$ and the collection
 $(\zeta(p, X), R(p,\eta,X), R_0(p,X))$ are suitably measurable transformations of  $W$ and $p$. (iii) Uniformly over $p \in \mathcal{P}$ and $\eta \in \mathcal{T}_N$,  (a)  the moments of the model are boundedly heteroscedastic, namely  $0<c_{\zeta} \leq \E[ \zeta^2(p, \eta)  \mid X ] \leq C_{\zeta} < \infty \text{ a.s. }$ for some positive finite constants that do not depend on $\eta$ and (b) $\max_{j \in [p_X]} \E[ | Z_j(X) \zeta(p, \eta) |^3 +  | Z_j(X) \mathcal{Y}(p, \eta) |^3 ] \leq C_{\zeta}$.  (iv).  For some sequence $K_N$ such that $K^2_N \bar{\sigma}^2_N \leq  \zeta_N $ and  $\log (a_N) \leq \zeta_N N^{1/3}$,    the dictionary functions, approximation errors, and empirical errors obey the following regularity
conditions: (a) $c_Z \leq \E Z^2_j(X) \leq C_Z$ for $j \in [p_X]$ and $\| Z(X) \|_{\infty} \leq K_N \text{ a.s. }$ and $\max_{j \in [p_X]} \E (Z^4_j(X) + Z^6_j(X)) \leq C_Z$.  (b) With probability $1-\Delta_N$, the following statements hold\\
\begin{align}
\sup_{p \in \mathcal{P}} \E_N (R(p, \eta, X) + R_0(p, X))^2 &\leq C \bar{\sigma}^2_N \label{eq:b10}  \\
A_{11N} (\eta) + A_{12N} (\eta) &\leq \zeta_N \label{eq:b11}    \\
 \log (a_N) ( B_N(\eta) + G_N (\eta))^{1/2} &\leq \zeta_N \label{eq:b12}    \\
C_N(\eta) &= o_P ( \zeta_N N^{-1/2} ) \label{eq:b13}    
\end{align}
(v) With probability $1-\Delta_N$,  the empirical minimum and maximum sparse eigenvalues of the basis covariance matrix are bounded from zero and above.
\end{assumption}

Define the following rates
\begin{align}
 \delta_{1N} = \left(\dfrac{ \log (N p_X K_N)}{N}\right)^{1/2} + \dfrac{K^2_N \log (N p_X K_N)}{N^{2/3}} \label{eq:delta1n}  \\
 \delta_{2N} :=    \left(\dfrac{ \log (N p_X K_N)}{N^{3/2}}\right)^{1/2} + \dfrac{K_N \log (N p_X K_N)}{N^{5/6}}. \label{eq:delta2n} \\
\lambda_{1N} = K^2_N (\sqrt{ \eta_N\log a_N } + N^{-1/2}  \log a_N).  \label{eq:lambda1n}
 \end{align}

\begin{lemma}[Lemma J.5 in \cite{BelCherWei}]
\label{lem:properties}
Let $\zeta(p, \eta_0)$ be as in \eqref{eq:zetapeta}. Suppose that for all $p \in \mathcal{P}$, we have that $\mathcal{Y} (p, \eta_0) = H(Y, p)$, where $Y$ is a random variable and $\{ H (\cdot, p), p \in \mathcal{P}\}$ is a VC-subgraph class of functions bounded by a constant $M_{UL}$ with index $C_Y$ for some constant $C_Y  \geq 1$. Moreover, suppose $Z(X)$ and $\zeta(p, \eta_0)$ obey Assumption \ref{lem:rieszappendix}(iii) and (iv)(a). If Assumption \ref{ass:jacobian} holds, 
\begin{align}
G_N (\eta_0) \lesssim N^{-1/2}.  \label{eq:b6}  
\end{align}
and, with probability at least $ 1 - (\log N)^{-1}$, 
\begin{align}
&  A_{11N} (\eta_0) + A_{12N} (\eta_0)  \leq \delta_{1N} \label{eq:b1} \\
&C_N (\eta_0) \leq  \delta_{2N}  \label{eq:b5}    
\end{align}

\end{lemma}

Lemma  \ref{lem:properties}  is not an exact restatement of Lemma J.5 in \cite{BelCherWei}, but follows from its proof. 

\begin{proof}[Proof of Lemma \ref{lem:properties}]
I verify the conditions of Lemma J.5 in \cite{BelCherWei} with 
$$
q_Z=6, \bar{K}_N := K_N, \nu=1, \ubar{C}:= c_{\zeta} c_Z, \bar{C}:=(1+C_{\zeta}) C_Z.
$$
I show that $\sup_{p_1, p_2 \in \mathcal{P}} \E (\mathcal{Y} (p_1, \eta_0) - \mathcal{Y} (p_2, \eta_0))^4 \leq \bar{C}_P \| p_1 - p_2 \|$ with $\bar{C}_P$ large enough.  Invoking 
$$
|\mathcal{Y} (p_1, \eta_0) - \mathcal{Y} (p_2, \eta_0)| \leq M_{UL} 1\{    \mathcal{E}_{+}(p_1,p_2) \cup   \mathcal{E}_{-}(p_1,p_2) \}
$$
implies
$$
\E |\mathcal{Y} (p_1, \eta_0) - \mathcal{Y} (p_2, \eta_0)|^k \leq M^k_{UL} \Pr ( \mathcal{E}_{+}(p_1,p_2) \cup   \mathcal{E}_{-}(p_1,p_2)) \leq \bar{C}_P \| p_1 - p_2 \|.
$$
Then, the bound \eqref{eq:b1} is established similarly to (J.10), \eqref{eq:b5} similarly to (J.11) and \eqref{eq:b6} is established in the first line (unnumbered display) on page 94 of the \cite{BelCherWei}. 
\end{proof}

\begin{lemma}[Effect of estimation error of $\eta$]
\label{lem:properties1}
(1) The following inequalities hold for $A_N(\eta), B_N (\eta), C_N(\eta)$:
\begin{align}
A_{11N} (\eta) &\leq 2 A_{11N} (\eta_0) +  2 F_N (\eta)   \label{eq:a1n} \\
A_{12N} (\eta) &\leq 2 A_{12N} (\eta_0) +  2 E_N (\eta) \label{eq:a2n} \\ 
B_N(\eta) &\leq 3 ( E_N(\eta) + B_N(\eta_0) + E_N(\eta)) \label{eq:bn} \\
G_N(\eta) &\leq  3 ( L_N(\eta) + G_N(\eta_0) + L_N(\eta))  \label{eq:gn} \\
C_N(\eta) &\leq C_N(\eta_0) + 2 D_N (\eta)  \label{eq:cn} 
\end{align}
(2) The following properties hold for $k=1$ and $k=2$
\begin{align}
\E [ (\zeta(p, \eta) - \zeta(p, \eta_0))^{2k} \mid X] &\leq  C_{k,\zeta} \| \eta(X) - \eta_0(X) \|  \label{eq:c2} \\
\E [ (\mathcal{Y}(p, \eta) -\mathcal{Y}(p, \eta_0))^{2k} \mid X] &\leq  C_{k,Y}  \| \eta(X) - \eta_0(X) \|  \label{eq:c3} 
\end{align}
where $C_{k,\zeta}$ and $C_{k,Y}$ are some finite constants. (3)  The following statements hold
\begin{align}
A_{11N} (\eta) + A_{12N} (\eta) &= O_P (N^{-1/2} (\delta_{1N} +\lambda_{1N})),  \label{eq:c4}  \\
B_N (\eta) &= O_P (N^{-1/2} (\delta_{1N} +\lambda_{1N}))  \label{eq:c5}  \\ 
 C_N (\eta) &= O_P ( \delta_{2N} + N^{-1/2} (\lambda_{1N})) \label{eq:c6}  \\
G_N(\eta) &= O( \eta_N ) \label{eq:c7}
 \end{align}

\end{lemma}

\begin{proof}[Proof of Lemma \ref{lem:properties1}]
Steps 1-2 establish \ref{eq:a1n}--\eqref{eq:cn}. Step 3 verifies (2). Steps 4--6 verifies (3).\textbf{Step 1. } Decomposing
\begin{align*}
(\zeta(p_1, \eta) - \zeta(p_2, \eta)) &=  ((\zeta(p_1, \eta)-\zeta(p_1, \eta_0)) \\
&+ (\zeta(p_1, \eta_0) - \zeta(p_2, \eta_0)) \\
&+(\zeta(p_2, \eta)-\zeta(p_2, \eta_0)).
\end{align*}
For any $f_j(\cdot)$ and $g_j(\cdot), j \in [p_X]$,   $$\sup_{p \in \mathcal{P}, j \in [p_X}  f_j(p) + g_j(p) \leq \sup_{p \in \mathcal{P}, j \in [p_X]} f_j(p) + \sup_{p \in \mathcal{P}, j \in [p_X]} g_j(p).$$ Multiplying $(\zeta(p_1, \eta) - \zeta(p_2, \eta)) $ by $Z_j(X)$ and taking $\sup_{p \in \mathcal{P}, j \in [p_X]} [\cdot]$ gives
\begin{align*}
C_N (\eta) &\leq {\sup}_{\substack{p_1, p_2 \in \mathcal{P}  \\  d_{\mathcal{P}} (p_1, p_2) \leq 1/N \\ j \in [p_X]}}    | \E_N Z_j(X) (\zeta(p_1, \eta) - \zeta(p_1, \eta_0)) | \\
&+C_N (\eta_0) +  {\sup}_{\substack{p_1, p_2 \in \mathcal{P}  \\  d_{\mathcal{P}} (p_1, p_2) \leq 1/N \\ j \in [p_X]}}    | \E_N Z_j(X) (\zeta(p_2, \eta) - \zeta(p_2, \eta_0)) | \\
&\leq 2 D_N (\eta) + C_N(\eta_0),
\end{align*}
which gives \eqref{eq:cn}.

\textbf{Step 2. } Taking squares and multiplying by $Z^2_j(X)$ with $Z^2_j(X) \geq 0$ gives
\begin{align*}
Z^2_j(X) (\zeta(p_1, \eta) - \zeta(p_2, \eta))^2  &\leq  3  Z^2_j(X) ( (\zeta(p_1, \eta)-\zeta(p_1, \eta_0))^2 \\
&+ (\zeta(p_1, \eta_0) - \zeta(p_2, \eta_0))^2 +(\zeta(p_2, \eta)-\zeta(p_2, \eta_0))^2).
\end{align*}
Likewise, 
$$
\zeta^2(p, \eta)  = (\zeta(p, \eta)  - \zeta(p, \eta_0) + \zeta(p, \eta_0))^2 \leq  2 (\zeta(p, \eta)  - \zeta(p, \eta_0))^2  + 2\zeta^2(p, \eta_0).
$$
A similar argument to Step 1 gives  \eqref{eq:a1n} and \eqref{eq:a2n} and \eqref{eq:bn} and \eqref{eq:gn}, which completes (1).

\textbf{Step 3. Verification of (2). } Invoking Assumptions \ref{ass:regularity} and \ref{ass:boundedwidth} and \eqref{eq:pp0} and \eqref{eq:lipbound}  from Lemma \ref{lem:powell} gives \eqref{eq:c3}. \eqref{eq:zetapeta} implies that
$\zeta(p, \eta) - \zeta(p, \eta_0)$ is the demeaned version of $\mathcal{Y}(p, \eta) - \mathcal{Y}(p, \eta_0)$ conditional on $X$.  Therefore, for any $p \in \mathcal{P}$ and for any $\eta \in \mathcal{T}_N$, 
\begin{align*}
\E [ (\zeta(p, \eta) - \zeta(p, \eta_0))^{2}  \mid X] &\leq \E [ (\mathcal{Y}(p, \eta) - \mathcal{Y}(p, \eta_0))^{2}  \mid X] \\
& \leq 2 C_P  M_h M^{2}_{UL}  \eta_N,
\end{align*}
which establishes \eqref{eq:c2}. Next, for any $\eta \in \mathcal{T}_N$, 
\begin{align*}
\E [ (\zeta(p, \eta) - \zeta(p, \eta_0))^{4}  \mid X] &\leq 16 \E [ (\mathcal{Y}(p, \eta) - \mathcal{Y}(p, \eta_0))^{2}  \mid X] \\
& \leq 32 C_P M_h M^{4}_{UL}  \eta_N.
\end{align*}
As a result, for any $\eta \in \mathcal{T}_N$, 
\begin{align}
\label{eq:lneta}
L_N(\eta) \leq  2 C_Z C_P  M_h M^{2}_{UL}  \eta_N
\end{align}

\textbf{Step 4. Verification of \eqref{eq:c4} and \eqref{eq:c6}. } Let $ (N^{-c}) \leq \eta_N$  for some $c>0$ as assumed. I verify the conditions of Lemma \ref{lem:maxineq} with the function class $\mathcal{F}_{1j}, \mathcal{F}_{2j}, \mathcal{F}_{3j}$ in Lemma \ref{lem:maxineq:suppfun2}, all of which obey \eqref{eq:entropyboundapp}. Take
$$
v' =1, \quad a_N = p_X \vee N, \quad \sigma'^2=  M^2_{UL} 2 C_P M_h  K^2_N \eta_N, \quad F(W):= 2 K_N M_{UL}
$$
gives  $$\log (a_N  F(W) / \sigma') =  O ( \log (a_N)) .$$ Likewise for $E_N(\eta)$ and $F_N(\eta)$, take
$$
v' =1, \quad a_N = p_X \vee N, \quad \sigma'^2=  M^4_{UL} 2 C_P M_h  K^4_N \eta_N, \quad F(W):= 4 K^2_N M_{UL}
$$
and $F(W):= 4 K^2_N M^2_{UL}$  gives  $\log (a_N  F(W) / \sigma')  =  O (\log a_N)$.  Thus,
$$
D_N (\eta) +E_N (\eta) + F_N(\eta) = O_P (N^{-1/2} \lambda_{1N}) 
$$
where $\lambda_{1N}$ as in \eqref{eq:lambda1n}. Invoking \eqref{eq:b1} and \eqref{eq:a1n}--\eqref{eq:a2n} gives \eqref{eq:c4}, and invoking \eqref{eq:b5} and \eqref{eq:cn} gives  \eqref{eq:c6}.

\textbf{Step 5. Verification of \eqref{eq:c5} } I verify the conditions of Lemma \ref{lem:maxineq} with $\mathcal{F}_{4j}$.  Note that
\begin{align*}
&\E Z^4_j(X) (\zeta(p_1, \eta_0) - \zeta(p_2, \eta_0))^4 \\
&\leq 16 K^4_N  \E (\mathcal{Y}(p_1, \eta_0) - \mathcal{Y}(p_2, \eta_0))^4 \leq 32 K^4_N M^4_{UL} C_{\delta} N^{-1}
\end{align*}
Invoking Lemma \ref{lem:maxineq} with
$$
v' =1, \quad a_N = p_X \vee N, \quad \sigma'^2=  32 K^4_N M^4_{UL} C_{\delta} N^{-1}
$$
and  $F(W):= K^2_N M^2_{UL}$ gives $\log (a  F(W) / \sigma') = O(\log (a_N)) $
\begin{align}
\label{eq:bneta0}
B_N(\eta_0) &= O \left( N^{-1/2}  \left( \sqrt{ K^4_N N^{-1}\log a_N }+ N^{-1/2} K^2_N\log a_N \right) \right) = O (  N^{-1/2}  \lambda_{1N} ) 
\end{align}
Invoking \eqref{eq:bn} and \eqref{eq:bneta0}  gives \eqref{eq:c5}.

\textbf{Step 6. Verification of \eqref{eq:c7}  }  Invoking \eqref{eq:b6} and  \eqref{eq:lneta} and \eqref{eq:gn} gives \eqref{eq:c7}.

 \end{proof}

\begin{assumption}[Additional assumptions for Lemma \ref{lem:riesz}]
\label{lem:rieszappendix1}
Suppose the conditions of Lemma \ref{lem:riesz} hold. In addition, suppose the following conditions hold on the basis functions $Z(X)$. (1) There exists constant $c_Z \leq C_Z$ and deterministic sequence $\{ K_N, N \geq 1\}$ where $c_Z \leq \E Z^2_j(X) \leq C_Z$ for all $j \in [p_X]$ and  $\| Z(X) \|_{\infty} \leq K_N \text{ a.s. }$ and  $\max_{j \in [p_X]} (\E  Z^6_j(X))^{1/2} \leq C_Z$. The sequences $\delta_{1N} + N^{1/2} \delta_{2N} + \lambda_{1N} + \log^{1/2} a_N ((\eta_N)^{1/2} + N^{-1/4} (\sqrt{\delta_{1N}} +\sqrt{ \lambda_{1N}})) = o(1)$. (v) With probability $1-\Delta_N$,  the empirical minimum and maximum sparse eigenvalues of the basis covariance matrix are bounded from zero and above.
\end{assumption}

\begin{proof}[Proof of Lemma \ref{lem:riesz}]

 \textbf{Step 0.} Define
 \begin{align*}
 R(p, \eta, X):=\E[ \mathcal{Y}(p, \eta) - \mathcal{Y} (p, \eta_0) \mid X].
 \end{align*}
 The function classes
$$
\mathcal{R}_{1,1} := \{ R(p, \eta, X), \quad p \in \mathcal{P} \} 
$$
is derived by taking expectation functions of $\{ \mathcal{Y}(p, \eta) - \mathcal{Y} (p, \eta_0),\quad p \in \mathcal{P} \}  \subseteq (Y_U - Y_L) \cdot (\mathcal{I}_{\eta}-\mathcal{I}_{\eta_0})$, and, therefore, obeys \eqref{eq:entropyboundapp}. The function class 
  $$
\mathcal{R}_{1,2}:=\{ R_0(p, X), \quad p \in \mathcal{P} \} \subseteq \mathcal{G}_{1, \eta_0} - \{ Z(X)' \eta_0(p), \quad p \in \mathcal{P} \}
$$
obeys \eqref{eq:entropyboundapp} because $ p \rightarrow \|\eta_0(p) \|_1$ is assumed Lipschitz in $p$.  Thus, the function class
$$
\{ (R_0(p,X) + R(p, \eta, X))^2, \quad p \in \mathcal{P} \}  \subseteq (\mathcal{R}_{1,1} + \mathcal{R}_{1,2}) \cdot (\mathcal{R}_{1,1} + \mathcal{R}_{1,2})
$$
obeys \eqref{eq:entropyboundapp}  for any fixed $\eta \in \mathcal{T}_N$. 

\textbf{Step 1. (i)--(iii).} Assumptions \ref{lem:rieszappendix} (i)--(ii) are directly assumed in Lemma \ref{lem:riesz}. Note that  (ii) automatically holds for a compact set $\mathcal{P} \in \subset \mathrm{R}^d$ with Eucledian distance. The bound \eqref{eq:c2} implies
\begin{align*}
&\E [(\zeta(p, \eta) - \zeta(p, \eta_0))^2 \mid X] = O( \eta^{\infty}_N). 
\end{align*}
Therefore, the condition (iii) for $\zeta(p, \eta_0)$ in Assumption \ref{lem:rieszappendix1} implies the condition (iii) for $\zeta(p, \eta)$ in Assumption \ref{lem:rieszappendix} with  $c_{\zeta}' = 1/2 c_{\zeta}$ and 
$\bar{C}_{\zeta} = 4 M^2_{UL}$ and $N$ large enough.  Cauchy inequality and Assumption \ref{ass:boundedwidth} gives
$$\max_{j \in [p_X]} \E[ | Z_j(X) \zeta(p, \eta) |^3 +  | Z_j(X) \mathcal{Y}(p, \eta) |^3 ] \leq  2  \max_{j \in [p_X]}  (\E Z^6_j(X))^{1/2} M^3_{UL}. $$ 

\textbf{Step 2. } I verify \eqref{eq:b10}. Note that for $p \in \mathcal{P}$, 
$$
\sup_{p \in \mathcal{P}} | R(p, \eta, X)| \leq M_{UL} \Pr (\mathcal{E}_{+-} (p) \mid X) \leq 2 C_P M_{UL} M_h \| \eta(X) - \eta_0(X) \| \text{ a.s. } 
$$
Therefore, for any $k=1$ and $k=2$, 
$$
\sup_{p \in \mathcal{P}} \E R^k(p, \eta, X) \leq (2 C_P M_{UL} M_h)^k \E \| \eta(X) - \eta_0(X) \|^k.
$$
\begin{align*}
\sup_{p \in \mathcal{P}} \E R_0^{2k} (p, X) \leq \E  (\rho_0(p,X) -  Z(X)' \eta_0(p))^{2k} \leq C \E  (\rho_0(p,X) -  Z(X)' \eta_0(p))^{2}  = O (\bar{\sigma}^2_N),
\end{align*}
which ensures that the approximation error converges at rate
\begin{align}
\label{eq:bias}
\sup_{p \in \mathcal{P}}  \E (R_0(p,X) + R(p, \eta, X))^2 = O( \bar{\sigma}^2_N + \eta^2_N) = O (\bar{\sigma}^2_N).
\end{align}
 Noting that
\begin{align*}
\E (R_0(p,X) + R(p, \eta, X))^4  &\leq 8 \E ( R^4_0(p,X) +  R^4(p,\eta,X))  \\
&= O ( \E \| \eta(X) - \eta_0(X) \|^4 + \bar{\sigma}^2_N) = O (\bar{\sigma}^2_N).
\end{align*}
Invoking Lemma \ref{lem:maxineq} with $$v'=d, \quad \sigma'^2 = C^2_{\sigma} \bar{\sigma}^2_N, \quad F(W) := \sup_{p \in \mathcal{P}} |R_0(p,X)| + M_{UL} $$  gives 
\begin{align}
\label{eq:demeaned}
\sup_{p \in \mathcal{P}} | [\E_N - \E] (R_0(p,X) + R (p, \eta, X))^2 | = O_P( N^{-1/2} \lambda_N ), 
\end{align}
where
\begin{align}
\label{eq:lambdan}
\lambda_N = \sqrt{   C^2_{\sigma}  \bar{\sigma}^2_N \log (1/\bar{\sigma}_N) } +  N^{-1/2}  C_{\sigma} \log (1/\bar{\sigma}_N) 
\end{align}
obeys $N^{-1/2} \lambda_N = O (\bar{\sigma}^2_N )$. Indeed, $\sqrt{\log (1/\bar{\sigma}_N)/N} = O (\sqrt{s_N \log a_N /N}) = O (\bar{\sigma}_N)$. Likewise,
$ N^{-1}  C_{\sigma} \log (1/\bar{\sigma}_N) = O (s_N \log a_N / N) = O (\bar{\sigma}^2_N)$.  Combining \eqref{eq:demeaned} and \eqref{eq:bias} gives \eqref{eq:b10}.
Taking
\begin{align*}
\zeta_N:&= \max (\delta_{1N} + N^{1/2} \delta_{2N} + \lambda_{1N} + \log^{1/2} a_N ((\eta_N)^{1/2} \\
&+ N^{-1/4} (\sqrt{\delta_{1N}} +\sqrt{ \lambda_{1N}})), N^{-1/3} \log a_N) 
\end{align*}
verifies the conditions \eqref{eq:b11} and \eqref{eq:b13} based on  \eqref{eq:c4} and \eqref{eq:c6}, Lemma \ref{lem:properties1}. To verify \eqref{eq:b12}, note that $B(\eta)$ and $G(\eta)$ are non-negative, and
$$
\sqrt{ B(\eta) + G(\eta) } \leq \sqrt{ B(\eta) } + \sqrt{ G(\eta) }  = O_P ( (\eta_N)^{1/2} + N^{-1/4} (\sqrt{\delta_{1N} + \lambda_{1N}})) = O_P (\zeta_N).
$$
Therefore, Assumption \ref{lem:rieszappendix} -- which coincides with Assumption 6.1 in \cite{Program} -- has been verified. Invoking Theorem 6.1 of \cite{Program}  gives the result.

\end{proof}

 \begin{assumption}[Additional assumptions for Lemma \ref{lem:rieszlogistic}]
\label{lem:rieszappendix2}
The following additional assumption hold. There exists a sequence $\rho_N = o(1)$ such that $\sup_{p \in \mathcal{P}} | R_0(p, X) | \leq \rho_N = o(1) \text{ a.s. }$. 
 \end{assumption}

 \begin{proof}[Proof of Lemma \ref{lem:rieszlogistic}]
Define
$$
 R(p, \eta, X) = \E [ \mathcal{M} (p, \eta) - \mathcal{M} (p, \eta_0) \mid X].
$$
 The function class
$$
\mathcal{R}_{1,1} := \{ R(p, \eta, X), \quad p \in \mathcal{P} \} 
$$
is derived by taking expectation functions of $\mathcal{I}_{\eta}$, and, therefore, obeys \eqref{eq:entropyboundapp}. The function class 
  $$
\mathcal{R}_{1,2}:=\{ R_0(p, X), \quad p \in \mathcal{P} \} \subseteq \mathcal{G}_{1, \eta_0} - \{ \Lambda(Z(X)' \eta_0(p)), \quad p \in \mathcal{P} \}
$$
obeys \eqref{eq:entropyboundapp} because $\| \eta_0(p) \|_1$ is assumed Lipschitz in $p$ and $\Lambda(\cdot)$ is Lipshitz (Lemma L.2 and L.3 in \cite{BelCherWei}). 
Invoking Lemma \ref{lem:maxineq} with $$v'=d, \quad \sigma' = C_{\sigma} \bar{\sigma}^2_N, \quad F(W) := 4$$ and  invoking \eqref{eq:demeaned} gives 
$$
\sup_{p \in \mathcal{P}} | [\E_N - \E] (R_0(p,X) + R (p, \eta, X))^2 | = O_P( N^{-1/2} \lambda_N ) =o_P (\bar{\sigma}^2_N),
$$
which verifies (iv)(a) of Assumption \ref{lem:rieszappendix}. The condition (iv) (b) is verified in Lemma \ref{lem:properties1}, where $$\mathcal{Y}(p, \eta):= \mathcal{M}(p, \eta), \quad M_{UL}:=1.$$ Finally,
$$
\sup_{x \in \mathcal{X}} |  R(p, \eta, x)| \lesssim_P \eta^{\infty}_N,
$$
which implies $\sup_{x \in \mathcal{X}} |  R(p, \eta, x) + R_0(p,x) | \lesssim_P \eta^{\infty}_N + \rho_N = o(1)$.  The rest of the proof follows similarly to the proof of Lemma \ref{lem:riesz}.

 \end{proof}

\subsection{Section \ref{sec:ver} Proofs}

 \begin{proof}[Proof of Lemma \ref{cor:powell}]
To cover the cases of Corollaries \ref{cor:plpiv} and \ref{cor:plpiv1}, I take $$\gamma(p, X):= \gamma_L(X) + 1/2 \gamma_{UL} (X), \quad \gamma_N:= 2 (\gamma_{LN} + \gamma_{UL, N})$$ 
in Corollary \ref{cor:plpiv} and
$$\gamma(p,X):=\gamma_{L} (X) + \rho_{0}(p,X), \quad \gamma_N:= \gamma_{LN}$$ in Corollary \ref{cor:plpiv1}.  Recall that
$$
Y(p, \eta) = Y_L + (Y_U - Y_L) 1\{ p' V(\eta) >0 \} = Y_L + \mathcal{Y} (p, \eta).
$$
Define remainder terms 
\begin{align}
I_1 (W,p,\eta):&= p' V(\eta_0) ( Y(p, \eta) -Y(p, \eta_0)) \label{eq:i1peta} \\
I_2 (W,p,\eta):&= p' V(\eta_0) (\log a_N - \gamma(p,X)) \\
I_3 (W,p,\eta) :&= p' (\eta_0(X)-\eta(X) )( Y(p, \eta_0) - \gamma_0(p, X)) \\
I_4 (W,p,\eta):&=  p' (\eta_0(X) - \eta(X))( Y(p, \eta) -Y(p, \eta_0)) \\
I_5(W,p,\eta) :&=p'  (\eta_0(X) - \eta(X)) (\log a_N-\gamma(p,X)). \label{eq:i5peta}
\end{align}
The estimation error of the orthogonal moment can be decomposed as 
\begin{align}
\label{eq:decomp}
 g(W,p,\xi(p)) - g(W,p,\xi_0(p)) = \sum_{j=1}^5 I_j (W,p, \eta).
\end{align}
Likewise, define the remainder terms
\begin{align}
R_1 (W,p,p_0):&=  p_0' V(\eta_0) ( Y(p, \eta_0) - Y(p_0, \eta_0) ) \label{eq:r1pp0} \\
R_2 (W,p,p_0):&=  p_0' V(\eta_0)( \gamma_0(p, X) - \gamma_0(p_0, X) ) \\
R_3 (W,p,p_0) :&= (p-p_0)' V(\eta_0)  ( Y(p_0, \eta_0) - \gamma_0(p_0, X)) \\
R_4 (W,p,p_0):&= (p-p_0)' V(\eta_0)  ( Y(p, \eta_0) - Y(p_0, \eta_0) ) \\
R_5 (W,p,p_0):&=  -(p-p_0)' V(\eta_0) ( \gamma_0(p, X) - \gamma_0(p_0, X) )  \label{eq:r5pp0}
\end{align}
and note that
$$
g(W,p,\xi_0(p)) - g(W,p_0,\xi_0(p_0)) = \sum_{j=1}^5 R_j (W,p,p_0).
$$

  \textbf{ Step 1.  Bound on $\mu_N$. }   The terms $I_1(W,p, \eta)$ and $I_4(W,p, \eta)$ coincide with $B_1(W, p, \eta)$ and $B_2(W, p, \eta)$ in 
   Lemma \ref{lem:powell}, that is
  $$
  I_1(W,p, \eta):=B_1(W, p, \eta), \quad I_4(W,p, \eta):=B_2(W, p, \eta).
  $$
  Invoking \eqref{eq:powell}  from Lemma \ref{lem:powell}  
 \begin{align*}
  \sup_{\eta \in \mathcal{T}_N} |\E [ I_1(W,p, \eta) + I_4(W,p, \eta) ] | &\leq  \sup_{\eta \in \mathcal{T}_N}  4 C^2_P M_{UL} M_h \E \| \eta(X) - \eta_0(X) \|^2 \\
  & \leq 4 C^2_P M_{UL} M_h  \eta^2_N.
 \end{align*}
 The terms $I_2(W,p, \eta)$ and $I_3(W,p, \eta)$ are mean zero by Law of Iterated Expectations
 \begin{align*}
 \E [ I_2(W,p,\eta) + I_3(W,p,\eta) ] =0, \quad \forall p \in \mathrm{R}^d \quad \forall \eta \in \mathcal{T}_N.
 \end{align*}
Finally, for any $\eta \in \mathcal{T}_N$, the term $I_5 (W, p, \eta)$ is bounded as
\begin{align*}
 | \E I_5(W,p, \eta) | &\leq C_P (\E \| \eta(X) - \eta_0(X) \|^2)^{1/2}  (\E | \gamma(p, X) - \gamma_0(p, X) |^2)^{1/2}.
\end{align*}
Combining the bounds give a valid bound on $\mu_N$, that is 
\begin{align}
\label{eq:munbound}
\mu_N \leq 4 C^2_P M_{UL} M_h  \eta^2_N+ C_P \eta_N \cdot \gamma_N.
\end{align}

 \textbf{ Step 2.  Bound on $r_N''$. }  Taking squares of \eqref{eq:decomp} gives 
 $$
\E ( g(W,p,\xi(p)) - g(W,p,\xi_0(p)))^2 = \E (\sum_{j=1}^5 I_j (W,p, \eta) )^2 \leq 5 \sum_{j=1}^5 \E I^2_j (W,p, \eta).
 $$
 We establish the bound on $\E I^2_j(W,p, \eta)$ for each $j=1,2,\dots, 5$. Invoking  \eqref{eq:peta0bound2} in Lemma \ref{lem:powell} and $Y_U - Y_L \leq M_{UL} \text{ a.s. }$  gives the bound on the first term
\begin{align*}
 \E I^2_1 (W,p, \eta) &= \E (p' V(\eta_0))^2 ( Y(p, \eta) -Y(p, \eta_0))^2  \leq M^2_{UL} \E (p' V(\eta_0))^2   1 \{\mathcal{E}_{+-}(p) \}.
 \end{align*}
 On the event $\mathcal{E}_{+-}(p)$, $|p' V(\eta_0)| \leq |p' (\eta_0(X) - \eta(X)|$. As a result, 
\begin{align*}
 \E I^2_1 (W,p, \eta) &\leq M^2_{UL} C^2_P \E \| \eta_0(X) - \eta(X) \|^2 1 \{\mathcal{E}_{+-}(p) \}  \\
 &\leq M^2_{UL} C^2_P \eta^2_N \quad   \forall p \in \mathrm{R}^d \quad \forall \eta \in \mathcal{T}_N.
 \end{align*}
Likewise, 
\begin{align*}
 \E I^2_4 (W,p, \eta) &\leq \E (p' (\eta_0(X) - \eta(X)))^2 (Y_U - Y_L)^2  1 \{\mathcal{E}_{+-}(p) \} \\
  &\leq M^2_{UL} C^2_P \eta^2_N \quad   \forall p \in \mathrm{R}^d \quad \forall \eta \in \mathcal{T}_N.
\end{align*}

 \textbf{ Step 3.  Bound on $r_N''$, cont. } Next, I bound $ \E I^2_2 (W,p, \eta)$ and $ \E I^2_3 (W,p, \eta)$.   Let $U$ be a conditionally on $X$ subGaussian random variable. Then, $\sup_{x \in \mathcal{X}}  \E [U^2 \mid X=x]$ is finite and bounded by some constant, and
\begin{align}
\label{eq:deltau}
\E  \Delta^2 (X) U^2 \leq \sup_{x \in \mathcal{X}} \E [U^2 \mid X=x] \E \Delta^2 (X).
\end{align}
Invoking the bound above with $$\Delta(X)= \log a_N - \gamma(p,X), \quad U = \| V(\eta_0) \|, \quad \sup_{x} \E[ U^2 \mid X=x] \leq \mathcal{C}_V  $$ gives
\begin{align*}
\E I^2_2 (W, p, \eta) &\leq C^2_P \bar C_V \E ( \log a_N - \gamma(p,X))^2  \leq \bar C_V C^2_P \gamma^2_N \quad \forall p \in \mathcal{P} \forall \gamma \in \Gamma_N.
\end{align*}
To bound  $ \E I^2_3 (W,p, \eta)$, we take $ U := Y(p, \eta_0) - \log a_N$ where $\log a_N = \E[ Y (p, \eta_0) \mid X]$. Invoking $$\E[ U^2 \mid X]  = \text{Var} ( Y (p, \eta_0) \mid X) \leq \E[ Y^2 (p, \eta_0) \mid X]$$ and $(a+b)^2 \leq 2 (a^2 + b^2)$ with $a= Y_L $ and $b = (Y_U - Y_L) 1\{ p' V(\eta_0) > 0 \}$ gives
\begin{align*}
\E[ U^2 \mid X]  \leq 2 (\E [ Y^2_L  \mid X] +   M^2_{UL})  \leq 2 (\bar C_L +  M^2_{UL}) =: \bar C_3 \text{ a.s }.
 \end{align*}
 Invoking \eqref{eq:deltau} with $U := Y(p, \eta_0) - \log a_N$ and $\Delta(X) = p' (\eta_0(X)-\eta(X) )$ gives
\begin{align*}
\E I^2_3 (W, p, \eta)  & \leq C^2_P \E \| \eta_0(X) - \eta(X) \|^2  \E [U^2 \mid X] \\
&\leq \bar C_3   C^2_P  \E \| \eta_0(X) - \eta(X) \|^2 \leq \bar C_3   C^2_P  \eta^2_N \quad \forall p \in \mathrm{R}^d \quad \forall \eta \in \mathcal{T}_N.
\end{align*}
Finally,
\begin{align*}
\E I^2_5 (W, p, \eta) &\leq C^2_P \E \| \eta_0(X) - \eta(X) \|^2 ( \log a_N - \gamma(p,X))^2 \\
&\leq C^2_P 4 M^2_{\gamma} \E \| \eta_0(X) - \eta(X) \|^2 \leq 4 M^2_{\gamma} C^2_P \eta^2_N.
\end{align*}
Collecting the bounds gives
\begin{align}
\label{eq:gammanbound}
\sum_{j=1}^5 \E I^2_j (W,p, \eta) \leq C' ( \eta^2_N +  \gamma^2_N)
\end{align}
for $C'$ large enough.

 \textbf{ Step 4. Bound on $r_N'$. }Let $(R_j(p, \eta))_{j=1}^5$ be as in \eqref{eq:r1pp0}-\eqref{eq:r5pp0} and $ \mathcal{E}_{-}(p,p_0),  \mathcal{E}_{+}(p,p_0)$
 as in  \eqref{eq:e-pp0}-\eqref{eq:e+pp0}.  Observe that
 \begin{align}
 \E R_1^2(W,p,p_0) &=\E (p_0' V (\eta_0))^2 (Y_U - Y_L)^2 1{\{   \mathcal{E}_{+-}(p,p_0) \} } \nonumber \\
 &\leq M^2_{UL} \E  ((p-p_0)' V (\eta_0))^2 1{\{   \mathcal{E}_{+-}(p,p_0) \} }  \nonumber \\
 & \leq M^2_{UL} \| p - p_0 \|^2 \E \| V(\eta_0) \|^2
 \end{align}
Note that
 \begin{align*}
| \gamma_0(p, X) - \gamma_0(p_0, X) |&= | \E  [ (Y_U - Y_L) 1\{   p' V(\eta_0)>0  \} \mid X] \\
& -  \E  [ (Y_U - Y_L) 1\{   p_0' V(\eta_0)>0  \} \mid X]  | \\
&\leq \E [ (Y_U - Y_L) | 1\{   p' V(\eta_0)>0  \} - 1\{   p_0' V(\eta_0)>0  \}  | \mid X] \\
&\leq \E [ (Y_U - Y_L) 1\{   \mathcal{E}_{+-}(p,p_0)   \} \mid X] \\
&\leq M_{UL} \Pr (  \mathcal{E}_{+-}(p,p_0) \mid X)  \text { a.s. }
 \end{align*}
 Taking the squares of each side gives 
 $$
 ( \gamma_0(p, X) - \gamma_0(p_0, X))^2 \leq M^2_{UL} \Pr^2 (  \mathcal{E}_{+-}(p,p_0) \mid X) \leq M^2_{UL} \Pr (  \mathcal{E}_{+-}(p,p_0) \mid X) \text{ a.s. },
 $$
 where the last bound follows from $t^2 \leq t$ if $t \in [0,1]$.  Invoking \eqref{eq:deltau} with $U:=   \| V (\eta_0) \| $ and $\Delta(X):=\gamma_0(p, X) - \gamma_0(p_0, X)$ gives
  \begin{align*}
 \E R_2^2(W,p,p_0) &=\E (p_0' V (\eta_0))^2  (\gamma_0(p, X) - \gamma_0(p_0, X))^2 \nonumber \\
 &\leq M^2_{UL} C^2_P \bar C_V  \E_{X}  [\Pr (  \mathcal{E}_{+-}(p,p_0) \mid X)] \\
 &= M^2_{UL} C^2_P \bar C_V  \Pr (  \mathcal{E}_{+-}(p,p_0) ) \\
 &\leq M^2_{UL} \bar C_V  \| p - p_0 \|.
 \end{align*}
Cauchy inequality gives a bound $\E R^2_3(W, p, p_0) $:
  \begin{align*}
\E R^2_3(W, p, p_0) &\leq C^2_P  \| p- p_0 \|^2 \E \| V(\eta_0) \|^2 ( Y(p_0, \eta_0) - \gamma_0(p_0, X))^2 \\
&\leq C^2_P \| p- p_0 \|^2 \| V(\eta_0) \|_{P, 4} \sup_{p_0 \in \mathcal{P}} \| Y(p_0, \eta_0) - \gamma_0(p_0, X) \|_{P, 4},
 \end{align*}
 where $ \| V(\eta_0) \|_{P, 4} < \infty$ follows from subGaussianity and $\| Y(p_0, \eta_0) - \gamma_0(p_0, X) \|_{P, 4} \leq \| Y(p_0, \eta_0)  \|_{P, 4} + M_{\gamma}  \leq \| Y_L \|_{P,4} + M_{UL} + M_{\gamma}$ is finite by assumption.   Finally, for $j=4 $ and $j=5$, the term $R_j(W, p, p_0)$ is a product of $( p- p_0)' V(\eta_0)$ and a random variable bounded by $M_{UL}$ a.s. Therefore, 
 $$
 \E R^2_j (W, p, p_0) \leq C^2_P M^2_{UL} \| p- p_0 \|^2  \E \| V(\eta_0) \|^2, \quad j=4,5.
 $$
Collecting the bounds shows that
\begin{align}
\label{eq:bound}
\sup_{p, p_0 \in \mathcal{P}, \| p - p_0 \| \leq \tau_N } \sum_{j=1}^5 \E R^2_j (W,p, p_0) \leq C' (\tau_N + \tau^2_N)
\end{align}
for $C'$ large enough.

 \textbf{ Step 5. Bound on $A_N$. } Define
\begin{align*}
M_1(W,\eta, \eta_0, m, m_0) :&= (\eta_0 (X) - \eta(X))(D -m_0(X))' \\
M_2(W,\eta, \eta_0, m, m_0) :&= (Z-\eta_0(X))(m_0(X)-m(X))' \\
M_3(W,\eta, \eta_0, m, m_0) :&=  (\eta_0 (X) - \eta(X)) (m_0(X)-m(X))'
\end{align*}
and note that 
\begin{align*}
A(W, \eta,m) - A(W, \eta_0,m_0) &= (Z - \eta(X)) (D - m(X))' - (Z - \eta_0(X)) (D - m_0(X))' \\
&= \sum_{k=1}^3 M_k(W, \eta, \eta_0, m, m_0).
\end{align*}
The terms $M_1(W,\eta, \eta_0, m, m_0)$ and $M_2(W,\eta, \eta_0, m, m_0) $ are zero mean  by LIE. The second-order term $M_3(W,\eta, \eta_0, m, m_0)$ is bounded as
\begin{align*}
 \| \E M_3(W,\eta, \eta_0, m, m_0) \| &\leq \E \| M_3(W,\eta, \eta_0, m, m_0) \| \\
 &\leq \E \| \eta_0 (X) - \eta(X) \| \| m_0 (X) - m(X) \| \\
&\leq (\E \| (\eta_0 (X) - \eta(X)) \|^2)^{1/2} (\E \| (m_0 (X) - m(X)) \|^2)^{1/2}.
\end{align*}
Collecting the bounds ensures that the condition \eqref{eq:an} holds with $A_N \leq  \eta_N \cdot m_N$. 

 \textbf{ Step 6. Bound on $\delta_N$. } 
\begin{align*}
\E  \| M_1(W,\eta, \eta_0, m, m_0)  \|^2 &\leq \E  \| M_1(W,\eta, \eta_0, m, m_0)  \|^2_F = \E \| \eta_0 (X) - \eta(X) \|^2 \| D -m_0(X) \|^2 \\
&\leq \sup_{x} \E [\| D -m_0(X) \|^2 \mid X=x]  \E \| \eta_0 (X) - \eta(X) \|^2  \\
&=: \bar C_D \E \| \eta_0 (X) - \eta(X) \|^2 .
\end{align*}
A similar argument gives
\begin{align*}
\E  \| M_2(W,\eta, \eta_0, m, m_0)  \|^2_F \leq  \bar C_V \E \| m_0 (X) - m(X) \|^2 .
\end{align*}
Finally,
\begin{align*}
\E  \| M_3(W,\eta, \eta_0, m, m_0)  \|^2_2 &\leq \E \| m_0 (X) - m(X) \|^2 \| \eta_0 (X) - \eta(X) \|^2 \leq 4 M^2_{\eta} \eta^2_N.  
\end{align*}
Collecting the bounds gives
\begin{align*}
\| A(W, \eta,m) - A(W, \eta_0,m_0)  \|_2 \leq \sum_{k=1}^3 \| M_k(W, \eta, \eta_0, m, m_0) \|_2
\end{align*}
which implies
\begin{align*}
\E \| A(W, \eta,m) - A(W, \eta_0,m_0)  \|^2_2 \leq 3 \sum_{k=1}^3 \E \| M_k(W, \eta, \eta_0, m, m_0) \|^2_2,
\end{align*}
which ensures that the condition \eqref{eq:an} holds with $$\delta^2_N \leq 3(4 M^2_{\eta}  \eta^2_N + \bar C_D \eta^2_N +   \bar C_V m^2_N  ).$$

 \textbf{ Step 7. (Conclusion)} Steps 1-6 verify Assumption \ref{ass:ratebound}. Take $A(W, \eta_0,m_0) = (V-\eta_0(X))  (D- m_0(X))'$. The bound on $\E \| A(W, \eta_0 ,m_0) \|_F^2$ is 
\begin{align*}
& \E \| A(W, \eta_0,m_0) \|_F^2 = \E \| Z - \eta_0(X) \|^2 \| D - m_0(X) \|^2  \\
& \leq \sqrt{\E \| Z-\eta_0(X) \|^4} \sqrt{\E \| D-m_0(X) \|^4} = \sqrt{\E \| V \|^4 \E \| E \|^4 } < \infty,
 \end{align*}
  since $\|V \| = \| V(\eta_0) \| $ and $\|E \| = \| E(m_0)\|$ are subGaussian. 
  \end{proof}
  
   \begin{proof}[Proof of Corollary \ref{cor:plpiv}]
Note that Assumptions \ref{ass:ratebound} and \ref{ass:ratebound2} are verified in Lemmas \ref{cor:powell} and \ref{cor:powell2}, respectively. Below, I verify Assumption \ref{ass:concentration:chap1}. Let $\mathcal{L}_{\eta} $ and $\mathcal{Y}_{\eta}$ and $ \mathcal{G}_{1,\eta}$ be the function classes  defined in Lemma 
 \ref{lem:maxineq:suppfun}. The following classes 
$$
\{ p' V(\eta) \cdot Y(p, \eta), \quad p \in \mathcal{P} \} \subseteq \mathcal{L}_{\eta} \cdot (Y_L + \mathcal{Y}_{\eta})
$$
and
$$
\{ p' V(\eta) \cdot  (\gamma_{L} (X) + 1/2 \gamma_{UL} (X)) ,  \quad p \in \mathcal{P} \} \subseteq \mathcal{L}_{\eta} \cdot  (\gamma_{L} (X) + 1/2 \gamma_{UL} (X))
$$
are derived from $\mathcal{L}_{\eta}, Y_L + \mathcal{Y}_{\eta}$ and integrable random variables $\gamma_L(X), \gamma_{UL} (X)$  whose  uniform covering entropies obey \eqref{eq:entropyboundapp}.   For every fixed $\eta$ and $\gamma_L$,  the function class
\begin{align*}
     \mathcal{F}_{\eta, \gamma_L, \gamma_{UL}} := \{ W \rightarrow  p' V(\eta) (Y(p, \eta) -  \gamma_{L} (X) -  1/2 \gamma_{UL} (X)), \quad p \in \mathcal{P} \}
\end{align*}
obeys \eqref{eq:entropyboundapp} with the envelope function 
\begin{align*}
F_{\eta, \gamma_L, \gamma_{UL}} &= C_P (\| V(\eta_0) \| + M_{\eta}) (Y_L  -  \gamma_{L,0} (X)) \\
&+ C_P (\| V(\eta_0) \| + M_{\eta}) (M_{\gamma} + M_{\gamma_{UL}}  + M_{UL}),
\end{align*}
whose $\| F_{\eta, \gamma_L, \gamma_{UL}}  \|_{P,c'}$ is integrable for $c'=2$ and $c' > 2$ by Assumption \ref{ass:regularity}. 

\end{proof}

\begin{proof}[Proof of Corollary \ref{cor:plpiv1}]
\textbf{Step 1. }  Define the first-order term 
$$
\bar{S}^v_{1} (p) := \sqrt{N}  \E_N p' V_i (\eta_0) ( \rho_0 (p, X_i) - \widehat \rho (p, X_i))  v_i
$$
and the second-order term
$$
\bar{S}^v_{2} (p) := \sqrt{N}  \E_N p' (V_i (\widehat \eta) - V_i (\eta_0)) ( \rho_0 (p, X_i) - \widehat \rho (p, X_i)) v_i
$$
and the cross-fit term
$$
\bar{S}^v_{3} (p) :=  \sqrt{N}  \E_N (g(W_i, p, \{ \widehat \eta, \widehat \gamma_L,  \rho_0(p, \cdot) \} ) - g(W_i, p_0(q), \{ \eta_0, \gamma_{L,0},  \rho_0(p, \cdot) \} )) v_i.
$$
Decompose the estimation error
\begin{align*}
&\sqrt{N} (\E_N g(W_i, \widehat p (q), \widehat \xi (\widehat p(q)) ) - \E_N g(W_i, p_0(q), \xi_0 (p_0(q)))) v_i \\
&= \bar{S}^v_{1} (\widehat p(q))+\bar{S}^v_{2} (\widehat p(q))+\bar{S}^v_{3} (\widehat p(q)).
\end{align*}
I show that $\sup_{p \in \mathcal{P}}  |\bar{S}^v_{1} (p)|= o_P(1)$ in Steps 2 and 3. Step 4 and Step 5 elaborates on $\sup_{p \in \mathcal{P}}  |\bar{S}^v_{2} (p)|= o_P(1)$ and $\bar{S}^v_{3} (\widehat p(q))$.

\textbf{Step 2.}   Let $\mathcal{L}_{\eta} $ and $\mathcal{Y}_{\eta}$ and $ \mathcal{G}_{1,\eta}$ and $\mathcal{M}$ be the function classes  defined in Lemma 
 \ref{lem:maxineq:suppfun}.  Take $a_N = p_X \vee N$ and $\bar{\sigma}_N = C \sqrt{s_N \log a_N / N}$.  Define the nuisance realization set
\begin{align*}
\mathcal{T}_p:= \{ \rho_0(p, \cdot) \} \cup \{ \| \eta \|_0 \leq C s_N, \|  \eta - \eta_0(p)  \| \leq  \bar{\sigma}_N,    \|   \eta - \eta_0(p)  \|_1 \leq \sqrt{s_N} \bar{\sigma}_N, \eta \in \mathrm{R}^{p_X} \}.
\end{align*}
The class
$$
\mathcal{F}_{2,1} = \{ W \rightarrow p' V(\eta_0) \cdot \Lambda (Z(X)' \eta), \quad p \in \mathcal{P}, \eta \in \mathcal{T}_p \setminus \{ \rho_0(p, \cdot)  \} \} \subseteq \mathcal{L}_{\eta} \cdot \Lambda(\mathcal{M})
$$
obeys \eqref{eq:entropyboundapp2} with $F_{2,1} :=C_P \| V(\eta_0) \| K_N $ (linear link) or $F_{2,1} :=C_P\| V(\eta_0) \| $ (logistic link), and the class
$$
\mathcal{F}_{2,2} =  \{ W \rightarrow  p' V(\eta_0) \cdot \rho_0(p,X) ,  \quad p \in \mathcal{P} \} \subseteq \mathcal{L}_{\eta} \cdot \mathcal{G}_{1,\eta_0}
$$
obeys \eqref{eq:entropyboundapp} with $F_{2,2} :=C_P M_{UL} \| V(\eta_0) \| $.   The function class
$$
\mathcal{F}_{2,3}:=\{ W \rightarrow p' V(\eta_0) \cdot (\Lambda (Z(X)' \eta) - \rho_0(p,X)), \quad p \in \mathcal{P} \}  \subseteq \mathcal{F}_{2,1}- \mathcal{F}_{2,2}
$$
obeys \eqref{eq:entropyboundapp2} with $F_{2,3}:= F_{2,1} + F_{2,2}$. For each $\mathcal{F} \in \{ \mathcal{F}_{2,1}, \mathcal{F}_{2,2}, \mathcal{F}_{2,3} \}$, the class $\mathcal{F}^e$ is obtained by multiplying the elements of $\mathcal{F}$ by an integrable random variable, and therefore retains the entropy properties of $\mathcal{F}$.  Finally,  wp $1-o(1)$, for $v=1$ (regular case) and $v=e$ (bootstrap case)
$$
 \sup_{p \in \mathcal{P}}  |\bar{S}^v_{1} (p)| = \| \G_N f^v(W) \|_{\mathcal{F}}.
$$
\textbf{Step 3.} For both linear and logistic link, $\sup_{t} | \Lambda' (t) | \leq 1$. By intermediate value theorem,
\begin{align*}
|\Lambda (Z(X)' \eta (p)) -  \Lambda (Z(X)' \eta_0 (p)) | &\leq \sup_{t} | \Lambda' (t) |  |Z (X)' (\eta (p) -\eta_0(p))| \\
 &\leq |Z (X)' (\eta (p) -\eta_0(p))|.
\end{align*}
Next, for any $p \in \mathcal{P}$.
\begin{align*}
&\E (\Lambda(Z(X)' \eta(p)) - \rho_0(p,X))^2  \leq 2 \E ( Z(X)' (\eta(p) - \eta_0(p)) )^2 + 2 \E R_0^2(p, X) \\
& \lesssim s_N \log a_N/N \equiv \bar{\sigma}_N^2
\end{align*}
Notice that
\begin{align*}
\sup_{f \in \mathcal{F}} \E f^2 (W) \leq C_P \sup_{x \in \mathcal{X}} \E [\| V(\eta_0) \|^2 \mid X=x] \E (\Lambda(Z(X)' \eta(p)) - \rho_0(p,X))^2 \lesssim \bar{\sigma}^2_N
\end{align*}
Invoking Lemma \ref{lem:maxineq} with 
$$
a_N =p_X \vee N, \quad v' = s_N, \quad \sigma' = C_{\sigma} \bar{\sigma}_N, \quad F_{2,3} (W):=F_{2,1} + F_{2,2} 
$$
gives for $\Lambda(t)=t$
\begin{align*}
 \sup_{p \in \mathcal{P}}  |\bar{S}^v_{1} (p)| \lesssim_P   \sqrt{s_N  \bar{\sigma}^2_N \log (a_N K_N /\bar{\sigma}_N)}  + s_N N^{-1/2 + 1/c' } \log (a_N K_N /\bar{\sigma}_N) =  o_P(1)
\end{align*}
and for $\Lambda(t) = \exp t/(\exp t+1)$
\begin{align*}
 \sup_{p \in \mathcal{P}}  |\bar{S}^v_{1} (p)| \lesssim_P   \sqrt{s_N  \bar{\sigma}^2_N \log (a_N /\bar{\sigma}_N)}  + s_N N^{-1/2 + 1/c' } \log (a_N /\bar{\sigma}_N) =  o_P(1)
\end{align*}

\textbf{Step 4.}    I show that $\sup_{p \in \mathcal{P}}  |\bar{S}^v_{2} (p)|= o_P(1)$.   By the argument in Step 3, 
$$
(\Lambda (Z(X_i)' \widehat \eta (p)) -  \Lambda (Z(X_i)' \eta_0 (p)))^2 \leq (Z (X_i)' (\widehat \eta (p) -\eta_0(p)))^2
$$
and summing over $i=1,2,\dots, N$ gives
$$
 \sup_{p \in \mathcal{P}}  \E_N (\Lambda (Z(X_i)' \widehat \eta (p)) -  \Lambda (Z(X_i)' \eta_0 (p)) )^2 \leq   \sup_{p \in \mathcal{P}}  \| \widehat \eta(p) - \eta_0(p) \|^2_{P_N,2}
$$
and
$$
\sup_{p \in \mathcal{P}}  \E_N  ( \rho_0 (p, X_i) - \widehat \rho (p, X_i))^2 \leq   \sup_{p \in \mathcal{P}}  2 \| \widehat \eta(p) - \eta_0(p) \|^2_{P_N,2} + N^{-1} 2 \E_N  R^2_0(p, X_i).
$$
As assumed in \eqref{eq:fsstage}, the RHS above is $O_P (\bar{\sigma}^2_N)$. By Assumption \ref{ass:regularity},
\begin{align*}
\sup_{p \in \mathcal{P}} ( \E_N v^2_i (p' (V_i (\widehat \eta) - V_i (\eta_0)) )^2 )^{1/2} \leq C_P (\E_N v^2_i \| \widehat \eta (X_i) - \eta_0(X_i) \|^2)^{1/2} = O_P (\eta_N + N^{-1/4}).
\end{align*}
As a result,
\begin{align*}
&\sup_{p \in \mathcal{P}} | \bar{S}^v_{2} (p)| \\
&=  \sup_{p \in \mathcal{P}} N^{1/2} | \E_N p' (V_i (\widehat \eta) - V_i (\eta_0))   ( \rho_0 (p, X_i) - \widehat \rho (p, X_i)) v_i | \\
&= \sup_{p \in \mathcal{P}} N^{1/2} (\E_N  ( \rho_0 (p, X_i) - \widehat \rho (p, X_i))^2)^{1/2}  ( \E_N v^2_i (p' (V_i (\widehat \eta) - V_i (\eta_0)) )^2 )^{1/2}   \\
&= o_P(N^{1/2} \bar{\sigma}_N \cdot (\eta_N + N^{-1/4})) =o_P(1).
\end{align*}

\textbf{Step  5.}  I show that $\sup_{p \in \mathcal{P}}  |\bar{S}_{3} (\widehat{p}(q)) - \G_N [h(W, q)]|= o_P(1)$. This follows from the Steps 1--5 of the proof of Theorems \ref{thm:limit}, where 
Assumptions \ref{ass:ratebound} and \ref{ass:ratebound2} and \ref{ass:concentration:chap1} need to be verified. Assumptions \ref{ass:ratebound} and \ref{ass:ratebound2} are verified in Lemmas \ref{cor:powell} and \ref{cor:powell2}, respectively. Assumption \ref{ass:concentration:chap1} is verified by the same argument as the proof of Corollary \ref{cor:plpiv} with 
$$
\{ p' V(\eta) \cdot Y(p, \eta), \quad p \in \mathcal{P} \} \subseteq \mathcal{L}_{\eta} \cdot (Y_L + \mathcal{Y}_{\eta})
$$
and
$$
\{ p' V(\eta) \cdot  (\gamma_{L} (X) + \rho_0(p,X)) ,  \quad p \in \mathcal{P} \} \subseteq \mathcal{L}_{\eta} \cdot  (\gamma_{L} (X) + \mathcal{G}_{1, \eta_0}),
$$
where both classes obey \eqref{eq:entropyboundapp}. The envelope function
$$
F(W):= (\| V(\eta_0) \| + M_{\eta}) (Y_L - \gamma_{L, 0} (X) + M_{\gamma} + M_{UL})
$$
is integrable by Assumption \ref{ass:regularity}. 

\end{proof}

 \subsection{Proof of Section \ref{sec:apd}}

\begin{proof}[Proof of Lemma \ref{lem:apd}]
The identified set can be written as
$$
\mathcal{B} = \{ \beta: \beta = \E V(\eta_0) \gamma (D, X) \},
$$
where
\begin{align}
\label{eq:bound}
\Pr ( \gamma_L (D, X) \leq \gamma (D, X) \leq \gamma_U (D, X)) = 1
\end{align}
The convexity argument is similar to \cite{Kaido}. For any $\beta_1$ and $\beta_2$, there exist $\gamma_1(D, X)$ and $\gamma_2(D,X)$ obeying \eqref{eq:bound} 
such that 
$
\beta_j = \E V(\eta_0) \gamma_j (D, X), \quad j=1, 2.
$
Therefore, for any $\alpha \in [0,1]$, $$\alpha \beta_1 + (1-\alpha) \beta_2 =  \E V(\eta_0) \gamma_{\alpha} (D, X),$$
where $\gamma_{\alpha} (D, X) = \alpha \gamma_1 (D, X)+ (1-\alpha)  \gamma_2 (D, X)$ also obeys \eqref{eq:bound}.   
To show strict convexity, invoke  Lemma \ref{lem:uniderivative} with $\Sigma = I_d$ gives 
$$
\sigma(q) - \sigma(q_0) = G(q_0)' (q-q_0) + R(q,q_0), \quad R(q,q_0) = o ( \| q -q_0 \|).
$$
 By Corollary 1.7.3 in Schneider (1993), the set $\mathcal{B}$ is strictly convex. Next, I show that the set $\mathcal{B}$ is bounded. For each $j=1,2,\dots, d$, take $l_j(D, X):= \nabla_j f_0(D \mid X)/ f_0(D \mid X)$. Cauchy inequality implies
\begin{align*}
| \beta_j | \leq \E | \gamma (D, X) l_j(D, X) | \leq \bar C_{\gamma} (\E \| l (D, X) \|^2)^{1/2} \leq C_{\gamma} C_L. 
\end{align*}
Finally, the set $\mathcal{B}$ contains its boundary by the argument of  the proof of Proposition 2 in \cite{BMM} and \cite{BM} and the proof of Theorem 2.1 in \cite{Kaido}, page 23. 

\end{proof}

\begin{proof}[Proof of Lemma \ref{lem:bias2}]
For the Example \ref{ex:apd}, recall that
$$
 \mathcal{E}_{+-}(q) = \{    0 < |q' V ( \eta_0) | \leq |q' (V ( \eta) - V ( \eta_0)) |  \}, \quad p:=q, \quad \mathcal{P} := \mathcal{S}^{d-1}, \quad C_P=1.
 $$
Let
$B_1(W, \eta, q )$ and $B_2(W, \eta, q )$ be the bias terms  as defined in Lemma \ref{lem:powell}.  Invoking \eqref{eq:peta0bound3} and \eqref{eq:peta0bound4} in Lemma \ref{lem:powell} gives 
\begin{align}
\label{eq:pp0}
Y(q, \eta) -Y(q, \eta_0)=(Y_U - Y_L) 1\{  \mathcal{E}_{+}(q) \cup \mathcal{E}_{-}(q)\} 
\end{align}
and
\begin{align*}
| \E  B_j(W, \eta, q ) | \leq  M_{UL} \E \| V ( \eta) - V (\eta_0) \| 1\{  \mathcal{E}_{+-}(q) \}, \quad j=1,2.
\end{align*}
Assumption \ref{ass:margin} and $\sup_{\eta \in \mathcal{T}_N} \| V ( \eta) - V (\eta_0) \| \leq \eta^{\infty}_N$ gives
$$
\sup_{\eta \in \mathcal{T}_N} \E \| V ( \eta) - V (\eta_0) \| 1\{  \mathcal{E}_{+-}(q) \} \leq \eta^{\infty}_N \Pr (\mathcal{E}_{+-}(q)) \leq \bar C_f (\eta^{\infty}_N)^2.
$$

\end{proof}

\begin{proof}[Proof of Corollary \ref{cor:apd}]
\textbf{Step 1 (Outline). } Assumption \ref{ass:ratebound2} holds trivially with known $\Sigma = I_d$ and $v_N = 0$. Therefore, Assumption \ref{ass:ratebound} can be invoked with $$A_N = \delta_N = \tau_N = r_N' = 0, \quad p=q \in \mathcal{S}^{d-1}.$$ Recall that
$$
\gamma_N = \gamma_{L,N} + \gamma_{UL,N}  + \gamma^1_{L,N} +\gamma^1_{UL,N}
$$

Step 2 verifies Assumption \ref{ass:concentration:chap1}. Steps 3 and 4 verify Assumption \ref{ass:ratebound}, the bounds on $\mu_N $ and $r_N''$ respectively.  Define
\begin{align*}
\mu_0(q, D, X) &:= \gamma_{L,0} (D, X) + \gamma_{UL,0} (D,X) 1 \{ - q'  \eta_0(D, X) > 0 \} \\
\nabla_D\mu_0(q, D, X) &:= \nabla_D \gamma_{L,0}(D, X) +\nabla_D \gamma_{UL,0} (D,X) 1 \{ - q'  \eta_0(D, X) > 0 \}. 
\end{align*}
Define the first-stage error terms
\begin{align*}
S_1 (D,X)  &= \gamma_L (D, X) - \gamma_{L,0} (D, X) \\
S_2 (D,X) &= \gamma_{UL} (D,X) - \gamma_{UL,0} (D,X) \\
S_3 (q, D,X) &=  1 \{  q' V(\eta) > 0 \} - 1 \{   q' V(\eta_0) > 0 \}
\end{align*}
and note that
\begin{align*}
\mu(q, D, X)  - \mu_0(q, D, X) &= S_1 (D,X) + S_2 (D,X) 1 \{  q'  \eta(D, X) > 0 \} \\
&+\gamma_{UL,0} (D,X) S_3 (q, D,X). \\
\nabla_D (\mu(q, D, X)  - \mu_0(q, D, X)) &= \nabla_D S_1 (D,X) + \nabla_D S_2 (D,X) 1 \{  q'  \eta(D, X) > 0 \} \\
&+ \nabla_D \gamma_{UL,0} (D,X) S_3 (q, D,X).
\end{align*}

\textbf{Step 2}.  I verify Assumption \ref{ass:concentration:chap1}. The function class 
$$
\mathcal{F}_{\xi} = \{  g(W, q, \xi(q)), q \in \mathcal{S}^{d-1} \} \subseteq \mathcal{H}_{\xi} + \mathcal{Y}_{\eta} - \mathcal{M}_{\eta},
$$
where
$$
\mathcal{Y}_{\eta} = (Y_U - Y_L)  \cdot  \mathcal{L}_{\eta}  \cdot \mathcal{I}_{\eta}  + Y_L
$$
and
\begin{align*}
\mathcal{M}_{\eta} &= \{ q' V(\eta) \mu (q, D, X), \quad q \in \mathcal{S}^{d-1} \} \\
& \subseteq \mathcal{L}_{\eta} \cdot (\gamma_{L} (D,X) + \gamma_{UL} (D,X) \cdot  \mathcal{I}_{\eta})
 \end{align*}
and
\begin{align*}
 \mathcal{H}_{\xi}&= \{ q' \nabla_D \mu(q, D, X), \quad q \in \mathcal{S}^{d-1} \} \\
 &\subseteq  \mathcal{L}_{\nabla_D \gamma_L} + \mathcal{L}_{\nabla \gamma_{UL}} \mathcal{I}_{\eta},
 \end{align*}
where all the classes above are $P$-Donsker and obey \eqref{eq:entropyboundapp}, which implies Assumption \ref{ass:concentration:chap1} (2).  The integrable envelope $F_{\Xi}$ for $\mathcal{F}_{\xi}$ can be taken as
\begin{align}
\label{eq:fxi}
F_{\Xi} :&= \| \nabla_D \gamma_L (D,X) \| + \| \nabla_D \gamma_{UL} (D,X) \|  \\
&+ \| V (\eta) \| (Y_L- \gamma_{L,0} (D,X) + M_{\gamma}) + \| V(\eta) \| ( M_{UL} + M_{\gamma}),
\end{align}
which is integrable by Assumption \ref{ass:regcond:apd}.   

\textbf{Step 3. Bound on $\mu_N$}. Decompose the first-stage estimation error
\begin{align*}
g(W, q, \xi(q))-g(W, q, \xi_0(q)) &= q' (\partial_D  \mu (q, D, X) - \partial_D  \mu_0 (q, D, X)) \\
&+q'V ( \eta_0) (  Y(q, \eta) - Y(q, \eta_0)) \\
&- q'V ( \eta_0) (  \mu (q, D, X) -   \mu_0 (q, D, X) ) \\
&+ (q'V ( \eta) - q'V ( \eta_0)) (  Y(q, \eta) - Y(q, \eta_0)) \\
&- (q'V ( \eta) - q'V ( \eta_0))  (  \mu (q, D, X) -   \mu_0 (q, D, X) ) \\
&= \sum_{j=1}^5 K_j(W,q, \xi),
 \end{align*}
Integration by parts implies
 $$
 \E [ K_1 (W,q, \xi) + K_3 (W,q, \xi)  ]  = 0 \quad \forall q \in \mathcal{S}^{d-1}. 
 $$
Note that the terms $K_2(W, q, \xi)$ and $K_4(W, q, \xi)$ coincide with $B_1(W, q, \eta)$ and $B_2(W, q, \eta)$, that is,
 \begin{align*}
 K_2(W, q, \xi)= K_2(W, q, \eta)=B_1(W, q, \eta), \\
  K_4(W, q, \xi) = K_4(W, q, \eta) = B_2(W, q, \eta).
\end{align*}
 Lemma \ref{lem:bias2} has shown that
 $$
\sup_{\eta \in \mathcal{T}_N} | \E  K_2(W,q, \xi) +  K_4(W,q, \xi)  | \leq 2 M_{UL} \bar C_f    (\eta^{\infty}_N)^2.
$$
Cauchy inequality gives
\begin{align}
\label{eq:murate}
&\E (\mu(q, D, X)  - \mu_0(q, D, X))^2 \\
&\leq 3 (\E S^2_1 (D,X) + \E S^2_2 (D,X) + \E \gamma^2_{UL,0} (D,X) S^2_3 (q, D,X) )\nonumber  \\
&\leq 3 (\E S^2_1 (D,X) + \E S^2_2 (D,X) + M^2_{UL} \Pr ( \mathcal{E}_{+-} (q))) \nonumber \\
&= O (\gamma^2_{N}  +  \eta^{\infty}_N). \nonumber
\end{align} 
Cauchy inequality gives 
 \begin{align*}
 \sup_{ \xi \in \Xi_N}  \E | K_5(W,q, \xi) | &\leq  \sqrt{ \E \| V(\eta) - V(\eta_0) \|^2  } \sqrt{(\E (\mu (q, D, X) -   \mu_0 (q, D, X))^2 } \\
&= O ( (\eta^{\infty}_N)^{3/2} + \eta^{\infty}_N \gamma_N).
\end{align*}

\textbf{Step 4.  Bound on $r_N''$}. For $\eta \in \mathcal{T}_N$, the following bounds hold for $j=2$ and $j=4$ follow from \eqref{eq:pp0}
\begin{align*}
\sup_{\eta \in \mathcal{T}_N}  \E  K^2_j(W,q, \eta)  &\leq \E \| V(\eta) - V ( \eta_0) \|^2  (Y_U - U_L)^2 1\{  \mathcal{E}_{+-}(q) \}  \leq M^2_{UL} (\eta^{\infty}_N)^2, \quad  j=2,4.
 \end{align*}
 For $\sup_{\eta \in \mathcal{T}_N} \| V(\eta) \| \leq \bar C \text{ a.s. }$ and $\sup_{\eta \in \mathcal{T}_N} \| V(\eta) - V(\eta_0)  \| \leq 2 \bar C \text{ a.s.}$, note that
$$
 \sup_{ \xi \in \Xi_N}  \E  K^2_j(W,q, \xi)  \leq 4 \bar C^2  \sup_{ \xi \in \Xi_N} \E (\mu (q, D, X) -   \mu_0 (q, D, X))^2 = O (\gamma^2_N + \eta^{\infty}_N), \quad j=3,5.
 $$
 and
\begin{align*}
 \sup_{ \xi \in \Xi_N}  \E  K^2_1(W,q, \xi)  &\leq 3 (\E \| \nabla_D \gamma_L (D, X) - \nabla_D \gamma_{L,0} (D, X)] \|^2 \\
 &+ \E \| \nabla_D \gamma_{UL} (D, X) - \nabla_D\gamma_{UL,0} (D, X) \|^2  \\
 & + C^2_{UL} \Pr ( \mathcal{E}_{+-} (q)))  = O (\gamma^2_{N}  +  \eta^{\infty}_N) 
 \end{align*}
Collecting the bounds gives $r_N' = O (\gamma_N + (\eta^{\infty}_N)^{1/2})$.

\end{proof}

\renewcommand{\thesection}{C}

  \section{Supplementary Tables }

\renewcommand{\theequation}{C.\arabic{equation}}
\renewcommand{\thelemma}{C.\arabic{lemma}}
\renewcommand{\thetable}{C.\arabic{table}}

\setcounter{equation}{0}
\setcounter{section}{0}
\setcounter{lemma}{0}
\medskip

\begin{table}[H]
\centering
\caption{Finite-sample performance of the non-ortho and ortho approaches (true first-stage) }
\begin{tabular}{c|cccccccc}
\toprule
  & \multicolumn{4}{c}{Non-ortho}& \multicolumn{4}{c}{Ortho} \\ 
   \hline \\
 $N$  & Total & Outer & Inner & Rej.freq & Total & Outer & Inner & Rej.freq   \\
  \\  
    & \multicolumn{8}{c}{Panel A: Bracket width $\Delta = 1$}\\
    \\
$250$ & 0.46 & 0.45 & 0.44 & 0.07 & 0.14 & 0.13 & 0.12 & 0.06 \\ 
$500$ & 0.34 & 0.33 & 0.33 & 0.05 & 0.10 & 0.09 & 0.08 & 0.06 \\ 
$700$ &   0.28 & 0.27 & 0.27 & 0.03 & 0.08 & 0.07 & 0.07 & 0.04 \\ 
 $1, 000$ & 0.24 & 0.23 & 0.23 & 0.03 & 0.07 & 0.06 & 0.06 & 0.08 \\ 
   & \multicolumn{8}{c}{Panel B: Bracket width $\Delta = 2$}\\
   \\
 $250$ & 0.49 & 0.46 & 0.45 & 0.06 & 0.18 & 0.15 & 0.14 & 0.08 \\ 
$500$ & 0.36 & 0.34 & 0.33 & 0.05 & 0.12 & 0.10 & 0.10 & 0.07 \\ 
$700$ &   0.29 & 0.27 & 0.27 & 0.03 & 0.10 & 0.09 & 0.08 & 0.03 \\ 
 $1, 000$ &  0.25 & 0.23 & 0.23 & 0.03 & 0.09 & 0.08 & 0.07 & 0.09 \\ 
 \\
   & \multicolumn{8}{c}{Panel C: Bracket width $\Delta = 3$}\\
   \\
 $250$ & 0.52 & 0.48 & 0.46 & 0.06 & 0.22 & 0.18 & 0.16 & 0.04 \\  
 $500$ & 0.38 & 0.34 & 0.33 & 0.06 & 0.15 & 0.13 & 0.12 & 0.05 \\
 $700$ &   0.31 & 0.28 & 0.28 & 0.03 & 0.13 & 0.11 & 0.10 & 0.03 \\ 
 $1, 000$ &  0.26 & 0.24 & 0.23 & 0.03 & 0.11 & 0.09 & 0.09 & 0.06 \\  
\bottomrule
\end{tabular}
\label{tab:simsapp}
\caption*{Notes. Results are based on 10, 000 simulation runs.  Panels A, B and C correspond to the bin width $\Delta = 1, 2, 3$.    Table shows the total risk \eqref{eq:total}, the outer and inner risks \eqref{eq:outerinner}, and the rejection frequency \eqref{eq:boot} for the nominal size $\alpha = 0.05$. The supremum over the unit circumference  is approximated by the maximum over the grid consisting of $ 50$ evenly spaced points.  Columns (1--4) and (5--8) correspond to the non-orthogonal and the orthogonal second-stage. The number of bootstrap repetitions $B=2, 000$. The true support function is in \eqref{eq:sigmatrue}.  The estimated support function is based on the true treatment fitted values (i.e., true $\alpha_1, \alpha_2$), zero outcome fitted values (non-ortho) and Lasso outcome fitted values (ortho).  }
\end{table}

\begin{table}[H]
\centering
\caption{Finite-sample performance of estimators }
\begin{tabular}{c|cccccccc}
\toprule
  & \multicolumn{4}{c}{Lasso-based (non-ortho)}& \multicolumn{4}{c}{Series-based (ortho)} \\ 
   \hline \\
 $N$  & Total & Outer & Inner & Rej.freq & Total & Outer & Inner & Rej.freq   \\
  \\  
    & \multicolumn{8}{c}{Panel A: Bracket width $\Delta = 1$}\\
    \\
$250$ & 0.65 & 0.64 & 0.63 & 0.02 & 0.21 & 0.16 & 0.20 & 0.17 \\
$500$ & 0.48 & 0.47 & 0.47 & 0.00 & 0.19 & 0.09 & 0.19 & 0.43 \\
$700$ &  0.42 & 0.41 & 0.41 & 0.00 & 0.18 & 0.08 & 0.18 & 0.62 \\ 
 $1, 000$ & 0.36 & 0.35 & 0.35 & 0.00 & 0.18 & 0.06 & 0.18 & 1.00 \\ 
   & \multicolumn{8}{c}{Panel B: Bracket width $\Delta = 2$}\\
   \\
 $250$ & 0.67 & 0.65 & 0.64 & 0.03 & 0.34 & 0.24 & 0.33 & 0.24 \\
$500$ & 0.49 & 0.48 & 0.47 & 0.00 & 0.33 & 0.13 & 0.33 & 0.87 \\
$700$ &   0.43 & 0.42 & 0.41 & 0.00 & 0.32 & 0.11 & 0.32 & 0.99 \\ 
 $1, 000$ &  0.37 & 0.36 & 0.35 & 0.00 & 0.32 & 0.08 & 0.32 & 1.00 \\ 
 \\
   & \multicolumn{8}{c}{Panel C: Bracket width $\Delta = 3$}\\
   \\
 $250$ & 0.69 & 0.67 & 0.64 & 0.04 & 0.48 & 0.33 & 0.47 & 0.18 \\  
 $500$ &0.51 & 0.49 & 0.48 & 0.00 & 0.47 & 0.18 & 0.47 & 0.95 \\ 
 $700$ &  0.44 & 0.42 & 0.42 & 0.00 & 0.47 & 0.14 & 0.47 & 1.00 \\ 
 $1, 000$ &  0.38 & 0.36 & 0.36 & 0.00 & 0.47 & 0.11 & 0.47 & 1.00 \\ 
\bottomrule
\end{tabular}
\label{tab:simsapp2}
\caption*{Notes. Results are based on 10, 000 simulation runs.  Panels A, B and C correspond to the bin width $\Delta = 1, 2, 3$.    Table shows the total risk \eqref{eq:total}, the outer and inner risks \eqref{eq:outerinner}, and the rejection frequency \eqref{eq:boot} for the nominal size $\alpha = 0.05$. The supremum over the unit circumference  is approximated by the maximum over the grid consisting of $ 50$ evenly spaced points.  The number of bootstrap repetitions $B=2, 000$. The true support function is in \eqref{eq:sigmatrue}.  }
\end{table}

 \bibliography{my_new_bibtex}
\bibliographystyle{apalike}
\end{document}